\def\year{2022}\relax
\documentclass[letterpaper]{article} % DO NOT CHANGE THIS
\usepackage{aaai22}  % DO NOT CHANGE THIS
\usepackage{times}  % DO NOT CHANGE THIS
\usepackage{helvet}  % DO NOT CHANGE THIS
\usepackage{courier}  % DO NOT CHANGE THIS
\usepackage[hyphens]{url}  % DO NOT CHANGE THIS
\usepackage{graphicx} % DO NOT CHANGE THIS
\usepackage{natbib}  % DO NOT CHANGE THIS AND DO NOT ADD ANY OPTIONS TO IT
\usepackage{caption} % DO NOT CHANGE THIS AND DO NOT ADD ANY OPTIONS TO IT
\DeclareCaptionStyle{ruled}{labelfont=normalfont,labelsep=colon,strut=off} % DO NOT CHANGE THIS
\usepackage[ruled,linesnumbered, nofillcomment]{algorithm2e}
\usepackage{amssymb}
\usepackage{mathrsfs}
\usepackage{bbm}
\usepackage{dsfont}
\usepackage{bbold}
\usepackage{mathbbol}
\usepackage{newtxmath}
\usepackage{color}
\usepackage{float}
\usepackage{subfigure}
\usepackage{booktabs}
\usepackage{epstopdf}
\usepackage{newfloat}
\usepackage{listings}
\usepackage{stfloats}

\newtheorem{theorem}{Theorem}
\newtheorem{asm}{Assumption}
\newtheorem{dfi}{Definition}
\newtheorem{lea}{Lemma}

\floatstyle{ruled}
\newfloat{listing}{tb}{lst}{}
\floatname{listing}{Listing}
\title{The $f$-Divergence Reinforcement Learning Framework}
\author {
    Chen Gong*\textsuperscript{\rm 1,2,3},
    Qiang He* \textsuperscript{\rm 1,2},
    Yunpeng Bai* \textsuperscript{\rm 1,2},
    Zhou Yang \textsuperscript{\rm 3},
    Xiaoyu Chen\textsuperscript{\rm 1},
    Xinwen Hou**\textsuperscript{\rm 1}, 
    Xianjie Zhang \textsuperscript{\rm 4},
    Yu Liu \textsuperscript{\rm 1},
    Guoliang Fan \textsuperscript{\rm 1}
}
\affiliations {
    \textsuperscript{\rm 1} Institute of Automation, Chinese Academy of Sciences\\
    \textsuperscript{\rm 2} University of Chinese Academy of Sciences, School of Artificial Intelligence\\
    \textsuperscript{\rm 3} School of Computing and Information Systems, Singapore Management University \\ 
    \textsuperscript{\rm 4} School of Software, Dalian University of Technology \\
    \{gongchen2020, heqing2019, baiyunpeng2020, xiaoyu.chen, xinwen.hou, guoliang.fan, yu.liu\}@ia.ac.cn, zyang@smu.edu.sg, zhangxianjie@mail.dlut.edu.cn
}
\usepackage{bibentry}
\begin{document}

\maketitle

\begin{abstract}
The framework of deep reinforcement learning (DRL) provides a powerful and widely applicable mathematical formalization for sequential decision-making. This paper present a novel DRL framework, termed \emph{$f$-Divergence Reinforcement Learning (FRL)}. In FRL, the policy evaluation and policy improvement phases are simultaneously performed by minimizing the $f$-divergence between the learning policy and sampling policy, which is distinct from conventional DRL algorithms that aim to maximize the expected cumulative rewards. We theoretically prove that minimizing such $f$-divergence can make the learning policy converge to the optimal policy. Besides, we convert the process of training agents in FRL framework to a saddle-point optimization problem with a specific $f$ function through Fenchel conjugate, which forms new methods for policy evaluation and policy improvement. Through mathematical proofs and empirical evaluation, we demonstrate that the FRL framework has two advantages: (1) policy evaluation and policy improvement processes are performed simultaneously and (2) the issues of overestimating value function are naturally alleviated. To evaluate the effectiveness of the FRL framework, we conduct experiments on Atari 2600 video games and show that agents trained in the FRL framework match or surpass the baseline DRL algorithms. 

% \noindent \textbf{Key words:} $f-$Divergence; Fenchel Conjugate; Reinforcement Learning; Policy gradient; Value Estimation
\end{abstract}
\section{Introduction}
Deep reinforcement learning (DRL) algorithms, which learn to make decisions from trial and error, have recently achieved successes in a wide variety of fields~\cite{mnih_human-level_2015,levine_end--end_2016,silver_mastering_2017}. Researchers generally consider reinforcement learning (RL) from views of dynamic programming \cite{sutton_reinforcement_2018}, Bayes inference \cite{ghavamzadeh2015bayesian}, or linear programming~\cite{dualdice,algaedice}. 

The majority of DRL algorithms train agents to learn a {\em learning policy} that can maximize the expected cumulative rewards from the trajectories generated by the {\em sampling policy}. The learning policy and sampling policy are defined as follows.
\begin{dfi} The learning policy $\pi$ is the target policy that an agent tries to learn, which is usually represented by a neural network model.
\end{dfi}
\begin{dfi}
The sampling policy $\tilde{\pi}$ is the policy that generates the sampling trajectories during the training phases. 
\end{dfi}
The sampling policy is induced by the learning policy through a specific transformation, \textit{e.g.,} the $\epsilon$-greedy is also used in many previous works, makeing the value function of sampling policy always greater than that of the learning policy \cite{mnih_human-level_2015}.

In the famous Soft Actor-Critic (SAC) algorithm \cite{sac}, the policy gradient formulation is derived from making the learning policy close to ``old'' policies; the latter are represented by a Boltzmann distribution whose energy function of is defined as the negative state-action value function~\cite{landau2013course}.
In the Motivation Section, we show that the sampling policy can be represented as the Boltzmann distribution. As a result, the policy iteration in SAC is a process of minimizing the Kullback-Leibler (KL) divergence between learning policy and sampling policy \cite{sac}.
Besides, we show that when obtaining the optimal policy in a Markov Decision Process (MDP), the Bellman optimality equation, which is induced by the learning policy, and the Bellman expectation equation, which is induced by the sampling policy, should have the same values (see the proof of Eq. (\ref{eq:theorem1})).
This also inspires our proof (see in Lemma \ref{plicy_improvement}) that when minimizing the distance (\textit{i.e.}, $f$-divergence) between the learning and sampling policy can lead the learning policy converge to the optimal policy. 

The above facts motivate us to present a novel DRL framework, termed \emph{$f$-Divergence Reinforcement Learning (FRL)}, which trains agents by minimizing the $f-$divergence between the learning policy and the sampling policy instead of maximizing the expected cumulative rewards as the conventional DRL algorithms do. 
We also show that the FRL framework can be compatible with any $f$ functions that satisfy certain properties (convex, proper and semi-continuous) (see Lemma \ref{lemma:1}). 
We mathematically show that the objective of minimizing the $f-$divergence can be transformed into a saddle-point optimization problem using Fenchel conjugate \cite{nachum2020Duality}, which makes the FRL framework able to evaluate and improve policy simultaneously while the conventional algorithms like SAC \cite{sac} perform the policy evaluation and policy improvement separately. 
We also provide the algorithm of performing policy gradient and policy evaluation that can train agents by solving this saddle-point optimization problem.

We conduct sufficient experiments to evaluate this proposed framework from different perspectives. 
On the one hand, the experiments demonstrate that in the Atari 2600 \cite{atari} environments, agents trained using FRL consistently match or outperform the baselines algorithms investigated.
On the other hand, we empirically show that the notorious problem of overestimating value functions is alleviated in the FRL framework. The above two points highlight the superiority of the FRL framework.

The contributions of this paper are summarised as follows. First, this paper proposes a novel framework called \emph{$f$-Divergence Reinforcement Learning} that provides a new paradigm of training agents. Second, we provide detailed proofs of the theoretical foundations behind the framework. Third, we provide the algorithm of implementing the framework. Then, experiments are conducted to show that the framework can alleviate the overestimation of value functions and outperforms (or at least matches) baselines algorithms. 
It is shown that utilizing the convex duality and the usage of the $f$-divergence concept can benefit the DRL theory, which hopefully provides new perspectives for the development of DRL theory.

\section{Related works}
This section first briefly discusses the developments of conventional DRL algorithms and then introduces that DRL algorithms using Fenchel conjugate and $f-$divergence.

\subsection{DRL Algorithms}
Value iteration methods play essential roles in the DRL research. Deep $Q$-network (DQN) \cite{mnih_human-level_2015} approximates the state-action value function using deep neural networks. But as pointed out by Van et al. \cite{doubledqn}, the $Q$ function can sometimes be overestimated, which results in imprecise value and lead to a sub-optimal policy. To alleviate the issue of overestimation, Van et al.~\cite{doubledqn} proposed Double DQN that utilizes two different networks for value estimation and action choice, respectively. Bellemare et al. proposed Distributional RL (C51) \cite{bellemare_distributional_2017} to address the same issue. Conventional DRL algorithms aim to maximize the expectation of the value function, but C51 directly optimizes the distribution of the value function. C51 cannot generally be minimized using stochastic gradient methods \cite{bellemare_distributional_2017}. To address this problem, Dabney et al.~\cite{dabney_distributional_2018} measured the loss function with Wasserstein distance \cite{arjovsky2017wasserstein}, which is approximated by quantile regression \cite{dabney_distributional_2018}.
Hessel et al. integrated the aforementioned methods and proposed Rainbow~\cite{hessel_rainbow_2018}, which further boosts the performance.

The value-based methods are hard to be applied in situations where the action space is very large or continuous.
Besides, the value-based methods usually need to consume millions of samples for training, revealing inefficiency in sampling~\cite{hessel_rainbow_2018}. As another important class of DRL algorithms, the policy gradient methods can alleviate problems of inefficiency. The policy gradient with function approximation was proposed by \cite{sutton_policy_2000}. Silver et al. proposed the Deterministic Policy Gradient method (DPG) \cite{dpg} to optimize the expected reward with a deterministic policy, avoiding the problem of the large computational cost of a random policy. Lillicrap et al. combined the DPG algorithm with deep neural networks and develop the Deep Deterministic Policy Gradient (DDPG) algorithm \cite{lillicrap_continuous_2015}. A well-known issue in the DPG algorithm is the lack of sufficient exploration. Lillicrap et al. claimed that adding noise drawn from the Ornstein-Uhlenbeck process to actions can help DDPG explore better \cite{OUprocess}. 
% \cite{OUprocess}  Twin Delayed Deep Deterministic Policy Gradient (TD3) \cite{td3} algorithm relieves the overestimation problem in DDPG by the twin network structure.

All the objectives of the aforementioned algorithms are to maximize the expected cumulative rewards. To the best of our knowledge, our framework is the first to study the $f$-divergence between learning policy and sampling policy and prove that the policy evaluation and policy improvement phases are induced by minimizing such $f$-divergence, which is a novel paradigm to train DRL algorithms.

\subsection{Fenchel Conjugate in DRL Algorithms}
The Fenchel conjugate and its variants, which belongs to a mature class of optimization methods, are at the heart of a number of DRL algorithms. 
The computation of $f-$divergence is usually approximated by a maximization problem with Fenchel conjugate \cite{algaedice,dai2020coindice,nachum2020Duality}. In off-policy policy evaluation (OPE), Nachum et al. proposed Dual stationary DIstribution Correction Estimation (DualDICE) \cite{dualdice} to accurately estimate the discounted stationary distribution ratios \cite{liu2018breaking} for any behavior policies that used to generate the static dataset. Zhang et al. extended DualDICE to undiscounted policy evaluation and proposed GenDICE \cite{gendice}. Yang et al. unified the DICE family of estimators as regularized Lagrangian of the same linear program \cite{yang2020off}. In policy gradient, Fenchel AlgaeDICE \cite{algaedice} is proposed to constrain the $f-$divergence between target and behavior policy distribution as a regularization, to exactly obtain the gradient of the off-policy objective.

DICE families separately derive the policy evaluation and policy improvement. However, policy evaluation and policy improvement in the FRL framework can be both derived by minimizing the $f$-divergence.

\section{Preliminaries}
In this section, we briefly review the basis of Fenchel conjugate, $f$-Divergence and DRL.
\subsection{Fenchel Conjugate}
The conjugate tools are important in machine learning and optimization. Fenchel conjugate, which is widely applied,  allows researchers to reformulate hard-to-solve optimization problems into alternative problems that are easier to be solved \cite{gail,dualdice,algaedice,nachum2020Duality,gendice}. 
Details of convex optimization are well explained in works by \cite{rockafellar2015convex,boyd2004convex}, and we refer interested readers to them for further details. Here we briefly introduce the notations and definitions that are related to this paper.

For a function $f:\Omega \to \mathbb{R}$, where $\Omega$ is the input domain and $\mathbb{R}$ is the output domain, we define its Fenchel conjugate $f_\ast:\Omega^\ast \to \mathbb{R}$ as follows:
\begin{equation}
    f_\ast(z) = \max_{x \in \Omega} \langle x,z \rangle - f(x).
\label{eq:conjugate}
\end{equation}
Definition \ref{dfi:3}, \ref{dfi:4} and \ref{dfi:5} explains the properties a function $f$ should have to be {\em convex}, {\em proper} and {\em lower semi-continuous}, respectively \cite{boyd2004convex}.

\begin{dfi}
When a function $f$ is convex, $f:\mathbb{R}^n \to \mathbb{R}$ meet that $\text{dom}(f) \subset \mathbb{R}^n$ is convex set, and $\forall x,y$ in $\text{dom}(f), \forall t \in [0,1], f(tx +(1-t)y) \leq tf(x) + (1-t)f(y)$.
\label{dfi:3}
\end{dfi}
\begin{dfi}
When a function $f$ is proper, $f$ meet that $\{x\} \in \Omega : f(x) < \infty$ is non-empty and $\forall x \in \Omega, f(x) > -\infty$.
\label{dfi:4}
\end{dfi}
\begin{dfi}
When a function $f$ is lower semi-continuous, $f$ meet that $\{x\} \in \Omega : f(x) > \alpha$ is an open set for $\forall \alpha \in \mathbb{R}$.
\label{dfi:5}
\end{dfi}

If a function $f$ is proper, convex, lower semi-continuous function, its Fenchel conjugate function $f_\ast$ is also proper, convex, lower semi-continuous. Rockafellar showed that for such a function $f$, $f_{\ast\ast}$ (the Fenchel conjugate of $f_\ast$) is $f$ itself \cite{rockafellar2015convex}. In other words, we have the following equation:
\begin{equation}
    f(x) = \max_{z \in \Omega^\ast} \langle x,z \rangle - f_\ast(z),
\end{equation}
where $\Omega^\ast$ denotes the domain of $f_\ast$. In this paper, a function $f$ used for computing $f$-divergence is proper, convex, lower semi-continuous unless otherwise stated. 

\subsection{$f-$Divergence}
$f$-Divergence is used to measure the difference between two probability distributions \cite{f_divergence}. For a function $f$ and a distribution $q$ over the domain $\mathcal{H}$, the $f-$Divergence is defined as,
\begin{equation}
    \mathbb{D}_f(x \parallel q) := \mathbb{E}_{h \sim q}\left[ f\left(\frac{x(h)}{q(h)}\right) \right].
\label{eq:f_divergence}
\end{equation}
$x \in \Delta (\mathcal{H})$, the simplex over $\mathcal{H}$, and $\mathbb{D}_f(x \parallel q)$ measure the divergence between $x$ and $q$. Beside, the domain of $\mathbb{D}_f(x \parallel q)$ may be extend to the set of real-valued functions $x: \mathcal{H} \to \mathbb{R}$ \cite{nachum2020Duality}. Different $f$ makes different distribution discrepancy. The chosen $f$ function needs to satisfy two key conditions: 1) $f$ function is a convex function; 2) $f(1) = 0$ \cite{rockafellar2015convex}. In the family of $f-$divergence, the KL-divergence $\mathbb{D}_{\text{KL}}(x\|q) := \mathbb{E}_{h\sim x}[\log(x(h)/q(h))]$ is admittedly the most commonly used one \cite{sac,bellemare_distributional_2017}. Actually, when $f(x) = x \log x$, Eq. (\ref{eq:f_divergence}) specifies the KL-Divergence.

\subsection{Reinforcement Learning}
The standard paradigm of training RL algorithm is to let an agent interact with the environment and learn from the feedback. The RL agent needs to continuously explore the environment and learn from its trial-and-error experiences to be properly trained. We abstract the standard RL training as a Markov Decision Process (MDP), which is defined by a tuple $(\mathcal{S ,A ,R , T} ,\rho_0 ,\gamma)$, \textit{i.e.}, a finite set of states $\mathcal{S}$; a finite set of actions $\mathcal{A}$; a reward function: $\mathcal{R}: \mathcal{S} \times \mathcal{A} \to \mathbb{R}$; a transition function $T: \mathcal{S} \times \mathcal{A} \to \Delta(\mathcal{S})$, where $\Delta(\mathcal{S})$ means a probability distribution over $\mathcal{S}$; an initial state distribution $\rho_0$; and a discount factor: $\gamma \in [0,1)$.

The policy is a function $\pi: \mathcal{S} \to \Delta(\mathcal{A})$ that tells an agent what actions to take under different states, and we define that the policy belongs a cluster $\Pi = \{\pi: \mathcal{S} \mapsto \Delta(\mathcal{A}) \}$. In each episode, an agent starts at the initial state $s_0$ and stops at a termination state. At each time step $t$, an agent takes an action with respect to the current state, i.e., $a_t \sim \pi(\cdot|s_t)$. Then, the agent receives an instant reward $r_t = r(s_t,a_t)$ from the environment and move a new state $s_{t+1} \sim T(s_t,a_t)$. The process continues until the agent arrives at a termination state. The discounted sum of rewards from state $s$ is defined as $R_t(s,a) = \sum_{i=t}^\infty \gamma^{i-t}r_i$, which is simplified as $R(s,a)$ in this paper. The agent aims to find the optimal policy $\pi^\ast$ that maximizes the expected rewards $\mathbb{E}_{\tau \sim \pi}[R_0|s_0]$, where $\tau$ is a trajectory $(s_0,a_0,r_0,s_1,\cdots)$. The state value function is defined as $V^\pi(s_t) = \mathbb{E}_{\tau \sim \pi}[R_t|s_t]$, the expected rewards of following the policy $\pi$ in state $s$. Similarly, the action-state value function is defined as $Q^\pi(s_t,a_t) = \mathbb{E}_{\tau \sim \pi}[R_t|s_t,a_t]$. Based on the principle of dynamic programming, $V^\pi$ and $Q^\pi$ are usually estimated by Bellman expectation equation:
\begin{equation}
\begin{gathered}
    V^\pi(s) = \mathbb{E}_{a \sim \pi( \cdot \mid s)}[Q^\pi(s,a)], \quad \forall s \in \mathcal{S}, a \in \mathcal{A} \\
    Q^\pi(s,a) = r(s,a) + \gamma \mathbb{E}_{s' \sim P(s,a)} [V^\pi(s')].
\end{gathered}
\label{eq:bellman_expectation}
\end{equation}

A stationary and deterministic policy can maximize $V^\pi(s)$ for all $s \in \mathcal{S}$ and $Q^\pi(s,a)$ for all $s \in \mathcal{S}, a \in \mathcal{A}$ simultaneously \cite{puterman2014markov} in value-based methods. For simplicity, when obtaining the optimal policy $\pi^\ast$, we denote the state value function and action value function under $\pi^\ast$ as $V^\ast$ and $Q^\ast$. The $V^\ast$ and $Q^\ast$ satisfy the following Bellman optimality equations \cite{bellman1956dynamic}:
\begin{equation}
    \begin{gathered}
        V^\ast(s) = \max_{a \in \mathcal{A}} Q^\ast(s,a), \quad \forall s \in \mathcal{S}, a \in \mathcal{A} \\
        Q^\ast(s,a) = r(s,a) + \gamma \mathbb{E}_{s' \sim P(s,a)}[V^\ast(s')].
    \end{gathered}
    \label{eq: bellman equation in method section}
\end{equation}
Actually, when we obtain $Q^\ast$, $\pi^\ast$ can be obtained by greedily choosing actions that have the highest values, \textit{i.e.}, $\pi^{\ast}(s)=\arg \max_{a \in \mathcal{A}} Q^{\ast}(s, a),$ for all $ s \in \mathcal{S}$. We utilize $\pi_Q$ to represent the procedure of turning a $Q$ function into greedy policy, so $\pi^\ast = \pi_{Q^\ast}$.
DRL usually uses neural networks to approximate both the policy and value function. Assuming that $\theta$ represents the parameters of a new work, we use $V^\pi(s; \theta)$ and $\pi(a|s;\theta)$ to denote the value function and policy. The agents learn to update $\theta$ over the trajectory $\tau$ with stochastic gradient descent.

\section{$f-$Divergence Reinforcement Learning}
In this section, we firstly introduce the theoretical foundation behind the FRL framework as well as the details of the $f-$divergence Reinforcement Learning Framework. Secondly, we explain how to use Fenchel conjugate to transforming the training of agents into a saddle point optimization problem in FRL. Finally, we discuss an example of using a specific $f$ function in our framework.

\subsection{Theoretical Foundation of FRL}

In this paper, we present a novel framework, termed as the \textit{$f-$divergence reinforcement learning (FRL)} framework, which trains agents by minimizing $f-$divergence between the learning policy and sampling policy, instead of maximizing the expected cumulative rewards as conventional DRL algorithms. Formally speaking, our framework solve the below optimization problem:
\begin{equation}
    \pi^\ast = \min_{\pi \in \Pi} \mathbb{D}_f(\pi\|\tilde{\pi}).
\label{eq:targe}
\end{equation}
where $\pi$ is the learning policy and $\tilde{\pi}$ is the sampling policy. The essential theoretical foundation of the FRL framework is the fact that we can achieve the optimal learning policy by minimizing the distance between the learning policy and sampling policy (Lemma \ref{plicy_improvement}). We gradually discuss the framework by (1) explaining the motivation behind the framework and (2) proving Lemma \ref{plicy_improvement}.

\subsubsection{Motivation}
Our motivation of designing the FRL framework comes from two aspects: (1) the learning policy is asymptotic to the sampling policy when DRL algorithms converge and (2) a prior algorithm perform policy improvement by minimizing KL-divergence between two policies. 

On the one side, under the optimal policy $\pi^\ast$, the Bellman equation is equivalent to the Bellman optimality equations. In other words, we have the following equation:
\begin{equation}
    \begin{aligned}
        Q^\ast(s,a) = & r(s,a) + \gamma \mathbb{E}_{s' \sim P(s,a),a'\sim \pi^\ast(\cdot \mid s')} [Q^\ast(s',a')] \\
        = & r(s,a) + \gamma \mathbb{E}_{s' \sim P(s,a)}\left[\max_{a'\in \mathcal{A}}\left[Q^\ast(s',a')\right]\right].
    \end{aligned}
    \label{eq:theorem1}
\end{equation}
The top right formula is the Bellman equation, and the bottom formula is the Bellman optimality equation when the learning policy is optimal. 
Intuitively speaking, if we regard the maximum operator as a deterministic policy, the sampling policy $\tilde{\pi}$ is equivalent to the learning policy $\pi$ when the latter is optimal.
Due to the limited space of the paper, we put the formal proof of Eq.~(\ref{eq:theorem1}) into the Appendix A.

When the learning policy converges to the optimal policy $\pi^\ast$, the $\mathbb{E}_{a'\sim \pi^\ast(\cdot \mid s')} [Q^\ast(s',a')]$ equals to $\max_{a'\in \mathcal{A}}[Q^\ast(s',a')]$; there is no difference between using Bellman equation and Bellman optimality equation to calculate $Q^\pi(s,a)$ for all $s \in \mathcal{S}, a\in \mathcal{A}$. We get the same value from the two equations, which is viewed as the consistency between the learning policy and sampling policy. We explain how to generate sampling policy through the learning policy. Firstly, the action-value $Q^\pi(s,a)$ function of learning policy $\pi$ is calculated by Bellman equation; secondly, for a state, we select the action that leads to the highest action value: $a = \arg\max_{a'} Q^\pi(s,a')$. The distribution of all the selected actions in each state is the sampling policy, $\tilde{\pi}(a|s) = \arg\max_{a'} Q^\pi(s,a') $.

% We assume that the attack has no prior information about the victim agent, e.g., parameters

On the other side, the Soft Actor Critic algorithm (SAC) algorithm performs the policy improvement by directly minimizing the expected KL divergence between the learning policy and ``old'' policies; the latter are represented as the Boltzmann distribution. The energy function of Boltzmann distribution is the negative of value function~\cite{haarnoja_soft_2018}. In other words, the improved policy (\textit{i.e.}, $\pi_{\text{new}}$) can be obtained using the following operation:
\begin{equation}
    \pi_{\text{new}}=\arg\min_{\pi'\in \prod}\mathbb{D}_{\mathrm{KL}}\left(\pi'\left(\cdot \mid s\right) \left\| \frac{\exp \left(Q^{\pi_{\text{old}}}\left(s, \cdot\right)\right)}{Z^{\pi_{\text{old}}}\left(s\right)}\right)\right.
\label{eq:sac}
\end{equation}
where the $Q^{\pi_{\text{old}}}(s,\cdot)$ is the $Q$ function of the previous policy. The partition function $Z^{\pi_{\text{old}}}(s)$ normalizes this distribution.

In Eq. (\ref{eq:sac}), the $\pi'(\cdot|s)$ is considered as the learning policy. Actually, the learning policy in each iteration is to be constrained as the Boltzmann distribution, and the action selected to interact with environment is generated with the Boltzmann distribution. Therefore, we consider that the samples are taken from the Boltzmann distribution, which is regarded as the sampling policy: $\tilde{\pi} = \frac{\exp (Q^\pi(s,\cdot))}{Z^\pi(s)}$. When the SAC algorithm converges, the KL-divergence in Eq. (\ref{eq:sac}) is close to zero, \textit{i.e.}, $\pi'$ equals to $\frac{\exp \left(Q^{\pi_{\text{old}}}\left(s, \cdot\right)\right)}{Z^{\pi_{\text{old}}}\left(s\right)}$. 
So the consistency between learning policy and sampling policy also exists, which is inline with the insights from Eq. (\ref{eq:theorem1}).

The above phenomenons inspire us to consider the ``distance" between the learning and sampling policy and propose the FRL framework.

% \begin{theorem}
% Given a $\epsilon$-greedy policy $\tilde{\pi}$ w.r.t. policy $\pi$, then $Q^{\tilde{\pi}}$ converges to $Q^{\pi}$.
% \label{theorem:2}
% \end{theorem}

% \bigskip

\subsubsection{Proof of Policy Improvement}
We now show that the optimal policy can be obtained by solving the optimization problem in Eq. (\ref{eq:targe}). The proof can be deduced to proving that there exists a method to minimize $f$-divergence that can gradually improve state-action value function of learning policy.

It is impossible to obtain an optimal policy $\pi^\ast$ by minimizing the $f$ divergence between the learning policy and an arbitrary sampling policy. We will show that FRL can work when the sampling policy has larger state-action value function than the learning policy. 
In SAC, a conventional RL algorithm, the improvement of the learning policy $\pi$ is done by minimizing the distance (\textit{i.e.}, KL-divergence) between $\pi$ and the old polices $\tilde{\pi}$ (see Eq. (\ref{eq:sac})). The old polices used in SAC also has larger state-action value function than the learning policy.

Then, we propose and prove the Lemma \ref{plicy_improvement}, saying that minimizing the $f$-divergence between the learning policy and sampling policy can improve the state-action value function obtained by the learning policy.
\begin{lea}
    (Fenchel Policy Improvement) Let $\pi_{\text{old}} \in \prod$ and $\pi_{\text{new}}$ present the optimizer of the minimization problem defined in Eq. (\ref{eq:targe}). The $\pi_{\text{old}}$ is the policy before iteration, and $\pi_{\text{new}}$ is the policy after iteration. For all $(s,a) \in \mathcal{S} \times \mathcal{A}$ with $|\mathcal{A}| < \infty$, we have $Q^{\pi_{\text{old}}}(s,a) \leq Q^{\pi_{\text{new}}}(s,a)$. 
    \label{plicy_improvement}
\end{lea}
For the formal proof of Lemma \ref{plicy_improvement}, we refer readers to Appendix A. Now we explain that solving the optimization problem in Eq. (\ref{eq:targe}) can make the learning policy converge to the optimal policy.

We first explain that the learning policy converges. Let $\pi_i$ be the policy at iteration $i$. Lemma \ref{plicy_improvement} suggests that $Q^{\pi_i}$ is monotonically increasing (\textit{i.e.}, $Q^{\pi_i} \geq Q^{\pi_{i-1}}$ ). Besides, the state-action value function for any policy  $Q^\pi$ is bounded because the rewards are always bounded. The two conditions guarantee $Q^{\pi_i}$ to converge, and the learning policy converges to some $\pi^\ast$.

Next, we explain that the converged policy $\pi^\ast$ is indeed optimal. At convergence, we have that $D_f(\pi^\ast \| \tilde{\pi^\ast}) \leq D_f(\pi \| \tilde{\pi^\ast})$ for all $\pi \in \prod, \pi \neq \pi^\ast$. Utilizing the same process of proof  as in the proof of Lemma \ref{plicy_improvement}, we get the results that $ Q^{\pi}(s,a) \leq Q^{\pi^\ast}(s,a)$ for all $(s,a) \in \mathcal{S} \times \mathcal{A}$. Therefore, $\pi^\ast$ is optimal in $\prod$. Based on the above analysis, we present one of the key contribution in FRL: the policy improvement phase can be operated by minimizing the $f$ divergence between learning policy and sampling policy. Such policy improvement method can lead to the optimal policy $\pi^\ast$.

\subsection{Saddle Point Reformulation}
Directly minimizing $f$-divergence is non-trivial so we explain how to convert the problem into a saddle point optimization problem that is easier to be solved. Following the settings in \cite{nachum2020Duality}, we first make the following two assumptions.
\begin{asm}
For any state-action pair $(s,a)$, $\pi(s,a) > 0$ and $\tilde{\pi}(s,a) > 0$. Besides, the $\frac{\pi}{\tilde{\pi}}$ are bounded by a finite constant C: $\|\frac{\pi}{\tilde{\pi}}\|_\infty \leq C$.
\label{asm:1}
\end{asm}
\begin{asm}
The function $f$ is convex within domain $\Omega$ and its derivative $f'$ is continuous. The Fenchel conjugate of $f$, denoted by $f_\ast$, , is closed and strictly convex.  The derivative of $f_\ast$, denoted by $f_\ast'$, is continuous and satisfies $\|f_\ast'\|_\infty \leq C$.
\label{asm:3}
\end{asm}
The two assumptions are made in the work using $f$-divergence as a Lagrange regularization term for training agents and the Assumption 1 is called {\em reference distribution property} \cite{nachum2020Duality}.
Using Fenchel conjugate, the Eq. (\ref{eq:targe}) can be rewritten as,
\begin{equation}
    \pi^\ast = \min_{\pi \in \Pi} \mathbb{E}_{\tilde{\pi}}\left[ \max_{\omega \in \mathbb{R}} \left\langle \pi/\tilde{\pi}, \omega \right\rangle - f_\ast(\omega) \right].
\label{eq:targe_f}
\end{equation}
As we mentioned in earlier sections, the $f$ function discussed in this paper is convex, proper and semi-continuous. As a result, for any given policy $\pi$, the maximizer of the inner function in Eq. (\ref{eq:targe_f}) (\textit{i.e.}, $\max_{\omega \in \mathbb{R}} \left\langle \pi/\tilde{\pi}, \omega \right\rangle - f_\ast(\omega)$) surely exists  \cite{dai2017learning}. Then we show that Eq. (\ref{eq:targe_f}) can be converted into a saddle point optimization problem via proving the Lemma \ref{lemma:1}.
\begin{lea}
Let $\xi$ is a random variable: $\forall \xi \in \Xi$, a function $g(\cdot,\xi) \to (-\infty,+\infty)$ is a proper, convex, lower semi-continuous function, then the following equation holds:
$$\mathbb{E}_{\xi}[\max_{\omega \in \mathbb{R}} g(\omega,\xi)] = \max_{\omega(\cdot) \in \mathcal{G}(\Xi)} \mathbb{E}_\xi[g(\omega(\xi),\xi)],
$$
where $\mathcal{G}(\Xi) = \{\omega(\cdot): \Xi \to \mathbb{R}\}$ is the functional space.
\label{lemma:1}
\end{lea}
The proof of Lemma \ref{lemma:1} is in Appendix B. As we mentioned in Preliminaries Section, if the $f$ funcion is proper, convex, lower semi-continuous, it Fenchel conjugate function $f_\ast$ is also proper, convex, lower semi-continuous.  Having the Lemma \ref{lemma:1}, we can replace the inner max operation over scalar $\omega$ to a max operation over function $\omega: \mathcal{S} \times \mathcal{A} \to \mathbb{R} $. As a result, we can rewrite Eq. (\ref{eq:targe}) into a saddle-point optimization problem:
\begin{equation}
\begin{aligned}
    \mathcal{L}_f (\pi^\ast, \omega^\ast)
    % = & \min_{\pi \in \Pi}\mathbb{D}_f(\pi\|\tilde{\pi}) \\
    = & \min_{\pi \in \Pi} \max_{\omega(\cdot) \in \mathcal{G}(\Xi)}  \\
    & \mathbb{E}_\pi\left[ \omega\left(\pi/\tilde{\pi}\right) \right]  - \mathbb{E}_{\tilde{\pi}}\left[ f_\ast\left( \omega\left( \pi/\tilde{\pi} \right)\right) \right],
\end{aligned}
\label{eq:target2}
\end{equation}
where $\Xi = \mathcal{S} \times \mathcal{A}$ and $\mathcal{G}(\Xi) = \{\omega(\cdot): \Xi \to \mathbb{R}\}$ is the functional space on $\Xi$. It should be highlighted that the maximization operators in Eq. (\ref{eq:targe_f}) and Eq. (\ref{eq:target2}) have entirely different meanings: in Eq. (\ref{eq:targe_f}) the $\omega$ is taken over a single variable, and over all possible function $\omega(\cdot) \in \mathcal{G}(\Xi)$ in Eq. (\ref{eq:target2}). By the optimality condition of the inner optimization problem \cite{algaedice}, for all $\pi \in \prod$, the value of optimal $\omega^{\ast}_{\pi / \tilde{\pi}}$ in Eq. (\ref{eq:target2}) satisfies that
\begin{equation}
    \pi = \tilde{\pi} \cdot f_\ast'\left(\omega^{\ast}_{\pi / \tilde{\pi}}\right),
\end{equation}
where $\omega^{\ast}_{\pi / \tilde{\pi}}$ is the shorthand for $\omega^\ast\left( \frac{\pi}{\tilde{\pi}} \right)$. With the fact that the derivative of a convex function $f'$ is the inverse function of the derivative of its Fenchel conjugate function $f_\ast'$ \cite{rockafellar2015convex}, we have the following relationship between $\omega^\ast(\cdot)$ and $f'$,
\begin{equation}
    \omega^\ast_{\pi / \tilde{\pi}} = f'\left( \pi/\tilde{\pi}\right).
\label{eq:optimal}
\end{equation}
%{\color{red}(the connection between optimal and solution. May be here need a proof)} 
To solve the Eq. (\ref{eq:target2}), we make a further assumption, the motivation of which is explained in Appendix C: 
\begin{asm}
We solve the inner optimization problem in Eq. (\ref{eq:target2}), \textit{i.e.},  $\omega(\cdot): \Xi \to \mathbb{R}$, in a constrained functional space $\Omega$: $\omega_{\pi / \tilde{\pi}}  = Q^\pi(s,a) - Q^{\tilde{\pi}}(s,a)$.
\label{asm:4}
\end{asm}

Then, we can minimize the absolute value of difference between value function $Q^\pi(s,a)$ and $Q^{\tilde{\pi}}(s,a)$. Under the Assumption \ref{asm:4}, the saddle-point optimization in Eq. (\ref{eq:target2}) can be reformulated as
\begin{equation}
\begin{aligned}
    \mathcal{L}_f (\pi^\ast, Q^\ast)
    % = & \min_{\pi \in \Pi}\mathbb{D}_f(\pi\|\tilde{\pi}) \\
    = \min_{\pi \in \Pi} \max_{Q^\pi \in \Omega}  \mathbb{E}_\pi\left[ Q^\pi(s,a) - Q^{\tilde{\pi}}(s,a)\right] \\
    - \mathbb{E}_{\tilde{\pi}}\left[ f_\ast\left( Q^\pi(s,a) - Q^{\tilde{\pi}}(s,a)\right) \right],
\end{aligned}
\label{eq:target3}
\end{equation}
The $Q^{\tilde{\pi}}(s,a)$ is the $Q$ value of the sampling policy. Besides, $Q^\pi(s,a)$ and $Q^{\tilde{\pi}}(s,a)$ satisfy the Bellman equation,
\begin{equation}
    \begin{aligned}
        Q^\pi(s,a) = & r(s,a) + \gamma \mathbb{E}_{\pi}[Q^\pi(s',a')] \\
        Q^{\tilde{\pi}}(s,a) = & r(s,a) + \gamma \mathbb{E}_{\tilde{\pi}}[Q^{\tilde{\pi}}(s',a')] \\
        = & r(s,a) + \gamma \mathbb{E}_{\tilde{\pi}}[r(s',a')] + \gamma^2\mathbb{E}_{\tilde{\pi}}[r(s'',a'')] + \cdots\! .
    \end{aligned}
\label{eq:q_relation}
\end{equation}
% The $Q^\pi(s,a) - Q^{\tilde{\pi}}(s,a)$ is simplified as $x_{Q^\pi, Q^{\tilde{\pi}}}(s,a)$, 
% and the object function in Eq. (\ref{eq:target3}) can be rewritten as
% \begin{equation}
% \begin{aligned}
%     \mathcal{L}_f (\pi^\ast, Q^\ast)
%     % = & \min_{\pi \in \Pi}\mathbb{D}_f(\pi\|\tilde{\pi}) \\
%      & = \min_{\pi \in \Pi} \max_{Q^\pi \in \Omega}
%      \mathbb{E}_\pi\left[ Q^\pi(s,a) - Q^{\tilde{\pi}}(s,a)\right] \\
%     & \qquad - \mathbb{E}_{\tilde{\pi}}\left[ f_\ast\left( Q^\pi(s,a) - Q^{\tilde{\pi}}(s,a)\right) \right].
% \end{aligned}
% \label{eq:target4}
% \end{equation}

The FRL is a generic framework where different $f$ functions can be utilized and the forms of the object function in Eq. (\ref{eq:target3}) can vary along with the usage of different $f$ functions~\cite{nachum2020Duality}. In FRL, the optimization phase of the inner function corresponds to the policy evaluation phase, and the optimization phase of the outer function corresponds to the policy improvement (see in Leamm \ref{plicy_improvement} and Theorem \ref{theorem:5}). These two phases are performed simultaneously when solving the saddle-point optimization shown in Eq. (\ref{eq:target3}). In the subsequent subsection, we discuss an example of the FRL framework with a specific $f$ function, which can help readers better understand the FRL. 

\subsection{An example of FRL given a specific $f$ function}
In this subsection, our analysis can be divided into two steps: (1) we deduce the saddle-point optimization objective in special form with a certain $f$ function; and (2) we develop the policy evaluation and the policy improvement for solving this optimization problem.

The $f$-divergence needs satisfy two conditions: (1) $f(\cdot): \mathbb{R}^n \to \mathbb{R}$ is a convex, proper and semi-continuous function and (2) $f(1) = 0$. In this work, we take $f(x) = \frac{1}{2}(x - 1)^2$, which is a commonly utilized function in $f$-divergence research~\cite{dualdice,algaedice,nachum2020Duality}. According to the Eq. (\ref{eq:conjugate}), we have  $f_\ast(z) = \max_x x \cdot z - f(x)$. The conjugate function of $f(x)$ (\textit{i.e.}, $f_\ast(x)$) equals to $\frac{1}{2} x^2 + x$~\cite{boyd2004convex}.
Then, our objective function in Eq. (\ref{eq:target3}) is reformulated as a concrete form,
\begin{equation}
\begin{aligned}
    &\mathcal{L}_f (\pi^\ast, Q^\ast)
    % = & \min_{\pi \in \Pi}\mathbb{D}_f(\pi\|\tilde{\pi}) \\
    =  \min_{\pi \in \Pi} \max_{Q^\pi \in \Omega}  \mathbb{E}_\pi\left[ Q^\pi(s,a)- Q^{\tilde{\pi}}(s,a)\right] \\
   &\ -  \mathbb{E}_{\tilde{\pi}}\left[ \left(Q^\pi(s,a)- Q^{\tilde{\pi}}(s,a)\right)  + \frac{1}{2}\left( Q^\pi(s,a)- Q^{\tilde{\pi}}(s,a)\right)^2 \right].
\end{aligned}
\label{eq:target5}
\end{equation}
% With the help of Bellman equation of $Q^\pi(s,a)$ and $Q^{\tilde{\pi}}(s,a)$, we reduce the special form of saddle-point optimization problem and obtain the Theorem \ref{theorem:3}.
% \begin{theorem}
%     With the $f(x) = \frac{1}{2} (x-1)^2$, the FRL framework can be defined as a saddle point optimization problem by using convex conjugate,
%     \begin{equation}
%         \begin{aligned}
%             & \min_\pi \mathbb{D}_f(\pi \| \tilde{\pi}) = \min_\pi \max_{Q^\pi} \mathbb{E}_{\pi}\left[Q^\pi(s,a) - \mathbb{E}_{\tilde{\pi}}[R(s,a)] \right] \\
%             & \qquad - \frac{1}{2}\mathbb{E}_{\tilde{\pi}}\left[ \left(Q^\pi(s,a)-\mathbb{E}_{\tilde{\pi}}[R(s,a)] + 1\right)^2 - 1\right].
%         \end{aligned}
%         \label{eq:theorem3}
%     \end{equation}
% \label{theorem:3}
% \end{theorem}
% For the proof of Theorem \ref{theorem:3}, please refer to Appendix C. $R(s,a)$ is the discounted sum of rewards of a trajectory $\{s_0,a_0,r_0,s_1,a_1,r_1,\cdots\}$. Note that $R(s,a)$ satisfies different Bellman equations given different policy. Here we give two examples to illustrate. Firstly, for sampling policy (e.g., the greedy policy in DQN), i.e., the policy is induced by the trajectories, we have the equation that $\mathbb{E}_{\tilde{\pi}}[R(s,a)] = r(s,a) + \gamma \mathbb{E}_{\tilde{\pi}}[Q^{\tilde{\pi}}(s',a')]$ according to Bellman equation. Secondly, for learning policy, we have $\mathbb{E}_{\pi}[R(s,a)] = r(s,a) + \gamma \mathbb{E}_{\pi}[Q^{\pi}(s',a')]$. We elaborate on this viewpoint in Appendix C.

The policy evaluation and policy improvement are indispensable to solve the saddle-point optimization problem. It is noted that from Eq. (\ref{eq:target5}), we treat the inner max operation over $Q^\pi$ as the optimization process for policy evaluation, and the outer min operation over $\pi$ as the optimization process for policy improvement. 
Lemma \ref{plicy_improvement} suggests that the learning policy can be improve by minimizing $f$-divergence. Now we explain the theorem to compute the policy gradient to solve the saddle point optimization that is transformed from the objective of minimizing $f$-divergence. 

\begin{theorem}
    (Fenchel Policy Gradient Theorem) If the dual function $Q^\pi$ is fixed, the gradient of FRL objective function $\mathcal{L}_f (\pi, Q^\pi)$ with respect to $\pi$ is,
    \begin{equation}
        \begin{aligned}
            & \frac{\partial }{\partial \pi} \mathcal{L}_f (\pi, Q^\pi) = \\
            & \quad \mathbb{E}_{\tau \sim \pi}\left[ \left( Q^{\pi}(s,a) - Q^{\tilde{\pi}}(s,a)\right) \nabla \log \pi(a|s)\right]. \\
        \end{aligned}
    \end{equation}
    where $Q^\pi(s,a)$ and $Q^{\tilde{\pi}}(s,a)$ are the $Q$ function of learning policy $\pi$ and sampling policy $\tilde{\pi}$ respectively. 
\label{theorem:4} 
\end{theorem}
For the proof of Theorem \ref{theorem:4}, please refer to Appendix D. The optimization for the inner max over $Q^\pi$, can be viewed as the process of policy evaluation, and we have the following theorem.

\begin{theorem}
    (Fenchel Policy Evaluation) When the outer optimization object is fixed, the gradient of the $Q$ function of learning policy is defined as,
    \begin{equation}
        \begin{aligned}
            & \frac{\partial}{\partial Q^\pi} \mathcal{L}_f (\pi, Q^\pi) = \\
            &  \qquad - \mathbb{E}_{\tilde{\pi}}\left[ \left( Q^\pi(s,a) - Q^{\tilde{\pi}}(s,a) \right)\nabla Q^\pi(s,a) \right] \\
            & \qquad + \mathbb{E}_{\pi} \left[\nabla Q^\pi(s,a)\right] - \mathbb{E}_{\tilde{\pi}} \left[\nabla Q^\pi(s,a)\right].
        \end{aligned}
        \label{eq: policy evaluation in FRL}
    \end{equation}
\label{theorem:5}
\end{theorem}
For the proof of Theorem \ref{theorem:5}, we also refer the readers to Appendix D. The approximation of the state-action function of sampling policy $Q^{\tilde{\pi}}(s,a)$ is realized by minimizing the Bellman error \cite{sutton_reinforcement_2018},
\begin{equation}
    J(Q^{\tilde{\pi}}) = \frac{1}{2} \left( Q^{\tilde{\pi}}(s,a) - \left(r(s,a) + \gamma \mathbb{E}_{\tilde{\pi}} \left[ Q^{\tilde{\pi}}(s',a') \right]\right)\right)^2.
\end{equation}
The $Q^{\tilde{\pi}}(s,a)$ is also approximated by neural networks. The above analysis shows that the policy evaluation and the policy improvement phases can gradually minimize the $f-$divergence between learning policy and sampling policy, so the optimal learning policy can be obtained. One may also extend other proper, convex, lower semi-continuous $f$ functions beyond the quadratic function.  Algorithm \ref{alg:training process} illustrates the process of training agents in the FRL framework.
\begin{algorithm*}[!t]
\caption{$f-$Divergence Reinforcement Learning (FRL) with $f(x)=\frac{1}{2}(x-1)^2$}
\label{alg:training process}
\textbf{Requires}: \\
$\theta$: initialize policy network parameters;
$\phi$: initialize $Q^\pi$ network; $\tilde{\phi}$: initialize $Q^{\tilde{\pi}}$ network;
    $\epsilon_Q$: learning rate of $Q$ network; $\epsilon_P$: learning rate of policy ; 
    $K$: dimension of action; $T$: the counter; $B$: batch size; the sampling policy is induced by the learning policy in a fixed way, in our realization, the sampling policy is defined as: $\tilde{\pi} = \epsilon-\text{greedy}(Q^{\pi}_\phi)$ \\
    \Repeat{$T > T_{\text{max}}$}{
        % \# Sample from learning policy and sampling policy \\
        Sample actions $a^{(i)} \sim \pi_{\theta}(s^{(i)})$, for $i=1,\cdots,B$, get $\{s^{(i)},a^{(i)},s^{'(i)},r^{(i)}\}$\\
        Sample actions $\tilde{a}^{(i)} \sim \tilde{\pi}(s^{(i)})$, for $i=1,\cdots,B$, get $\{s^{(i)},\tilde{a}^{(i)},\tilde{s}^{'(i)},\tilde{r}^{(i)}\}$\\
        \# Calculate the gradient of policy  and $Q$ network \\
        $\nabla_\theta \mathcal{L}_f(\pi_\theta,Q^\pi_\phi) = \frac{1}{B}\sum_{i=1}^B(Q^\pi_\phi(s^{(i)},a^{(i)}) - \tilde{r}^{(i)} - \sum_{k=1}^K Q_\phi^\pi(s^{'(i)},a_k) \pi(a_k|s^{'(i)})) \nabla_\theta \log \pi_\theta(a^{(i)}|s^{(i)}) $\\
        $\nabla_\phi \mathcal{L}_f(\pi_\theta,Q^\pi_\phi) = -\frac{1}{B}\sum_{i=1}^B ( (Q^\pi_\phi(s^{(i)},\tilde{a}^{(i)}) - \tilde{r}^{(i)} - \sum_{k=1}^K Q_\phi^\pi(\tilde{s}^{'(i)},a_k) \pi(a_k|\tilde{s}^{'(i)}))\nabla_\phi Q^\pi_\phi(s^{(i)},\tilde{a}^{(i)})) + \frac{1}{B} \sum_{i=1}^B( \nabla_\phi Q^\pi_\phi(s^{(i)},\tilde{a}^{(i)}) - \nabla_\phi Q^\pi_\phi(s^{(i)},a^{(i)})) $\\
         $ \nabla_{\tilde{\phi}} Q^{\tilde{\pi}}_{\tilde{\phi}} = \frac{1}{B}\sum_{i=1}^B ( (\nabla_{\tilde{\phi}} Q^{\tilde{\pi}}_{\tilde{\phi}}(s^{(i)},\tilde{a}^{(i)}) - \tilde{r}^{(i)} - \sum_{k=1}^K \nabla_{\tilde{\phi}} Q^{\tilde{\pi}}_{\tilde{\phi}}(\tilde{s}^{'(i)},a_k) \pi(a_k|\tilde{s}^{'(i)}))\nabla_{\tilde{\phi}} Q^{\tilde{\pi}}_{\tilde{\phi}}(s^{(i)},\tilde{a}^{(i)})) $\\
        \# Update network parameters \\
        $\theta \gets \theta + \epsilon_P \nabla_\theta \mathcal{L}_f(\pi_\theta,Q^\pi_\phi) $ \\ 
        $\phi \gets \phi + \epsilon_Q \nabla_\phi \mathcal{L}_f(\pi_\theta,Q^\pi_\phi)$ \\ 
        $\tilde{\phi} \gets \tilde{\phi} + \epsilon_Q \nabla_{\tilde{\phi}} Q^{\tilde{\pi}}_{\tilde{\phi}}$ \\ 
        $T \leftarrow T + 1$ \\
    }
\end{algorithm*}

\subsection{Alleviating the overestimation issue}
In the conventional DRL methods, the overestimation arises due to the following process:
\begin{equation}
	\begin{aligned}
	\max_ a Q(s,a) &= \max_a \mathbb{E}_U [Q(s,a) + U(a)] \\
	&\leq \mathbb{E}_U [\max_a \{ Q(s,a) + U(a)\}],
	\end{aligned}
\end{equation}
where $U(a)$ is action noise and the inequality holds due to Jensen's inequality \cite{kuznetsov2020controlling}. Actually, in FRL framework, we can rethink that the overestimation issue is somehow caused by only using sampling policy in the evaluation phase. In our framework, we utilize both sampling policy and learning policy in the policy evaluation phase. The introduction of learning policy in the FRL framework creates a balance in the policy evaluation phase, \textit{i.e.}, alleviate the overestimation issue induced by Jensen's inequality. This dynamics process is directly reflected by Eq. (\ref{eq: policy evaluation in FRL}), where the additional two gradient terms balance the estimation issue in conventional policy evaluation objectives.

% {\color{red}In traditional DRL methods, the learning process of value function usually considered as minimizing the Bellman error under the sampling policy, which causes the value function overestimation easily \cite{dqn,a2c,ppo}. The traditional DRL framework is presented as,
% \begin{equation}
%     \max_{\pi} \max_{Q^\pi} \left[ \mathbb{E}_{\pi}\left[R(s,a)\right] - \frac{1}{2} \mathbb{E}_{\tilde{\pi}} \left[\left(Q^\pi(s,a) - R(s,a)\right)^2\right] \right] 
% \end{equation}
% Compared with traditional DRL framework, the FRL framework add the optimization for the difference of value function obtained by learning policy and sampling, which avoid the update of $Q$ function only depends on sampling policy and alleviate the value function overestimation.}

\section{Experimental Results}
We conduct experiments on Atari 2600 games \cite{atari}, which is commonly adopted for evaluating DRL algorithms. To demonstrate the effectiveness of the FRL framework, we compare it with three well-known baseline algorithms: Advantage Actor Critic (A2C), Proximal Policy Optimization (PPO) and Soft Actor Critic (SAC). We also show that the issue of overestimating value function can be alleviated in the FRL framework.

\subsection{Evaluation on Atari 2600}
To evaluate the FRL framework, we compare our framework on Atari 2600 games with three popular DRL algorithms. The details of baselines are listed as follows.
\begin{itemize}
    \item \textbf{A2C} is synchronous advantage actor critic approach, which is an efficient on-policy DRL algorithm;
    \item \textbf{PPO} uses a clipped surrogate objective to approximately guarantee that the policy is monotonically raised each time, improving training stability;
    \item \textbf{SAC} is trained to maximize expected cumulative rewards, which encourages exploration by adding policy entropy to the optimization goal.
\end{itemize}
We utilize the \texttt{Stable-Baselines3}\footnote{https://github.com/DLR-RM/stable-baselines3} implementations for these three algorithms. The FRL framework are implemented based on the \texttt{Stable-Baselines3} package as well. To make the comparison fair, we use the same hyper-parameters and architectures of policy and value function networks as for all the algorithms investigated in the paper.
For each algorithm, we take 10 million steps for training an agent. We train five agents simultaneously for each seed, and each agent utilizes eight-step returns for the policy gradient estimation. To reduce the effects of environmental uncertainty, we run each algorithm over five different random seeds and use the average numbers as the final result.

%\footnote{https://github.com/DLR-RM/stable-baselines3}

\begin{figure*}
\centering
\subfigure{
\includegraphics[width=1.8 in]{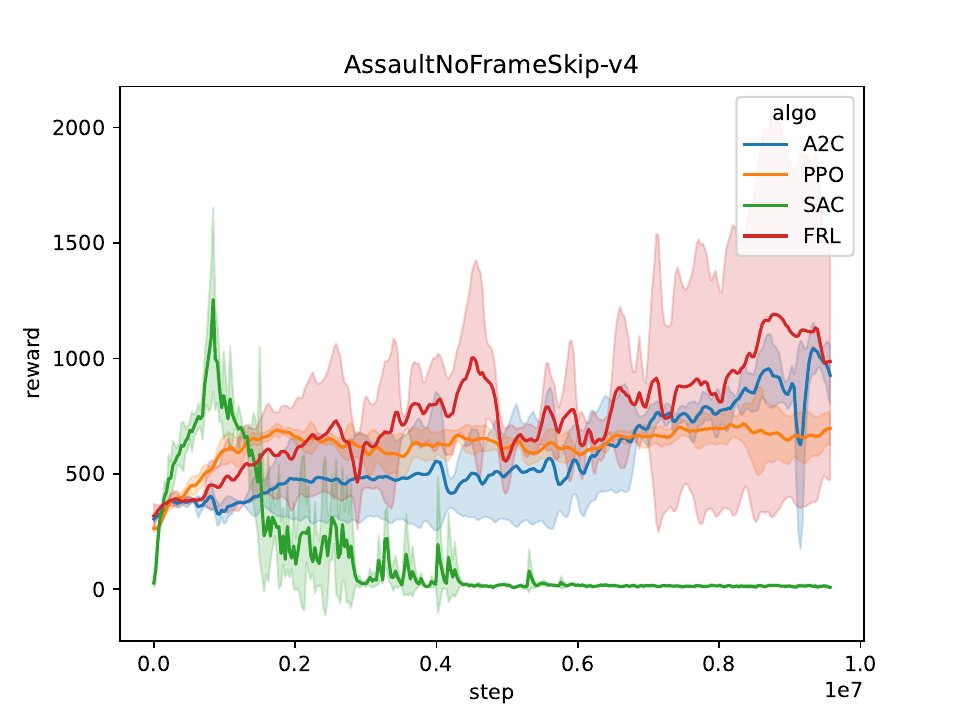}
}
\hspace{-0.7 cm}
\subfigure{
\includegraphics[width=1.8 in]{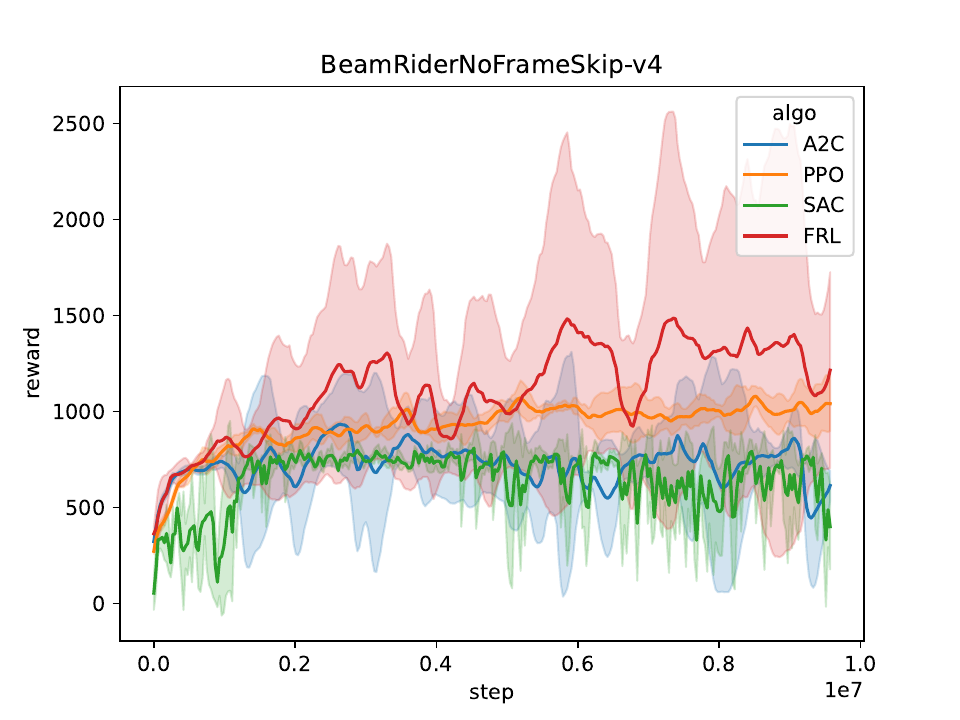}
}
\hspace{-0.7 cm}
\subfigure{
\includegraphics[width=1.8 in]{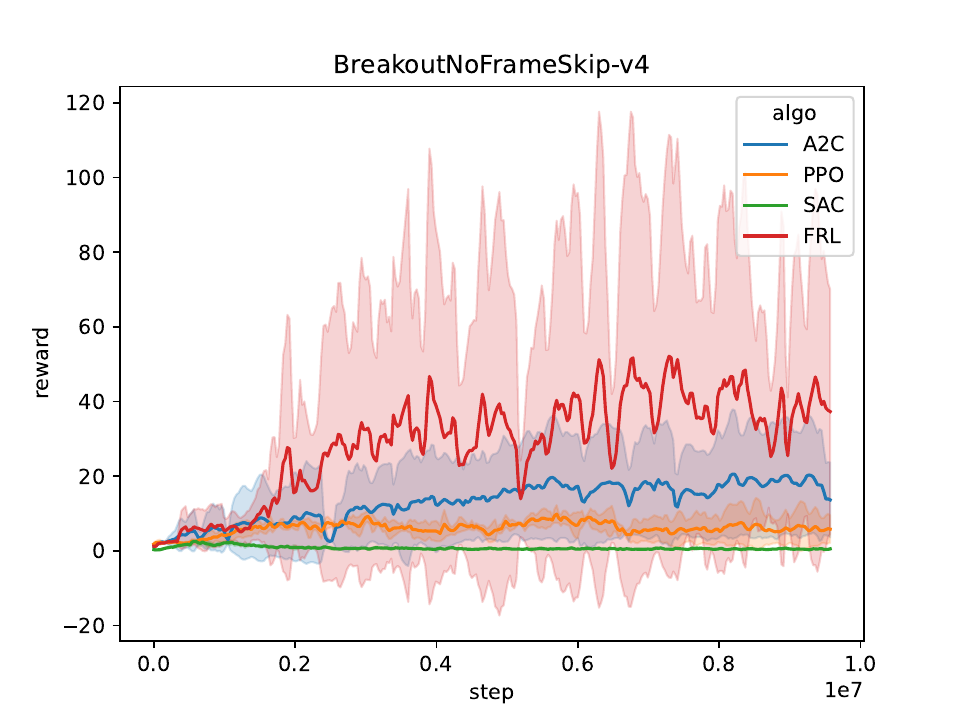}
}
\hspace{-0.7 cm}
\subfigure{
\includegraphics[width=1.8 in]{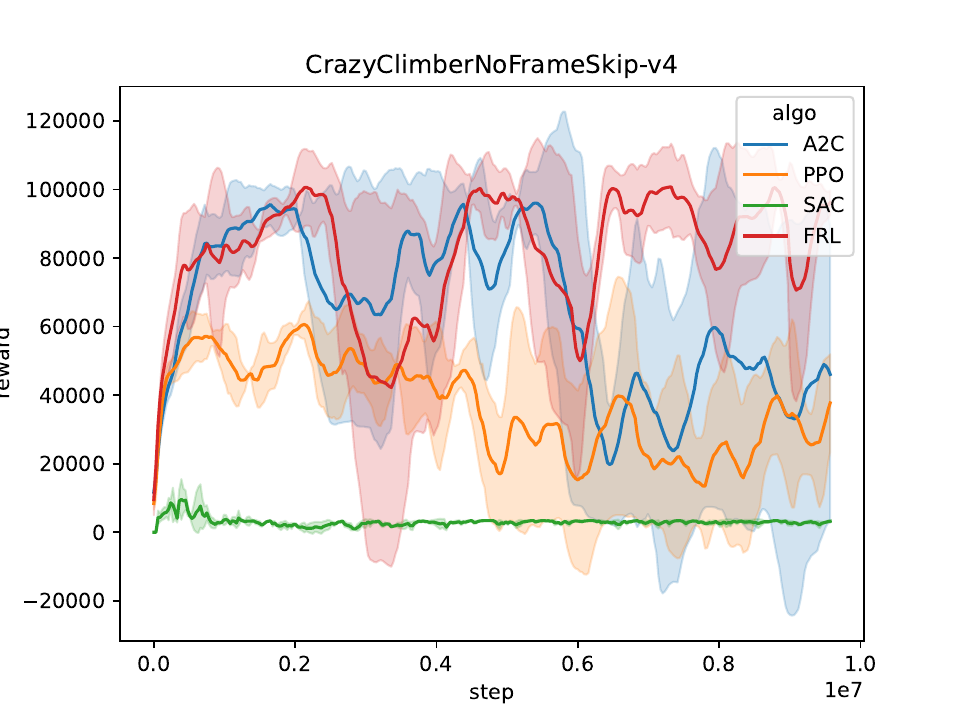}
}
\vspace{-0.3 cm}

\subfigure{
\includegraphics[width=1.8 in]{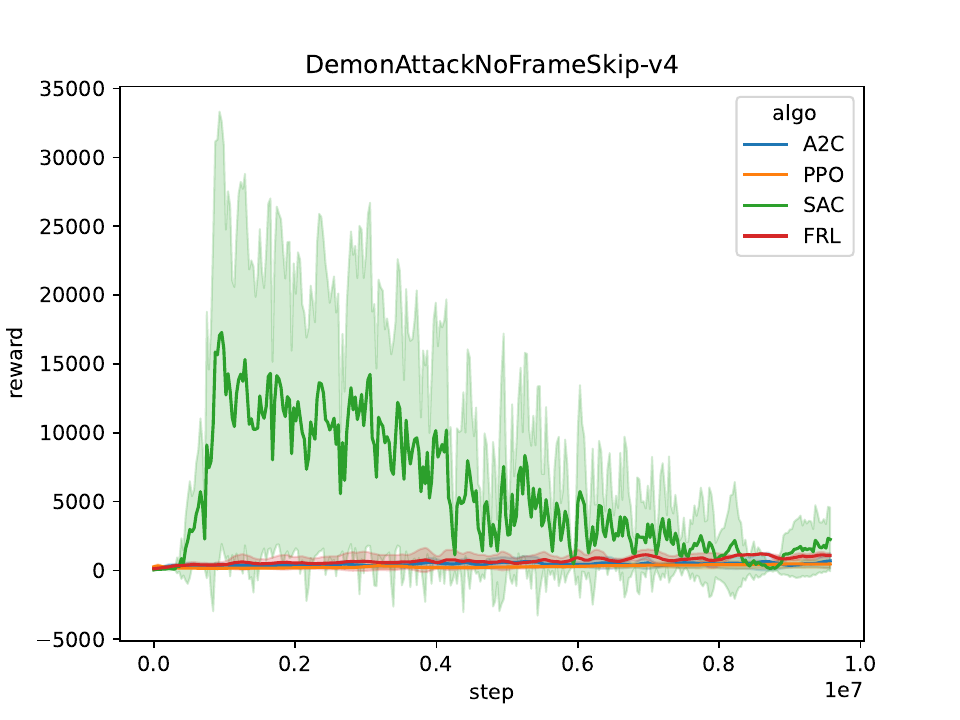}
}
\hspace{-0.7 cm}
\subfigure{
\includegraphics[width=1.8 in]{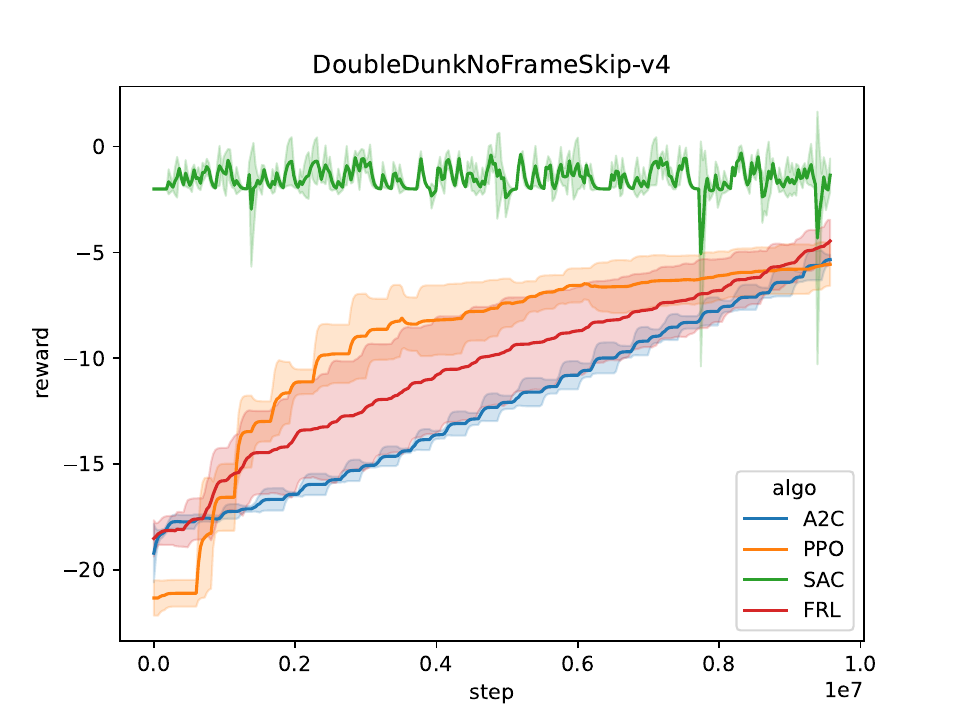}
}
\hspace{-0.7 cm}
\subfigure{
\includegraphics[width=1.8 in]{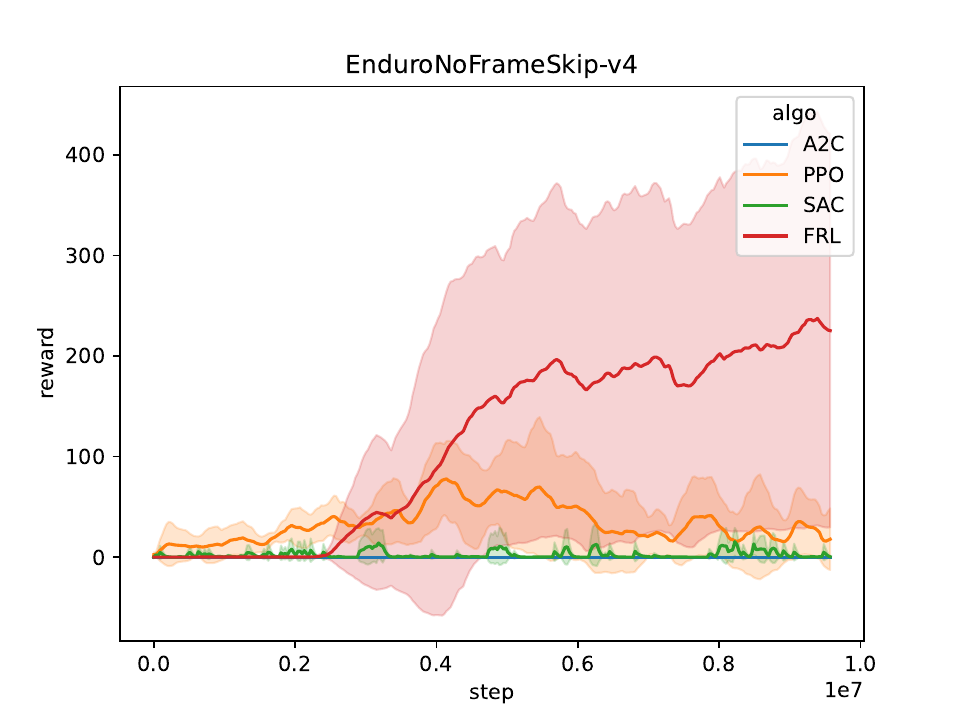}
}
\hspace{-0.7 cm}
\subfigure{
\includegraphics[width=1.8 in]{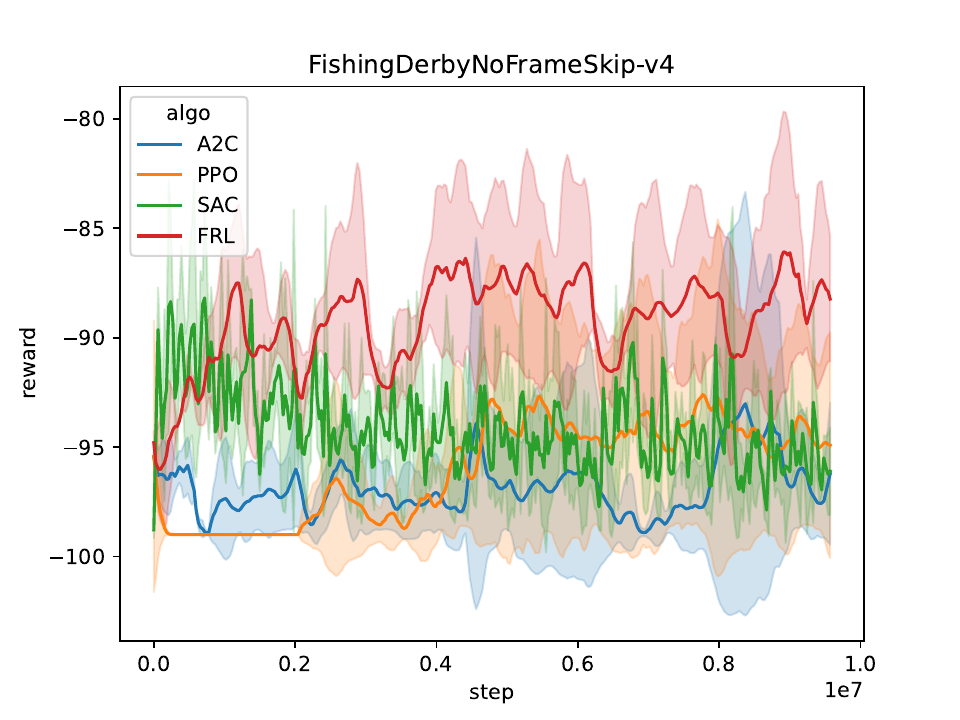}
}
\vspace{-0.3 cm}

\subfigure{
\includegraphics[width=1.8 in]{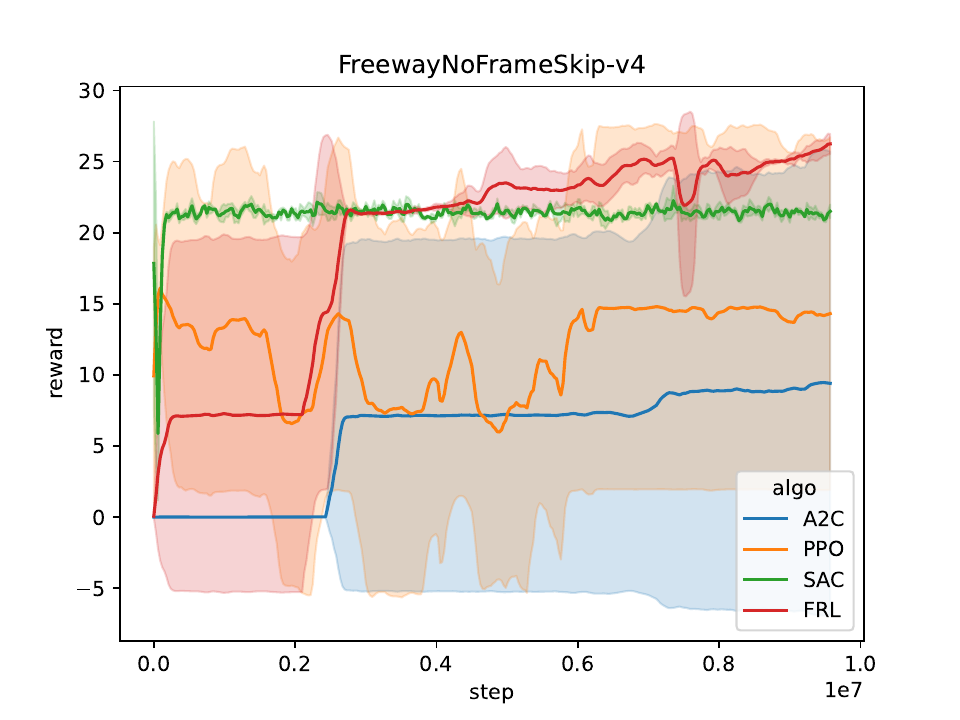}
}
\hspace{-0.7 cm}
\subfigure{
\includegraphics[width=1.8 in]{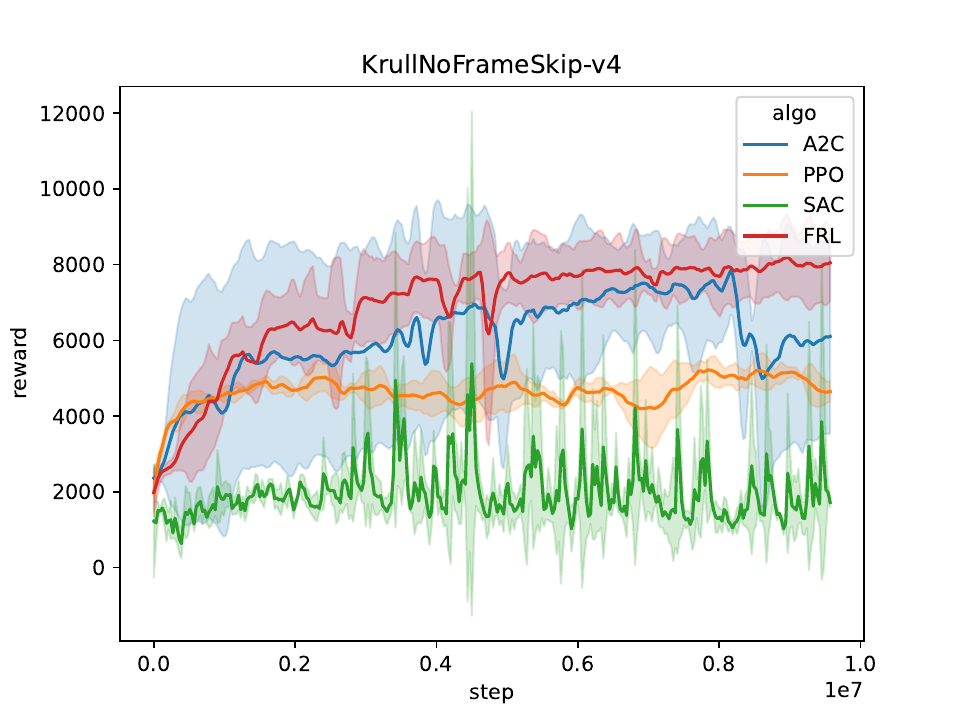}
}
\hspace{-0.7 cm}
\subfigure{
\includegraphics[width=1.8 in]{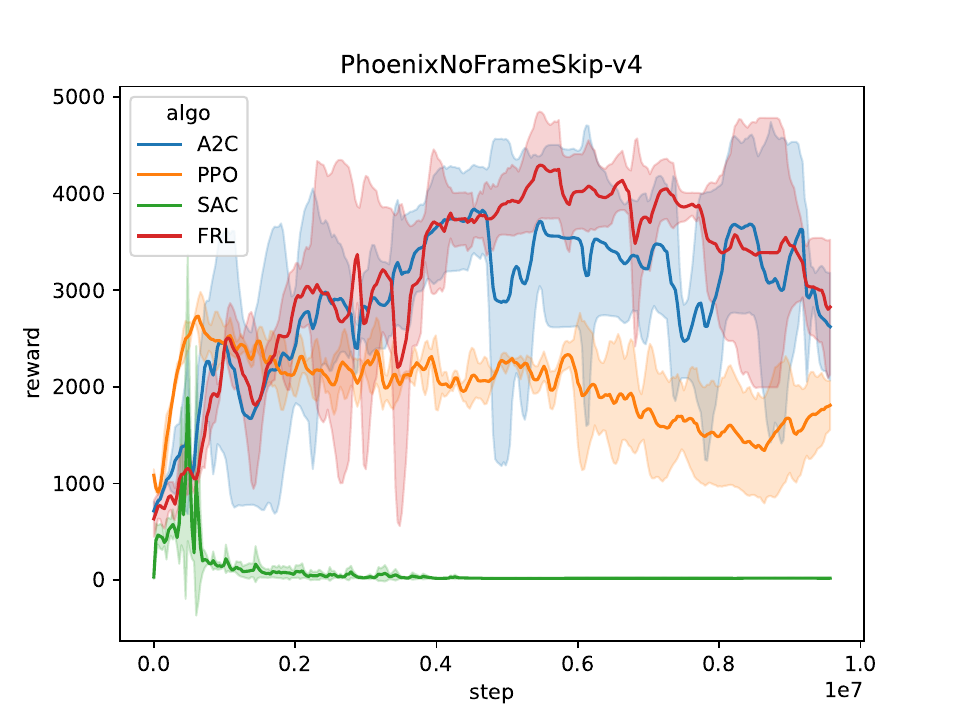}
}
\hspace{-0.7 cm}
\subfigure{
\includegraphics[width=1.8 in]{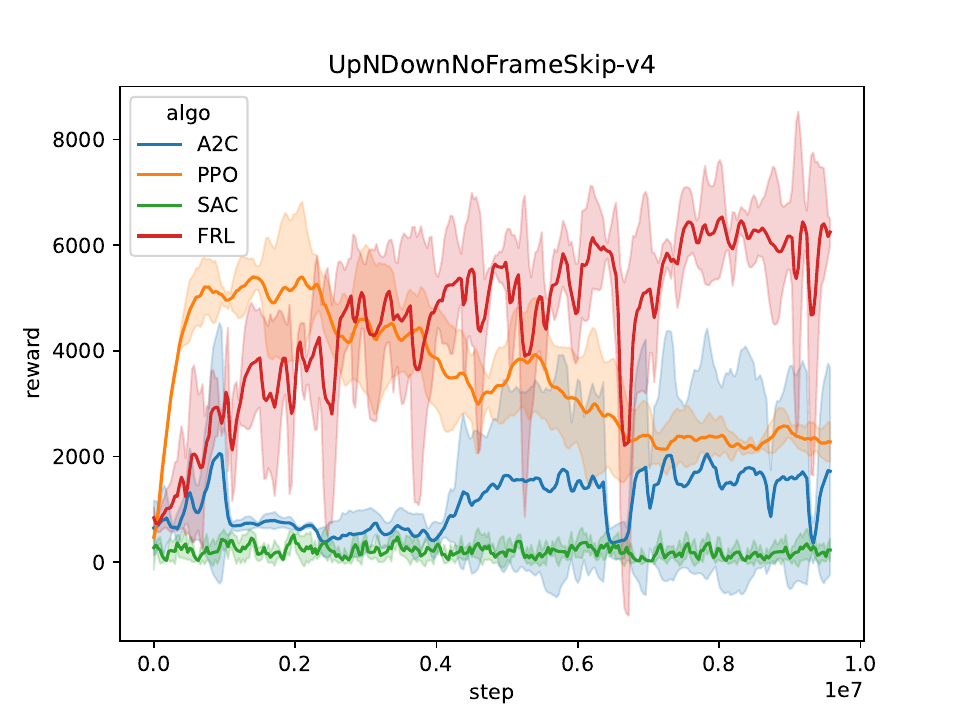}
}

\caption{\textbf{Performance curves. }Performance curves with FRL, A2C, PPO and SAC for the 12 Atari 2600 video games. Each curve are smoothed by the exponential moving average to with 3 step. The solid line is the mean and the shaded region represents a standard deviation of the average evaluation over five seeds.} 
\label{fig:rewards}
\end{figure*}

Due to the limited space, we only presenet the results on 12 Atari video games (NoFrameskip-v4) in Figure \ref{fig:rewards}. The full results of all the 57 Atrai 2600 games as well as the implementation details are shown in Appendix E. 
As illustrated in the Figure \ref{fig:rewards}, on the 10 (out of 12) games, the agents in the FRL framework outperform agents trained with the baseline algorithms. Out of all the 57 games investigated, the agents in our framework surpass other algorithms in 23 environments; it highlights the effectiveness of defining DRL from the viewpoint of minimizing the $f-$divergence between learning policy and sampling policy.

\subsection{Alleviating the Overestimation of Value Function}
\begin{figure*}
\centering
\subfigure[BeamRider]{
\includegraphics[width=1.8 in]{ 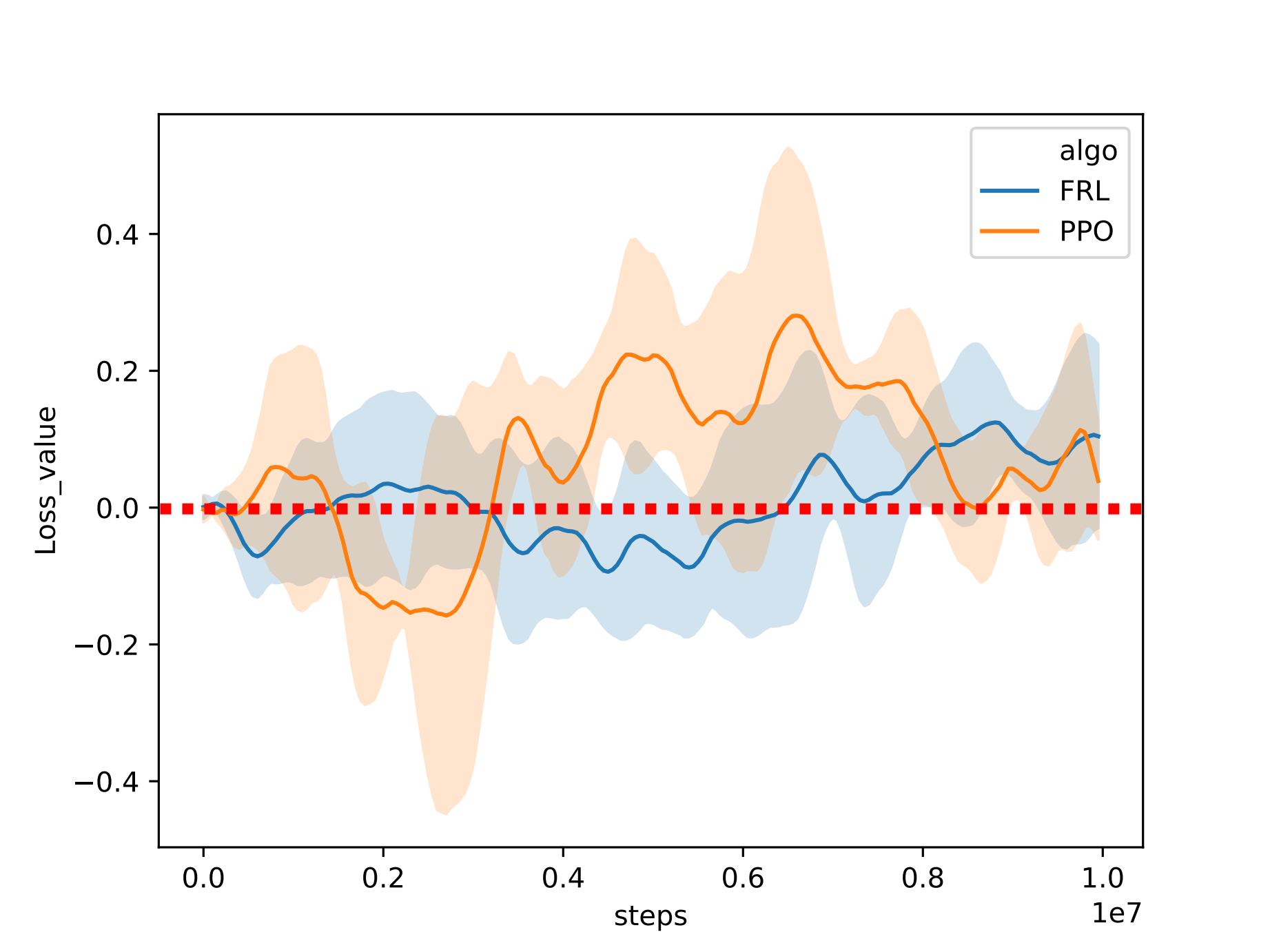}
}
\hspace{-0.7 cm}
\subfigure[Breakout]{
\includegraphics[width=1.8 in]{ 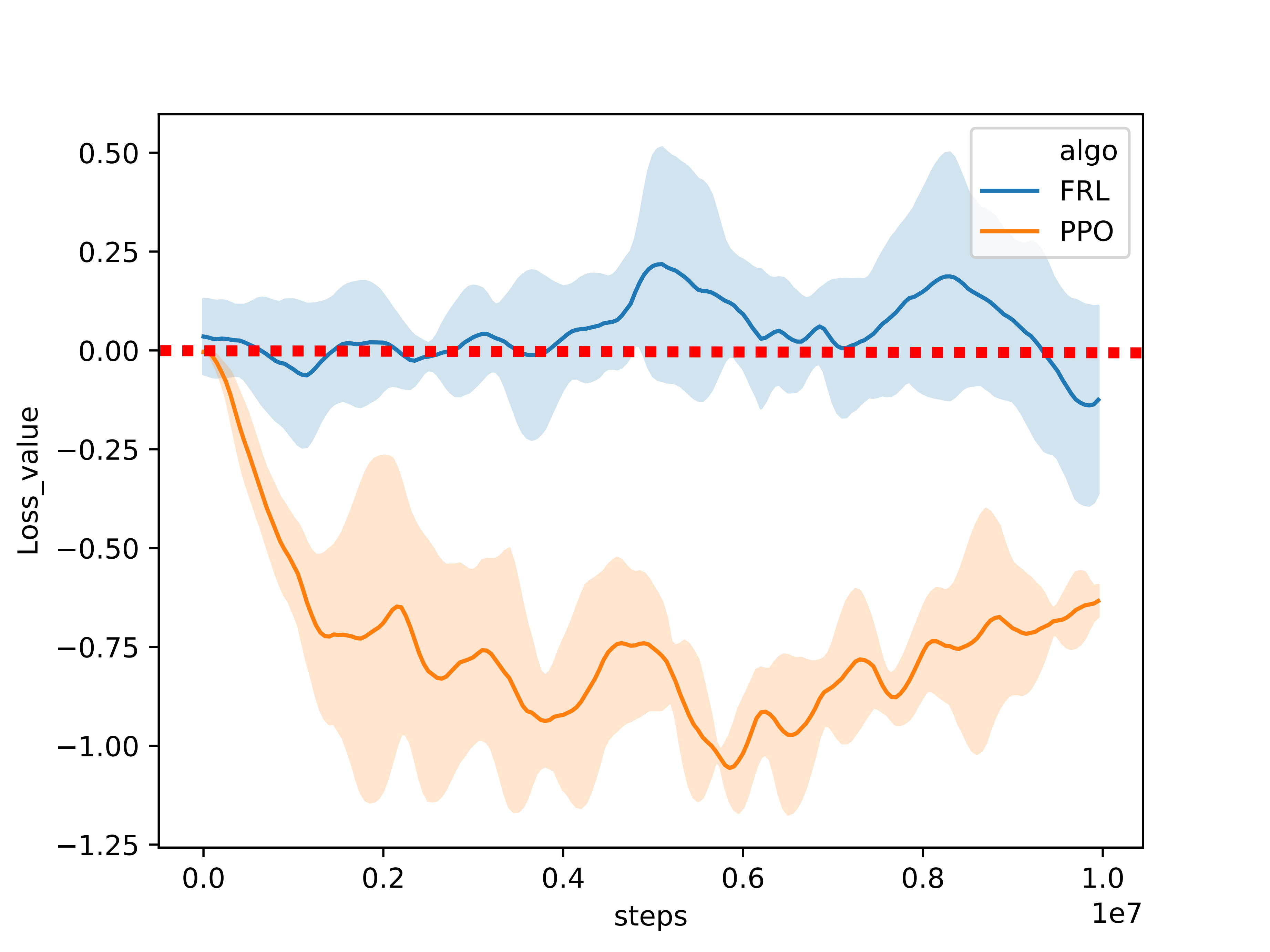}
}
\hspace{-0.7 cm}
\subfigure[DoubleDunk]{
\includegraphics[width=1.8 in]{ 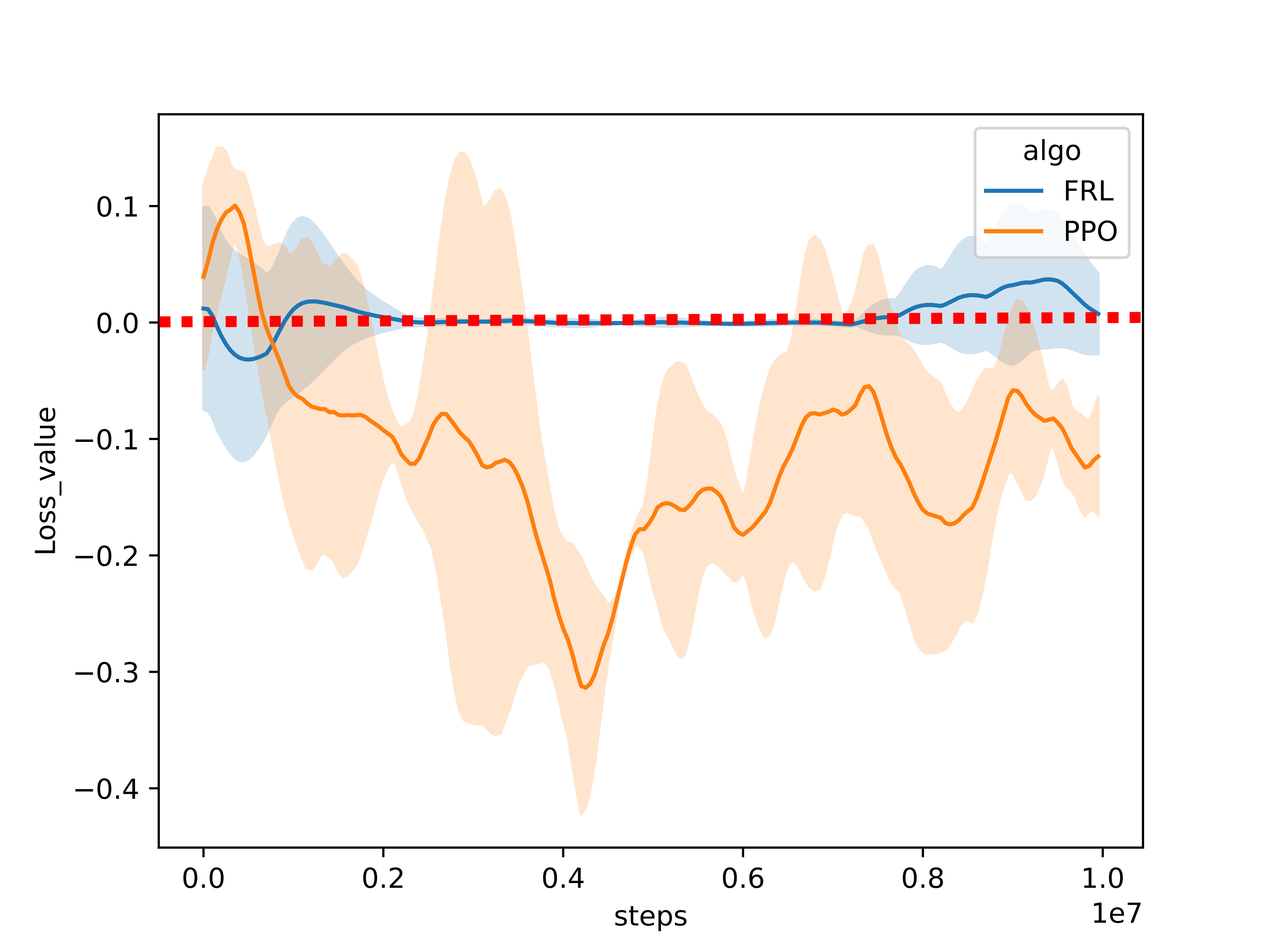}
}
\hspace{-0.7 cm}
\subfigure[Enduro]{
\includegraphics[width=1.8 in]{ 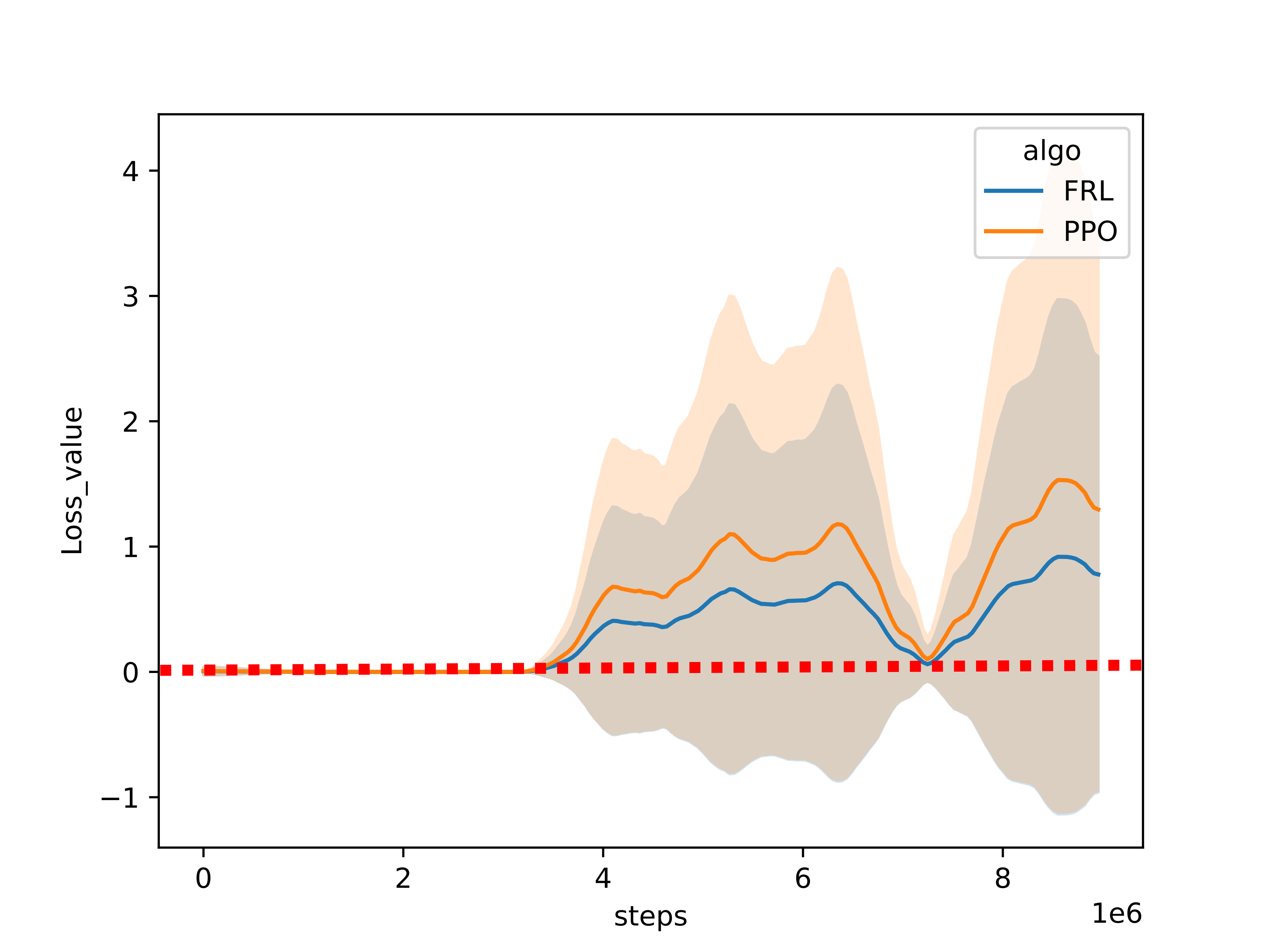}
}
\vspace{-0.3 cm}

\subfigure[FishingDerby]{
\includegraphics[width=1.8 in]{ 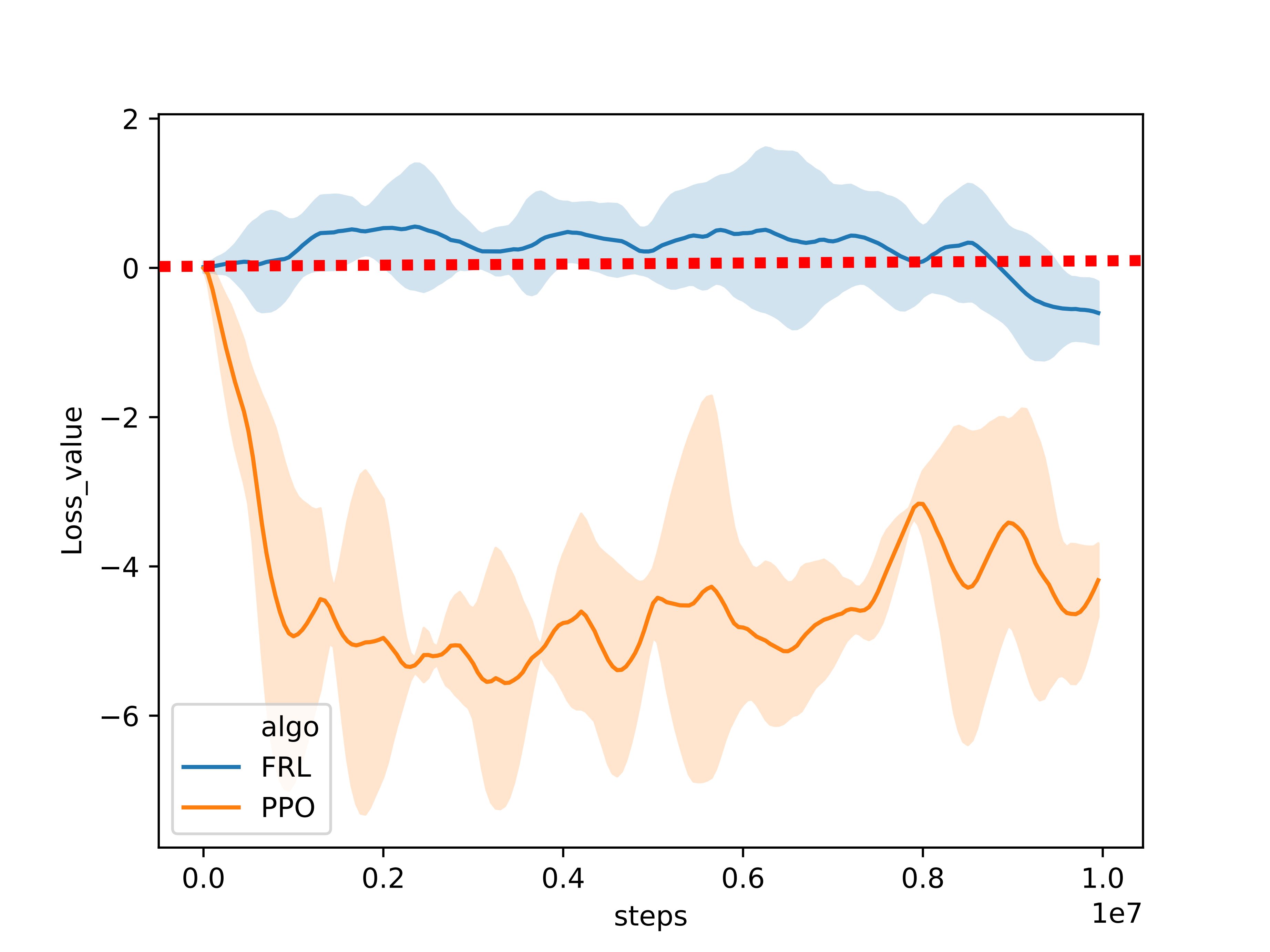}
}
\hspace{-0.7 cm}
\subfigure[Freeway]{
\includegraphics[width=1.8 in]{ 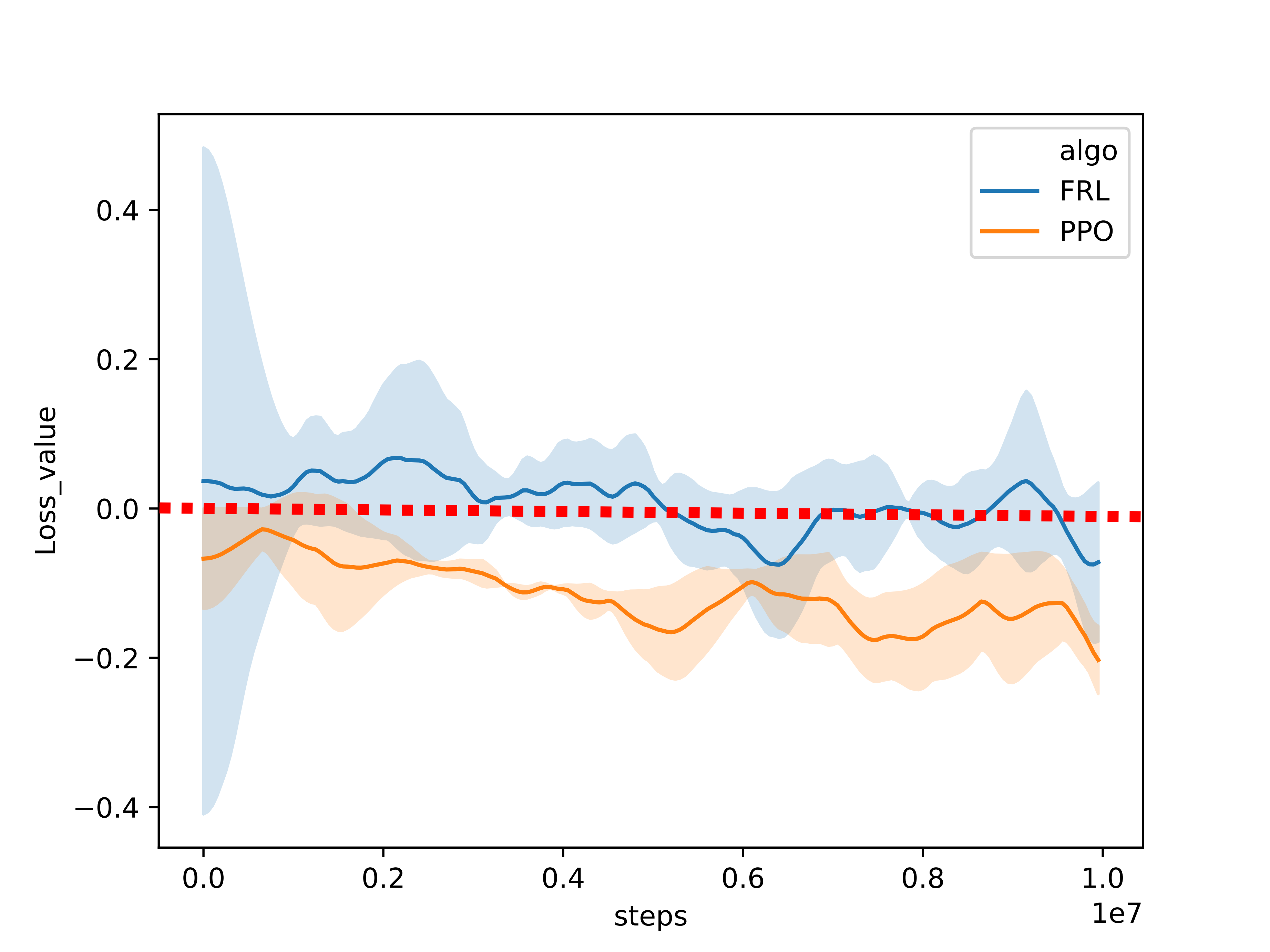}
}
\hspace{-0.7 cm}
\subfigure[Krull]{
\includegraphics[width=1.8 in]{ 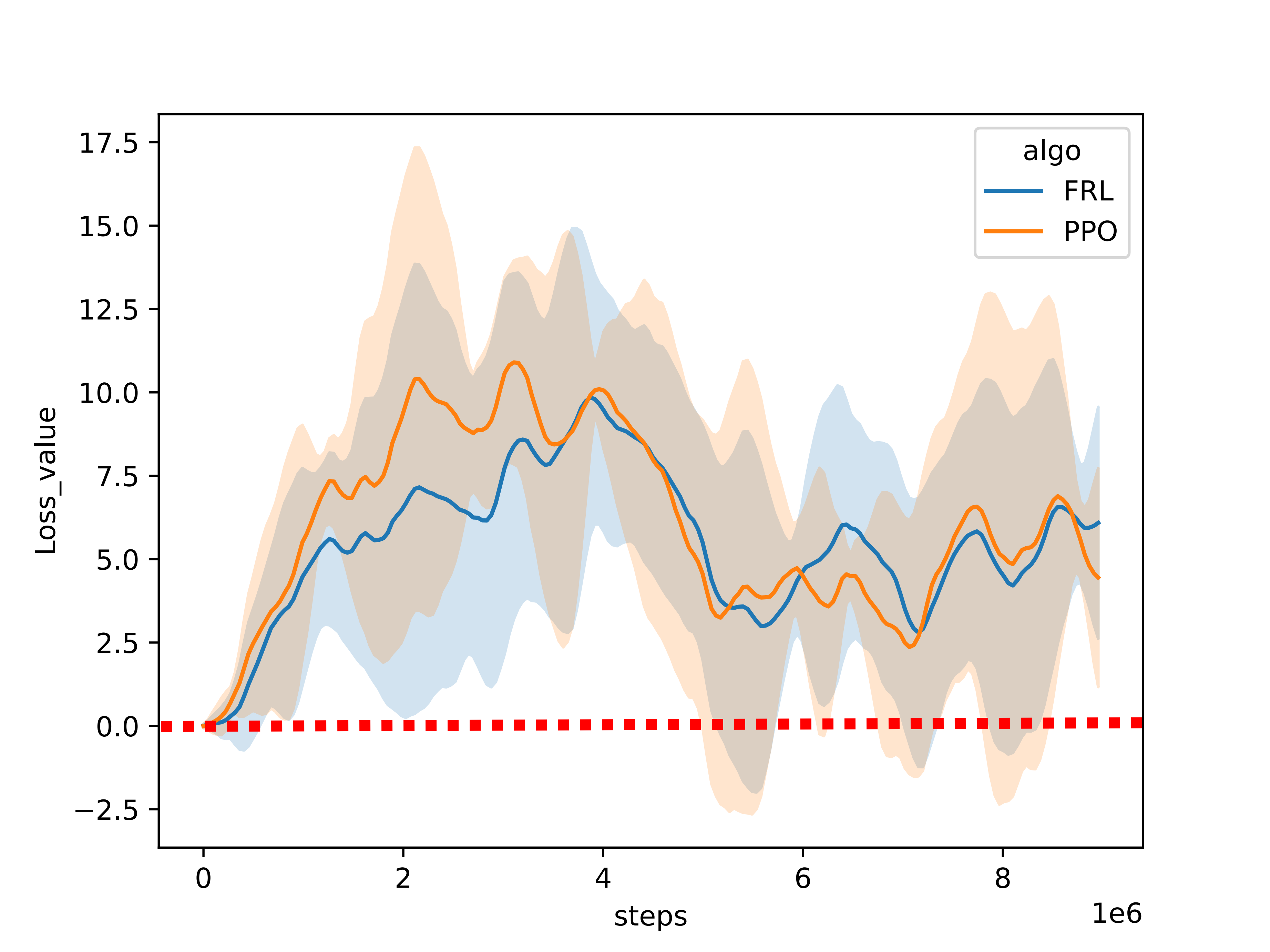}
}
\hspace{-0.7 cm}
\subfigure[Phoenix]{
\includegraphics[width=1.8 in]{ 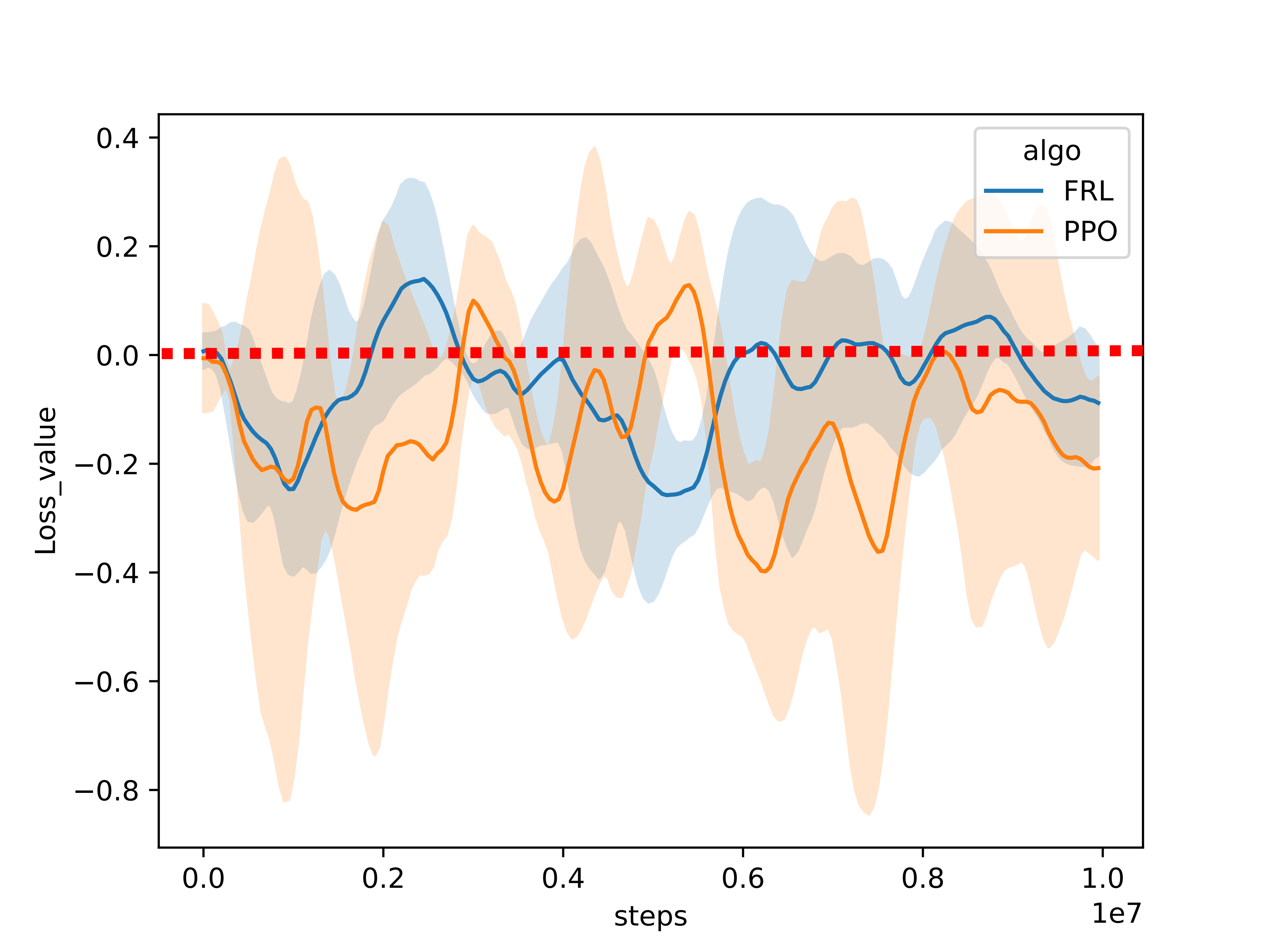}
}

\caption{\textbf{Loss curves. }Loss curves between true value and approximated value with FRL and PPO for the 8 Atari 2600 video games. Each curve are smoothed by the exponential moving average to with 3 step. The solid line is the mean and the shaded region represents a standard deviation of the average evaluation over five seeds. The red dotted line present $\text{Loss}_{\text{value}}=0$.} 
\label{fig:q_over_complete}
\end{figure*}

We compare the $\text{Loss}_{\text{value}}$ between the true value and the estimated value, where $\text{Loss}_{\text{value}}$ measures the severity of overestimation of the value function. The $\text{Loss}_{\text{value}}$ is defined as
\begin{equation}
    \text{Loss}_{\text{value}} = \mathbb{E}_\pi\left[Q^\pi(s,a)\right] - \text{Value}_{ \text{true}}, 
    \label{eq:loss_value}
\end{equation}
where the $\text{Value}_{\text{true}}$ is the cumulative rewards for a state in a trajectory, and $\mathbb{E}_\pi\left[Q^\pi(s,a)\right]$ is the inner product of the output of our policy network and $Q$ network. Ideally, $\text{Loss}_{\text{value}}$ is close to zero. 
We plot the trends of $\text{Loss}_{\text{value}}$ in Figure \ref{fig:q_over_complete}, which shows that the $\text{Loss}_{\text{value}}$ of FRL (the green lines) is significantly closer to zero than PPO's (the orange lines). It highlights that the FRL framework can alleviate the problem of overestimating value function. We refer readers to Appendix F for more detailed results.

\section{Conclusions}
In this paper, we propose the \textit{$f-$divergence reinforcement learning (FRL)} framework, where the policy evaluation and policy improvement phases are performed by minimizing $f-$divergence between the learning policy and sampling policy, instead of maximizing the expected cumulative rewards as conventional DRL algorithms. We mathematically prove that FRL framework can obtain the optimal policy and the objective of minimizing $f$-divergence can be converted into a saddle-point optimization problem. We also prove that the FRL framework is compatible with any $f$ functions that are convex, proper and semi-continuous. Therefore, the FRL defines a general form of policy evaluation and policy improvement. 
Our experiment results show that FRL outperforms the baselines in most cases. We find that the polices obtained from FRL suffer less from the issue of overestimating value function.
Our approach of minimizing $f$-divergence and utilizing convex duality tools to analyze DRL provides a new paradigm for studying DRL theory. In the future, we plan to explore more kinds of $f$ function in FRL, and investigate the most appropriate $f$ function for specific tasks.

\clearpage
\bibliography{mybib}

\clearpage
\onecolumn
\section*{Supplementary Materials}
\subsection*{Appendix A \label{Append:A}}

\bigskip

\noindent \emph{under the optimal policy $\pi^\ast$, the Bellman equation is equivalent to the Bellman optimality equations. In other words, we have the following equation}
$$
    \begin{aligned}
        Q^\ast(s,a) = & r(s,a) + \gamma \mathbb{E}_{s' \sim P(s,a),a'\sim \pi^\ast(\cdot \mid s')} [Q^\ast(s',a')] \\
        = & r(s,a) + \gamma \mathbb{E}_{s' \sim P(s,a)}\left[\max_{a'\in \mathcal{A}}\left[Q^\ast(s',a')\right]\right].
    \end{aligned}
$$
\emph{The top right formula is the Bellman equation, and the bottom formula is the Bellman optimality equation when the learning policy is optimal. 
Intuitively speaking, if we regard the maximum operator as a deterministic policy, the sampling policy $\tilde{\pi}$ is equivalent to the learning policy $\pi$ when the latter is optimal. }

\bigskip

\noindent\textbf{\emph{Proof.}} From eq (\ref{eq: bellman equation in method section}), we have
$$
    \begin{aligned}
        Q^\ast(s,a) & = \max_{\pi \in \Pi} Q^\pi(s,a)\\
         & = r(s,a) + \max_{\pi \in \Pi}\gamma \mathbb{E}_{s' \sim P(s,a), a' \sim \pi(\cdot \mid s')} [Q^\pi(s',a')] \\
         & = r(s,a) + \max_{\pi \in \Pi}\gamma \mathbb{E}_{s' \sim P(s,a)} [V^\pi(s')] \\
        & \overset{\text{(a)}}{=} r(s,a) + \gamma \mathbb{E}_{s' \sim P(s,a)} V^\ast(s') \\ 
        & = r(s,a) + \gamma \mathbb{E}_{s' \sim P(s,a)}[\max_{a' \in \mathcal{A}}[Q^\ast(s',a')]],
    \end{aligned}
\label{eq:e2o}
$$

\noindent where the equation (a) can be reduced based on follow relationship. First, we show the conditioned on $M_0: (s_0, a_0, r_0, s_1)=\left(s, a, r, s^{\prime}\right)$. The maximum discounted value is not a function of $(s,a,r)$. Specifically,

$$
\max _{\pi \in \Pi} \mathbb{E}\left[\sum_{t=1}^{\infty} \gamma^{t} r\left(s_{t}, a_{t}\right) \mid \pi,M_0=\left(s, a, r, s^{\prime}\right)\right]=\gamma V^{\star}\left(s^{\prime}\right).
$$

\noindent For all $\pi \in \Pi$, define an ``offset" policy $\pi_{(s,a,r)}$. The ``offset" policy chooses actions on a  trajectory $\tau$ according
to the same distribution that $\pi$ chooses actions on the trajectory $(s,a,r,\tau)$; \emph{e.g.,} $\pi_{(s,a,r)}(a_0=a'|s_0=s')$ equals to the probability $\pi(a_1=a'|M_0 = (s,a,r,s'))$. By the Markov property, we have that
$$
    \begin{aligned}
        & \mathbb{E}\left[\sum_{t=1}^{\infty} \gamma^{t} r\left(s_{t}, a_{t}\right) \mid \pi,M_0=\left(s, a, r, s^{\prime}\right)\right] \\ 
        = &  \gamma \mathbb{E}\left[\sum_{t=0}^{\infty} \gamma^{t} r\left(s_{t}, a_{t}\right) \mid \pi_{(s, a, r)}, s_{0}=s^{\prime}\right] \\
        = &\gamma V^{\pi_{(s, a, r)}}\left(s^{\prime}\right).
    \end{aligned}
$$
Besides, because $V^\pi(s')$ is not a function about $(s,a,r)$, we have 
$$
\begin{aligned}
    & \max _{\pi \in \Pi} \mathbb{E}\left[\sum_{t=1}^{\infty} \gamma^{t} r\left(s_{t}, a_{t}\right) \mid \pi,M_0=\left(s, a, r, s^{\prime}\right)\right] \\
    = & \gamma \cdot \max _{\pi \in \Pi} V^{\pi_{(s, a, r)}}\left(s^{\prime}\right) \\
    = & \gamma \cdot \max _{\pi \in \Pi}  V^{\pi}\left(s^{\prime}\right) \\ 
    = & \gamma V^{\star}\left(s^{\prime}\right)
\end{aligned}.
$$
Therefore, we prove that there exists a policy $\pi$ such for all $s \in \mathcal{S}$, $V^\pi(s) = V^\ast(s)$.

\noindent Then, we rewrite the Eq. (\ref{eq:theorem1}) as the form of inner, 
$$
\begin{aligned}
    Q^\ast(s,a) = & r(s,a) + \gamma \mathbb{E}_{s' \sim P(s,a)} [\langle \pi(\cdot|s),Q^\ast(s,a) \rangle] \\
    = & r(s,a) + \gamma \mathbb{E}_{s' \sim P(s,a)}[\langle \tilde{\pi}(\cdot|s),Q^\ast(s,a) \rangle] .
\end{aligned}
$$
Obviously, we can obtain the relationship $\langle \pi,Q^\ast(s,a) \rangle = \langle \tilde{\pi},Q^\ast(s,a) \rangle$, so
$$
\langle \pi - \tilde{\pi}, Q^\ast(s,a) \rangle = 0.
$$
Because of the arbitrariness of $Q^\ast(s,a)$, we reduce that $\pi = \tilde{\pi}$. 

\bigskip

\hfill $\Box$

\bigskip

\noindent \textbf{Lemma \ref{plicy_improvement}} \textit{(Fenchel Policy Improvement) Let $\pi_{\text{old}} \in \prod$ and $\pi_{\text{new}}$ present the optimizer of the minimization problem defined in Eq. (\ref{eq:targe}). The $\pi_{\text{old}}$ is the policy before iteration, and $\pi_{\text{new}}$ is the policy after iteration. For all $(s,a) \in \mathcal{S} \times \mathcal{A}$ with $|\mathcal{A}| < \infty$, we have $Q^{\pi_{\text{old}}}(s,a) \leq Q^{\pi_{\text{new}}}(s,a)$. } 

\noindent\textbf{\emph{Proof.}} Let $\pi_{\text{old}} \in \prod$, let $Q^{\pi_{\text{old}}}$ and $V^{\pi_{\text{old}}} \in \prod$ be the corresponding state-action value and state value, and let $\pi_{\text{new}}$ be defined as
$$
\begin{aligned}
    \pi^\ast = & \min_{\pi \in \Pi} \mathbb{D}_f(\pi\|\tilde{\pi}) = & \min_{\pi \in \Pi} J_{\pi_{\text{old}}}(\pi),
\end{aligned}
$$
We must have the fact that $J_{\pi_{\text{old}}}(\pi_{\text{new}}) \leq J_{\pi_{\text{old}}}(\pi_{\text{old}})$, since we select $\pi_{\text{new}} = \pi_{\text{old}} \in \prod$. As the $f-$divergence needs satisfy the condition: $f(1) = 0$~\cite{nachum2020Duality}, when $x$ closes to $1$, we have $f'(x) \approx \log x + C$, where $C$ represents a constant. Hence,
$$
\begin{aligned}
D_f(\pi_{\text{new}} \|\tilde{\pi}_{\text{old}} ) \leq & D_f(\pi_{\text{old}} \|\tilde{\pi}_{\text{old}}) \\
\int \tilde{\pi}_{\text{old}}(a|s) f\left(\frac{\pi_{\text{new}}}{\tilde{\pi}_{\text{old}}}\right)  d a \leq & \int \tilde{\pi}_{\text{old}}(a|s) f\left(\frac{\pi_{\text{old}}}{\tilde{\pi}_{\text{old}}}\right)  d a \\
\int \tilde{\pi}_{\text{old}}(a|s) \max_{\omega_{\text{new}} \in \mathbb{R}}\left[\frac{\pi_{\text{new}}}{\tilde{\pi}_{\text{old}}}\omega_{\text{new}} - f_\ast(\omega_{\text{new}})\right]  d a \leq & \int \tilde{\pi}_{\text{old}}(a|s) \max_{\omega_{\text{old}} \in \mathbb{R}}\left[\frac{\pi_{\text{old}}}{\tilde{\pi}_{\text{old}}}\omega_{\text{old}} - f_\ast(\omega_{\text{old}})\right]  d a \\
\max_{\omega_{\text{new}} \in \mathcal{G}(\Xi)}\mathbb{E}_{\tilde{\pi}_{\text{old}}(a|s)} \left[\frac{\pi_{\text{new}}}{\tilde{\pi}_{\text{old}}}\omega_{\text{new}} - f_\ast(\omega_{\text{new}})\right] \overset{\text{(a)}}{\leq} & \max_{\omega_{\text{old}} \in \mathcal{G}(\Xi)}\mathbb{E}_{\tilde{\pi}_{\text{old}}(a|s)}\left[\frac{\pi_{\text{old}}}{\tilde{\pi}_{\text{old}}}\omega_{\text{old}} - f_\ast(\omega_{\text{old}})\right]  \\
\mathbb{E}_{\tilde{\pi}_{\text{old}}(a|s)} \left[\frac{\pi_{\text{new}}}{\tilde{\pi}_{\text{old}}}\omega^\ast_{\text{new}} - f_\ast(\omega^\ast_{\text{new}})\right] \leq & \mathbb{E}_{\tilde{\pi}_{\text{old}}(a|s)}\left[\frac{\pi_{\text{old}}}{\tilde{\pi}_{\text{old}}}\omega^\ast_{\text{old}} - f_\ast(\omega^\ast_{\text{old}})\right],  \\
\end{aligned}
$$
\noindent where the equation (a) can be reduced based on the Lemma \ref{lemma:1}. Besides, the policy $\pi$ always can be represented as a Boltzmann distribution \cite{landau2013course}, where the energy function of Boltzmann distribution is defined as the negative value function (when the $\beta \to +\infty$, the policy is a random policy, and the policy is considered as a greedy policy based on $\beta \to 0$.). Then, we write the sampling policy as the form of Boltzmann distribution, where the energy function of Boltzmann distribution is defined as the negative value function, $\tilde{\pi} = e^{\frac{1}{\beta}Q^{\pi_{\text{old}}}} / \sum_a e^{\frac{1}{\beta}Q^{\pi_{\text{old}}}}$, and the $\sum_a e^{\frac{1}{\beta}Q^{\pi_{\text{old}}}}$ is marked as $Z^{\pi_{\text{old}}}(s)$. Besides, with the analysis of Eq. (\ref{eq:optimal}), the following equation holds:
$$
\begin{aligned}
  & \omega_{\text{new}}^\ast = f' \left(\frac{\pi_{\text{new}}}{\tilde{\pi}_{\text{old}}} \right) \approx \log \frac{\pi_{\text{new}}}{\tilde{\pi}_{\text{old}}} + C = \log \pi_{\text{new}} - \frac{1}{\beta} Q^{\pi_{\text{old}}} + \log Z^{\pi_{\text{old}}}(s) \\
  & \omega_{\text{old}}^\ast = f' \left(\frac{\pi_{\text{old}}}{\tilde{\pi}_{\text{old}}} \right) \approx \log \frac{\pi_{\text{old}}}{\tilde{\pi}_{\text{old}}} + C = \log \pi_{\text{old}} - \frac{1}{\beta} Q^{\pi_{\text{old}}} + \log Z^{\pi_{\text{old}}}(s). \\
\end{aligned}
$$

\noindent If $\mathbb{E}_{\tilde{\pi}_{\text{old}}(a|s)} \left[ f_\ast(\omega^\ast_{\text{new}})\right] \leq \mathbb{E}_{\tilde{\pi}_{\text{old}}(a|s)}\left[f_\ast(\omega^\ast_{\text{old}})\right]$, we have,
$$
\begin{aligned}
\mathbb{E}_{\tilde{\pi}_{\text{old}}(a|s)} \left[\frac{\pi_{\text{new}}}{\tilde{\pi}_{\text{old}}}\omega^\ast_{\text{new}} \right] \leq & \mathbb{E}_{\tilde{\pi}_{\text{old}}(a|s)}\left[\frac{\pi_{\text{old}}}{\tilde{\pi}_{\text{old}}}\omega^\ast_{\text{old}} \right] \\
\mathbb{E}_{\pi_{\text{new}}(a|s)} \left[\omega^\ast_{\text{new}} \right] \leq & \mathbb{E}_{\pi_{\text{old}}(a|s)}\left[\omega^\ast_{\text{old}} \right] \\
\mathbb{E}_{\pi_{\text{new}}(a|s)} \left[\log \pi_{\text{new}} - \frac{1}{\beta} Q^{\pi_{\text{old}}} + \log Z^{\pi_{\text{old}}}(s) \right] \leq & \mathbb{E}_{\pi_{\text{old}}(a|s)}\left[\log \pi_{\text{old}} - \frac{1}{\beta} Q^{\pi_{\text{old}}} + \log Z^{\pi_{\text{old}}}(s) \right] \\
\mathbb{E}_{\pi_{\text{new}}(a|s)} \left[  Q^{\pi_{\text{old}}} - \beta\log \pi_{\text{new}} \right] \geq & \mathbb{E}_{\pi_{\text{old}}(a|s)}\left[Q^{\pi_{\text{old}}} - \beta\log \pi_{\text{old}} \right]. \\
\end{aligned}
$$
Then,
$$
\begin{aligned}
Q^{\pi_{\text{old}}}(s,a) = & r(s,a) + \gamma \mathbb{E}_{s' \sim T, a'\sim \pi_{\text{old}}}[Q^{\pi_{\text{old}}}(s',a') - \beta \log \pi_{\text{old}}] \\
\leq & r(s,a) + \gamma \mathbb{E}_{s' \sim T, a'\sim \pi_{\text{new}}}[Q^{\pi_{\text{old}}}(s',a') - \beta \log \pi_{\text{new}}] \\
& \vdots \\
= & Q^{\pi_{\text{new}}}(s,a),
\end{aligned}
$$
where we have repeatedly expanded $Q^{\pi_{\text{old}}}$ on the RHS by utilizing the Bellman equation and bound in above equation. In reality, we always utilize the Bellman equation with entropy regularization \cite{a2c,sac}. In our paper, the $f(x) = \frac{1}{2}(x-1)^2$, its conjugate function $f_\ast$ is $f_\ast(\omega) = \frac{1}{2}\omega^2 + \omega$, where $\omega = x -1$. So
$$
\begin{aligned}
\mathbb{E}_{\tilde{\pi}_{\text{old}}(a|s)} \left[ f_\ast(\omega^\ast_{\text{new}})\right] = & \mathbb{E}_{\tilde{\pi}_{\text{old}}(a|s)}\left[\frac{1}{2}\left(\omega^\ast_{\text{new}}\right)^2 + \omega^\ast_{\text{new}} \right] \\
= & \mathbb{E}_{\tilde{\pi}_{\text{old}}(a|s)}\left[\frac{1}{2}\left(\frac{\pi_{\text{new}}}{\tilde{\pi}_{\text{old}}}-1\right)^2 + \left(\frac{\pi_{\text{new}}}{\tilde{\pi}_{\text{old}}}-1\right) \right] \\
\overset{\text{(b)}}{\leq} & \mathbb{E}_{\tilde{\pi}_{\text{old}}(a|s)}\left[\frac{1}{2}\left(\frac{\pi_{\text{old}}}{\tilde{\pi}_{\text{old}}}-1\right)^2 + \left(\frac{\pi_{\text{old}}}{\tilde{\pi}_{\text{old}}}-1\right) \right] \\
= & \mathbb{E}_{\tilde{\pi}_{\text{old}}(a|s)} \left[ f_\ast(\omega^\ast_{\text{old}})\right] \\
\end{aligned}
$$
where the equation (b) can be reduced based on the fact, \textit{e.g.,} $J_{\pi_{\text{old}}}(\pi_{\text{new}}) \leq J_{\pi_{\text{old}}}(\pi_{\text{old}})$.

\hfill $\Box$

\bigskip

\subsection*{Appendix B}

\bigskip

\noindent \textbf{Lemma \ref{lemma:1}} \emph{Let $\xi$ is a random variable: $\forall \xi \in \Xi$, a function $g(\cdot,\xi) \to (-\infty,+\infty)$ is a proper, convex, lower semi-continuous function, then the following equation holds:
$$
\mathbb{E}_{\xi}[\max_{\omega \in \mathbb{R}} g(\omega,\xi)] = \max_{\omega(\cdot) \in \mathcal{G}(\Xi)} \mathbb{E}_\xi[g(\omega(\xi),\xi)],
$$
where $\mathcal{G}(\Xi) = \{\omega(\cdot): \Xi \to \mathbb{R}\}$ is the functional space.}

\bigskip

\noindent\textbf{\emph{Proof.}} As we assume that $g(\cdot,\xi)$ is proper, convex, lower semi-continuous function, there certainly exists a maximizer $\omega^\ast_\xi$ that satisfies $\forall \omega \in \mathbb{R}, g(\omega,\xi) \leq g(\omega^\ast_\xi,\xi)$. Therefore, we can define a function $\omega^\ast(\cdot): \mathcal{X} \to \mathbb{R}$; let $\omega^\ast(\xi) = \omega^\ast_\xi$, where $\omega^\ast(\xi) \in \mathcal{G}(\Xi)$.

\noindent On the one hand, 
$$
\begin{aligned}
    \mathbb{E}_{\xi}[\max_{\omega \in \mathbb{R}} g(\omega,\xi)] 
    & = \mathbb{E}_{\xi}[ g(\omega^\ast(\xi),\xi)] \leq \max_{\omega(\cdot) \in \mathcal{G}(\Xi)} \mathbb{E}_\xi[g(\omega(\xi),\xi)].
\end{aligned}
$$

\noindent On the other hand, $\forall \omega(\cdot) \in \mathcal{G}(\Xi)$, $\xi \in \Xi$, we have $g(\omega(\xi),\xi) \leq \max_{\omega \in \mathbb{R}} g(\omega,\xi)$. So, for $\forall \omega(\cdot) \in \mathcal{G}(\Xi)$,  $\mathbb{E}_\xi[g(\omega(\xi),\xi)] \leq \mathbb{E}_\xi[ \max_{\omega \in \mathbb{R}} g(\omega,\xi)]$. This further implies that,
$$
\begin{aligned}
   \max_{\omega(\cdot) \in \mathcal{G}(\Xi)} \mathbb{E}_\xi[g(\omega(\xi),\xi)]
    & \leq  \mathbb{E}_{\xi}[\max_{\omega \in \mathbb{R}} g(\omega,\xi)].
\end{aligned}
$$
Combining with the above derivation, we can conclude that $\mathbb{E}_{\xi}[\max_{\omega \in \mathbb{R}} g(\omega,\xi)] = \max_{\omega(\cdot) \in \mathcal{G}(\Xi)} \mathbb{E}_\xi[g(\omega(\xi),\xi)]$. 

\bigskip

\hfill $\Box$

\bigskip

% \subsection*{Append C}

\subsection*{Appendix C}
\noindent \textbf{Assumption \ref{asm:4}} \textit{We solve the inner optimization problem in Eq. (\ref{eq:target2}), \textit{i.e.},  $\omega(\cdot): \Xi \to \mathbb{R}$, in a constrained functional space $\Omega$: $\omega_{\pi / \tilde{\pi}}  = Q^\pi(s,a) - Q^{\tilde{\pi}}(s,a)$.}

\noindent On the one hand, as the $f$ function is a proper, convex, lower semi-continuous function, we deduce the connection between $\omega^\ast(\cdot)$ and $f'$ in the Eq. (\ref{eq:optimal}) based on the convex conjugate property \cite{algaedice},
$$
    \omega^\ast_{\pi / \tilde{\pi}} = f'\left( \pi/\tilde{\pi}\right).
$$
When the $f$ function is defined as a specific form: $f(x) = \frac{1}{2} (x-1)^2$, we deduce 
$$
\omega^\ast_{\pi / \tilde{\pi}} = f'\left( \pi/\tilde{\pi}\right) = \frac{\pi}{\tilde{\pi}} - 1.
$$

On the other hand, the policy $\pi$ always can be represented as a Boltzmann distribution \cite{landau2013course}, where the energy function of Boltzmann distribution is defined as the negative value function,
$$
\pi = \frac{\exp{\left\{\frac{1}{\beta} Q^\pi(s,a)\right\}}}{\sum_a \exp{\left\{\frac{1}{\beta} Q^\pi(s,a)\right\}}}.
$$
For example, when the $\beta \to +\infty$, the policy is a random policy, and the policy can be considered as a greedy policy based on $\beta \to 0$. So the $w^\ast_{\pi/\tilde{\pi}}$ can be approximated as,
$$
w^\ast_{\pi/\tilde{\pi}} \propto \exp \left\{ \frac{1}{\beta} \left( Q^\pi(s,a) - Q^{\tilde{\pi}}(s,a) \right) \right\} \propto Q^\pi(s,a) - Q^{\tilde{\pi}}(s,a).
$$
That inspires us that constrain the the solution of $\omega(\cdot)$ in a functional space: $\omega_{\pi/\tilde{\pi}} = Q^\pi(s,a) - Q^{\tilde{\pi}}(s,a)$ to build to connection between the optimization problem and state-action value function.

\bigskip

\bigskip

\subsection*{Appendix D}

\noindent \textbf{Theorem \ref{theorem:4}}
\emph{(Fenchel Policy Gradient Theorem) If the dual function $Q^\pi$ is fixed, the gradient of FRL objective function $\mathcal{L}_f (\pi, Q^\pi)$ with respect to $\pi$ is,} 
$$
    \begin{aligned}
        & \frac{\partial }{\partial \pi} \mathcal{L}_f (\pi, Q^\pi) = \\
        & \quad \mathbb{E}_{\tau \sim \pi}\left[ \left( Q^{\pi}(s,a) - Q^{\tilde{\pi}}(s,a)\right) \nabla \log \pi(a|s)\right]. \\
    \end{aligned}
$$
\emph{where $Q^\pi(s,a)$ and $Q^{\tilde{\pi}}(s,a)$ are the $Q$ function of learning policy $\pi$ and sampling policy $\tilde{\pi}$ respectively.}

% \emph{Admittedly, in traditional RL algorithms, the state-action function $Q^\pi(s,a)$ is usually approximated by value function $V^\pi(s)$, which presents that the policy gradient based on Bellman error is equivalent to the policy gradient based on advantage function.}

\bigskip

\noindent\textbf{\emph{Proof.}} Note that all the gradient symbols $\nabla$ in this section represent the gradient operation w.r.t. policy $\pi$. The $\frac{\partial }{\partial \pi} \min_{Q^\pi} \mathcal{L}_f (\pi, Q^\pi)$ is decomposed into three aspects: 
$$
\begin{aligned}
    \begin{aligned}
            & (1) =  \nabla \mathbb{E}_{\pi} \left[ \left(Q^\pi(s,a) - Q^{\tilde{\pi}}(s,a) \right) \right] \\
            & (2) = - \nabla \mathbb{E}_{\tilde{\pi}}\left[\left(Q^\pi(s,a) - Q^{\tilde{\pi}}(s,a)\right)\right] \\
            & (3) = - \nabla \mathbb{E}_{\tilde{\pi}}\left[\left(Q^\pi(s,a) - Q^{\tilde{\pi}}(s,a)\right)^2\right] \\
        \end{aligned}
\end{aligned}
$$
Then, we calculate the gradient of each part separately,
$$
\begin{aligned}
    (1) & = \nabla \int \pi(a|s)\left[Q^\pi(s,a) - R(s,a)\right] da \\
    & = \int \nabla \pi(a|s)\left[Q^\pi(s,a) - Q^{\tilde{\pi}}(s,a)\right] da +  \int \pi(a|s) \nabla Q^\pi(s,a) da \\
      & \overset{\text{(a)}}{=} \int \nabla \pi(a|s)\left[Q^\pi(s,a) - Q^{\tilde{\pi}}(s,a)\right] da + \int \frac{\pi}{\tilde{\pi}} \nabla \pi(a|s) da \\
     & = \int \nabla \pi(a|s)\left[Q^\pi(s,a) - Q^{\tilde{\pi}}(s,a)\right] da +  \int \nabla \pi(a|s)\left[Q^\pi(s,a) - Q^{\tilde{\pi}}(s,a) + 1\right] da \\
     & = 2 \int \nabla \pi(a|s)\left[Q^\pi(s,a) - Q^{\tilde{\pi}}(s,a)\right] da \\
     & = 2 \mathbb{E}_{\pi}\left[ \nabla \log \pi(a|s)\left[Q^\pi(s,a) - Q^{\tilde{\pi}}(s,a)\right] \right]\\
\end{aligned}
$$
Where the equation (a) can be reduced based on follow relationship. According to the optimal connection in Eq. (\ref{eq:optimal}) and Assumption \ref{asm:4}, the fact that the derivatives $f'$ and $f_\ast'$ are inverses of each other ($f(x) = \frac{1}{2}(x-1)^2$), we can built the connection between the policy $\pi$ and its state-action function $Q^\pi(s,a)$,
$$
\begin{gathered}
    \omega^\ast_{\pi / \tilde{\pi}} = Q^\pi(s,a) - Q^{\tilde{\pi}}(s,a) = f'\left( \pi/\tilde{\pi}\right) \\
    Q^\pi(s,a) - Q^{\tilde{\pi}}(s,a) = \frac{\pi}{\tilde{\pi}} + 1.
\end{gathered}
$$
so we have $\nabla Q^\pi(s,a) = \frac{\nabla \pi}{\tilde{\pi}}$. The $Q^\pi(s,a)$ is induced by the Bellman error function with a certain learning policy $\pi$, so that when we calculate the policy gradient,  the gradient of $Q^\pi(s,a)$ with respect to $\pi$ can not be ignored. For the second term, we have
$$
\begin{aligned}
   (2) = & - \nabla \int \tilde{\pi}(a|s)\left[Q^\pi(s,a) - Q^{\tilde{\pi}}(s,a)\right] \\
   = & - \int \tilde{\pi}(a|s) \nabla Q^\pi(s,a) \\
   = & - \int \nabla \pi(a|s) da = 0.
\end{aligned}
$$
Analogous to the derivation for the second term, for the third term, we have
$$
\begin{aligned}
   (3) = & - \frac{1}{2} \nabla \mathbb{E}_{\tilde{\pi}}\left[(Q^\pi(s,a) - Q^{\tilde{\pi}}(s,a))^2\right] \\
   = & -\int \tilde{\pi}(a|s)\left[Q^\pi(s,a) - Q^{\tilde{\pi}}(s,a)\right] \nabla Q^\pi(s,a) da \\
   = & -\int \pi(a|s)\left[Q^\pi(s,a) - Q^{\tilde{\pi}}(s,a)\right] da \\
   = & - \mathbb{E}_{\pi}\left[ \nabla \log \pi(a|s)\left[Q^\pi(s,a) - Q^{\tilde{\pi}}(s,a) \right] \right].\\
\end{aligned}
$$
Then, merging the deriving results of (1), (2) and (3), the gradient of the reinforcement learning in $f-$divergence form $\mathcal{L}_f (\pi, Q^\pi)$ with respect to $\pi$ is obtained,
$$
    \begin{aligned}
        & \frac{\partial }{\partial \pi} \mathcal{L}_f (\pi, Q^\pi) =  \mathbb{E}_{\tau \sim \pi}\left[ \left(Q^{\pi}(s,a) - Q^{\tilde{\pi}}(s,a)\right) \nabla \log \pi(a|s)\right]. \\
    \end{aligned}
$$
As we analyzed in Appendix C, $Q^{\tilde{\pi}}(s,a)$ satisfies the Bellman equation, $Q^{\tilde{\pi}}(s,a) = r(s,a) + \gamma \mathbb{E}_{\tilde{\pi}}[Q^{\tilde{\pi}}(s',a')]$. Therefore, the policy gradient can be rewritten as,
$$
\begin{aligned}
 & \frac{\partial }{\partial \pi} \mathcal{L}_f (\pi, Q^\pi) =  \mathbb{E}_{\tau \sim \pi}\left[ \left(Q^{\pi}(s,a) - r(s,a) - \gamma \mathbb{E}_{\tilde{\pi}}[Q^{\tilde{\pi}}(s',a')]\right) \nabla \log \pi(a|s)\right].
\end{aligned}
$$

\bigskip

\hfill $\Box$

\bigskip

\noindent \textbf{Theorem} \ref{theorem:5} \emph{(Policy Evaluation in FRL) When the outer optimization object is fixed, the object function of policy evaluation is,}
$$
    \begin{aligned}
        & \frac{\partial}{\partial Q^\pi} \mathcal{L}_f (\pi, Q^\pi) =  - \mathbb{E}_{\tilde{\pi}}\left[ \left( Q^\pi(s,a) - Q^{\tilde{\pi}}(s,a) \right)\nabla Q^\pi(s,a) \right]  + \mathbb{E}_{\pi} \left[\nabla Q^\pi(s,a)\right] - \mathbb{E}_{\tilde{\pi}} \left[\nabla Q^\pi(s,a)\right]
    \end{aligned}
$$
\noindent\textbf{\emph{Proof.}} Note that all the gradient symbols $\nabla$ in this section represent the gradient operation w.r.t value function $Q$. Analogous to the proof for Theorem \ref{theorem:4}, the $\frac{\partial}{\partial Q^\pi}\min_{\pi} \mathcal{L}_f (\pi, Q^\pi)$ is decomposed into three aspects, 

For the first term,
$$
\begin{aligned}
    & -\frac{1}{2} \nabla \mathbb{E}_{\tilde{\pi}}\left[\left(Q^\pi(s,a) - Q^{\tilde{\pi}}(s,a)\right)^2\right] =  -\mathbb{E}_{\tilde{\pi}} \left[\left(Q^\pi(s,a) - Q^{\tilde{\pi}}(s,a)\right)\nabla Q^\pi(s,a)\right],
\end{aligned}
$$

For the second term,
$$
\begin{aligned}
- \nabla \mathbb{E}_{\tilde{\pi}}\left[\left(Q^\pi(s,a) - Q^{\tilde{\pi}}(s,a)\right)\right] = - \mathbb{E}_{\tilde{\pi}}\left[\left(\nabla Q^\pi(s,a)\right)\right],
\end{aligned}
$$

For the third term,
$$
\begin{aligned}
\nabla \mathbb{E}_{\pi}\left[\left(Q^\pi(s,a) - Q^{\tilde{\pi}}(s,a)\right)\right] = - \mathbb{E}_{\pi}\left[\left(\nabla Q^\pi(s,a)\right)\right].
\end{aligned}
$$

We merge the above derivation and obtain the result. 
%Obviously, the derivation of policy evaluation is relatively simple than policy improvement's. 

\hfill $\Box$

\bigskip

\subsection{Appendix E}
\begin{figure*}[htbp]
\centering
\subfigure{
\includegraphics[width=0.75 in]{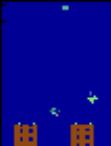}
}
\hspace{-0.3 cm}
\subfigure{
\includegraphics[width=0.75 in]{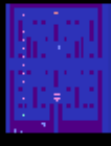}
}
\hspace{-0.3 cm}
\subfigure{
\includegraphics[width=0.75 in]{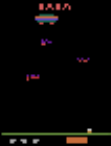}
}
\hspace{-0.3 cm}
\subfigure{
\includegraphics[width=0.75 in]{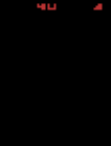}
}
\hspace{-0.3 cm}
\subfigure{
\includegraphics[width=0.75 in]{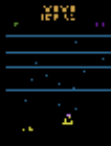}
}
\hspace{-0.3 cm}
\subfigure{
\includegraphics[width=0.75 in]{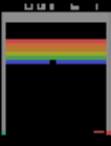}
}
\hspace{-0.3 cm}
\subfigure{
\includegraphics[width=0.75 in]{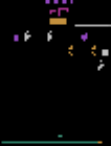}
}

\subfigure{
\includegraphics[width=0.75 in]{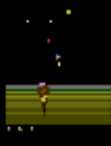}
}
\hspace{-0.3cm}
\subfigure{
\includegraphics[width=0.75 in]{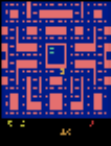}
}
\hspace{-0.3 cm}
\subfigure{
\includegraphics[width=0.75 in]{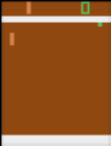}
}
\hspace{-0.3 cm}
\subfigure{
\includegraphics[width=0.75 in]{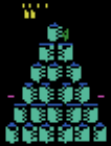}
}
\hspace{-0.3 cm}
\subfigure{
\includegraphics[width=0.75 in]{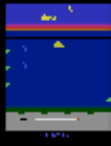}
}
\hspace{-0.3cm}
\subfigure{
\includegraphics[width=0.75 in]{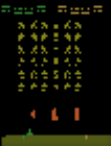}
}
\hspace{-0.3 cm}
\subfigure{
\includegraphics[width=0.75 in]{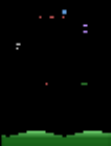}
}
\caption{\textbf{Some experiment environments.} A part of Atari 2600 games in our experiments are presented here, namely AirRaid, Alien, Assault, Asteroids, BeamRider, Breakout, Carnival, Jamesbond, MsPacman, Pong, Qbert, Seaquest, SpaceInvaders, StarGunner respectively (From left to right, up to down).} 
\label{fig:atari}
\end{figure*}

\begin{figure*}[t]
\centering
\subfigure{
\includegraphics[width=1.2 in]{atari_rewards/DoubleDunk.pdf}
}
\hspace{-0.6 cm}
\subfigure{
\includegraphics[width=1.2 in]{atari_rewards/Enduro.pdf}
}
\hspace{-0.6 cm}
\subfigure{
\includegraphics[width=1.2 in]{atari_rewards/FishingDerby.pdf}
}
\hspace{-0.6 cm}
\subfigure{
\includegraphics[width=1.2 in]{atari_rewards/Freeway.pdf}
}
\hspace{-0.6 cm}
\subfigure{
\includegraphics[width=1.2 in]{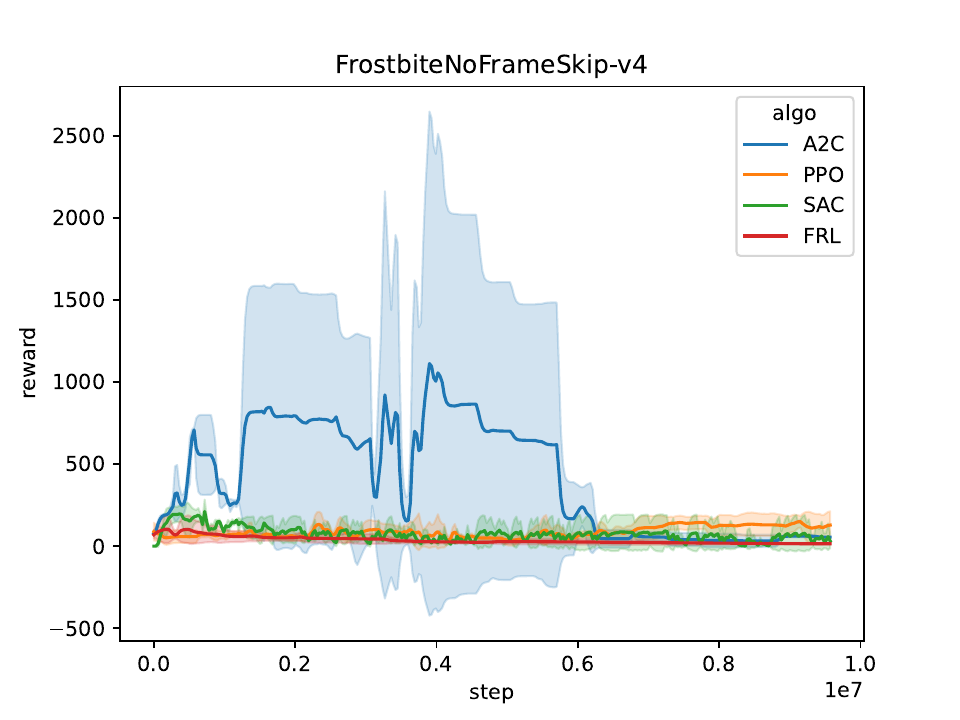}
}
\hspace{-0.6 cm}
\subfigure{
\includegraphics[width=1.2 in]{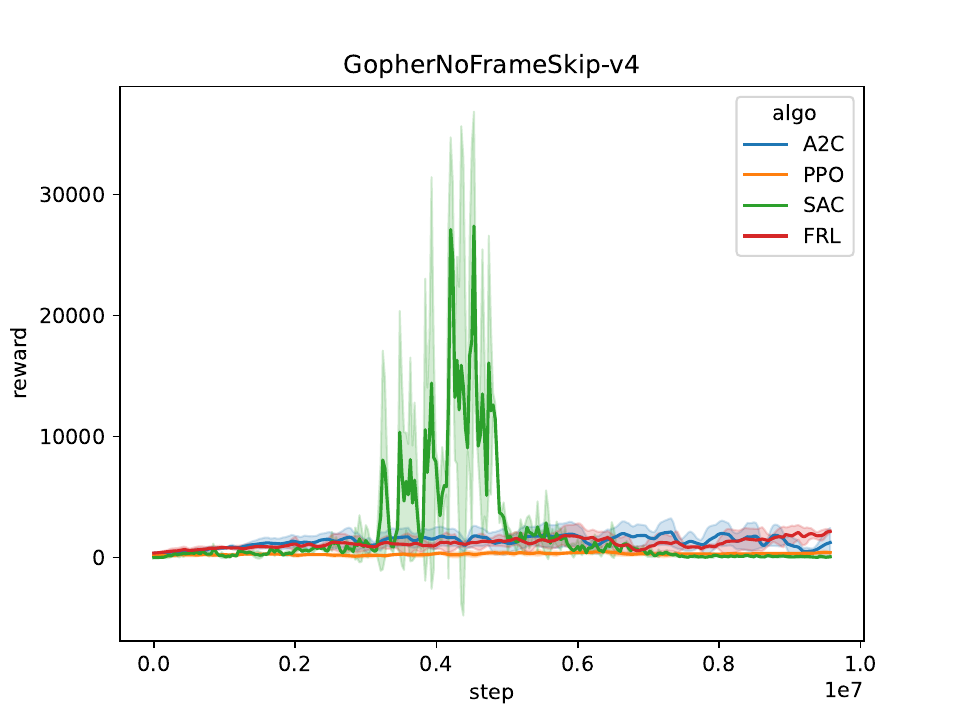}
}
\subfigure{
\includegraphics[width=1.2 in]{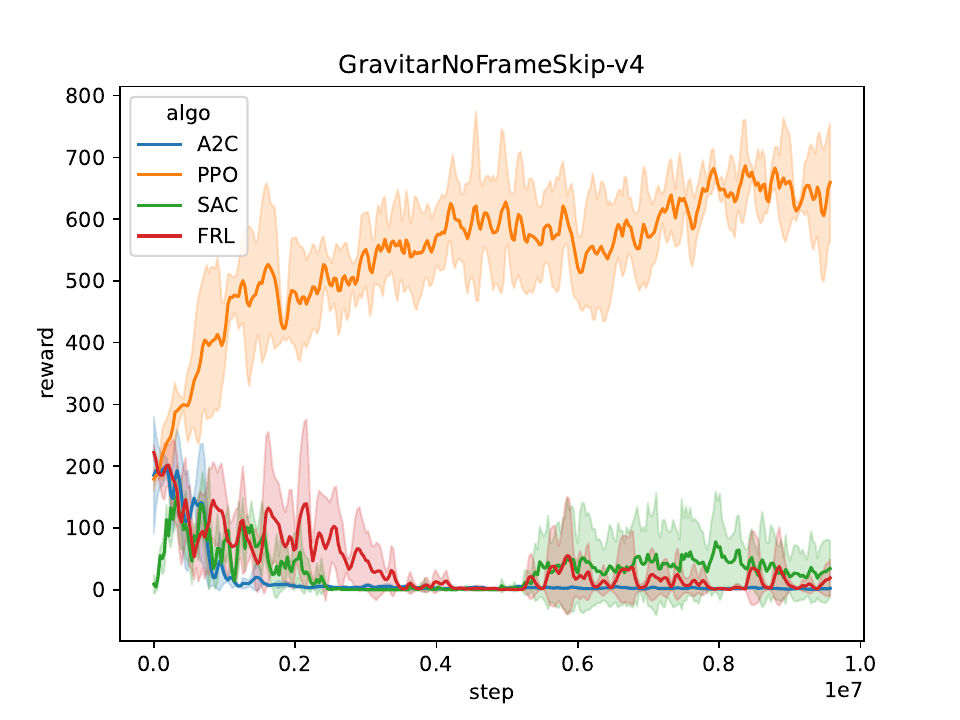}
}
\hspace{-0.6 cm}
\subfigure{
\includegraphics[width=1.2 in]{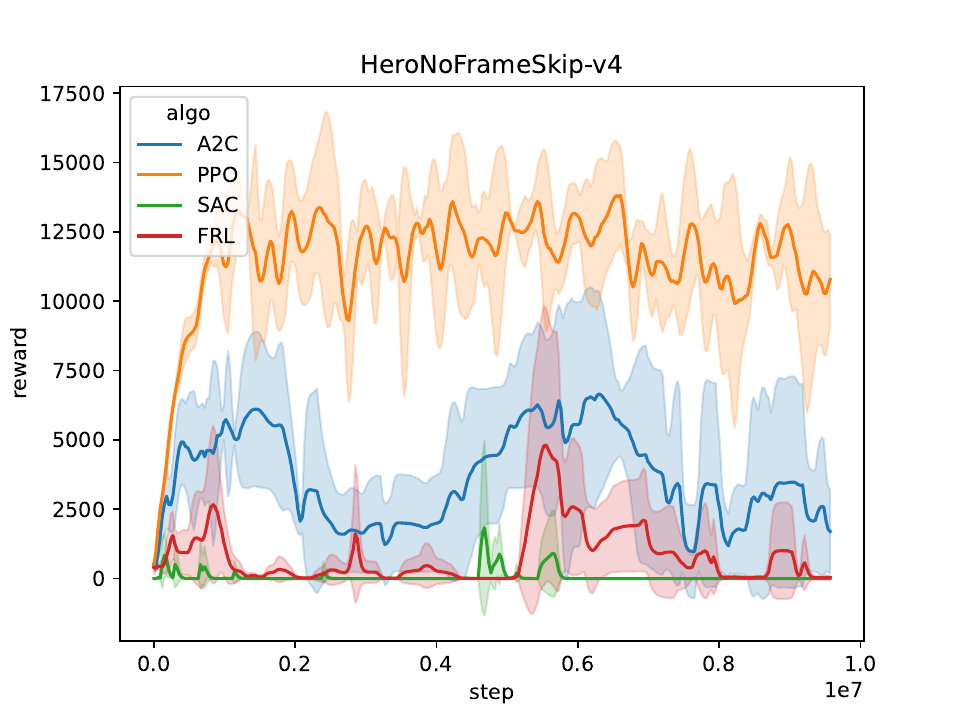}
}
\hspace{-0.6 cm}
\subfigure{
\includegraphics[width=1.2 in]{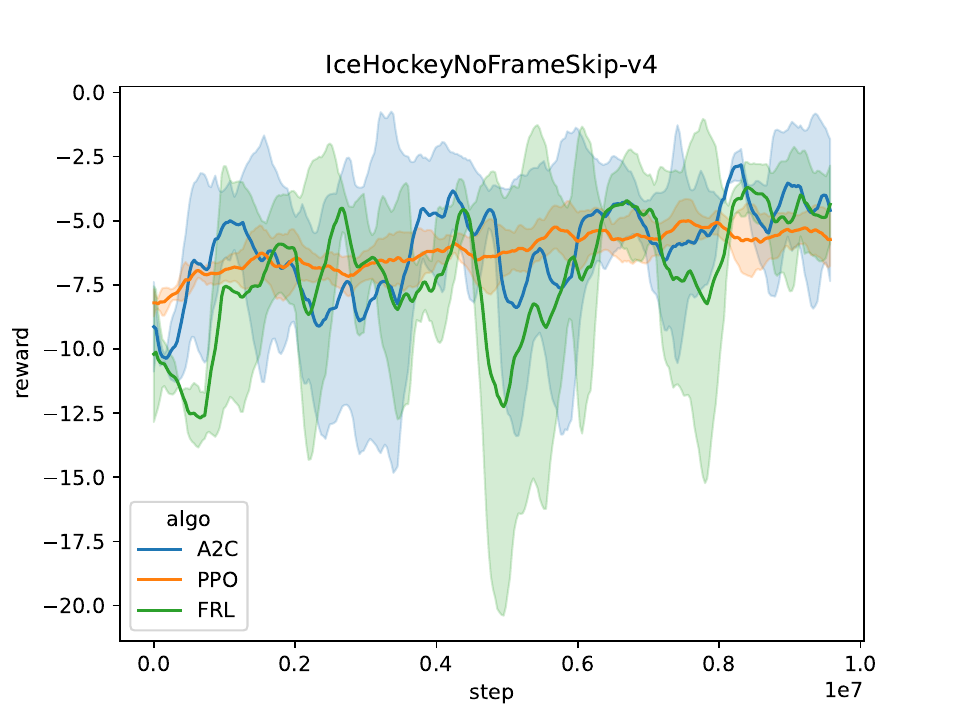}
}
\hspace{-0.6 cm}
\subfigure{
\includegraphics[width=1.2 in]{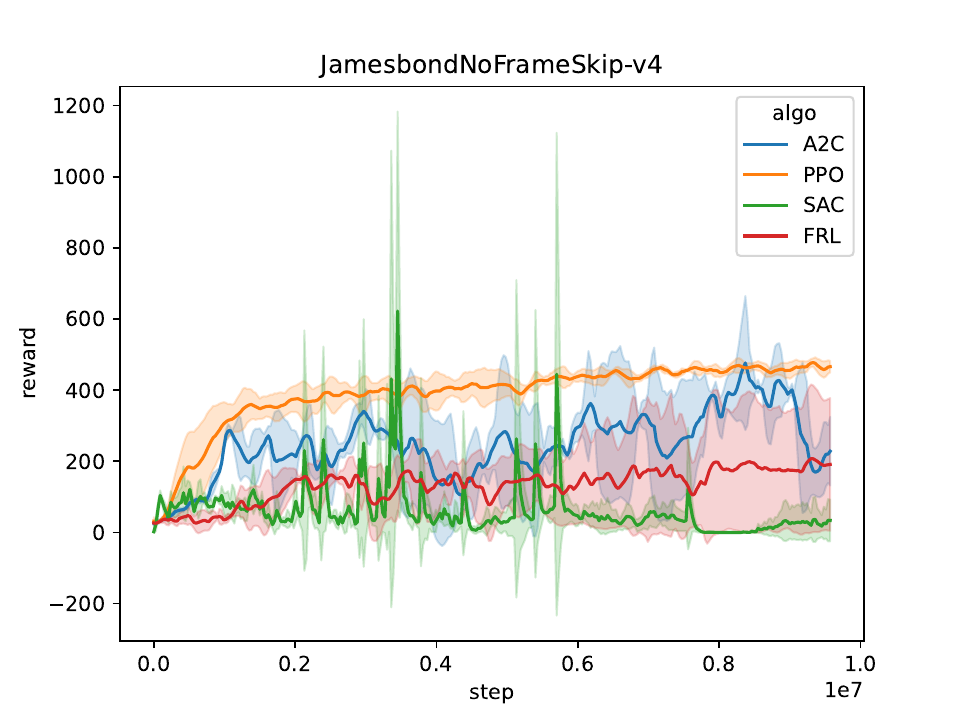}
}
\hspace{-0.6 cm}
\subfigure{
\includegraphics[width=1.1 in]{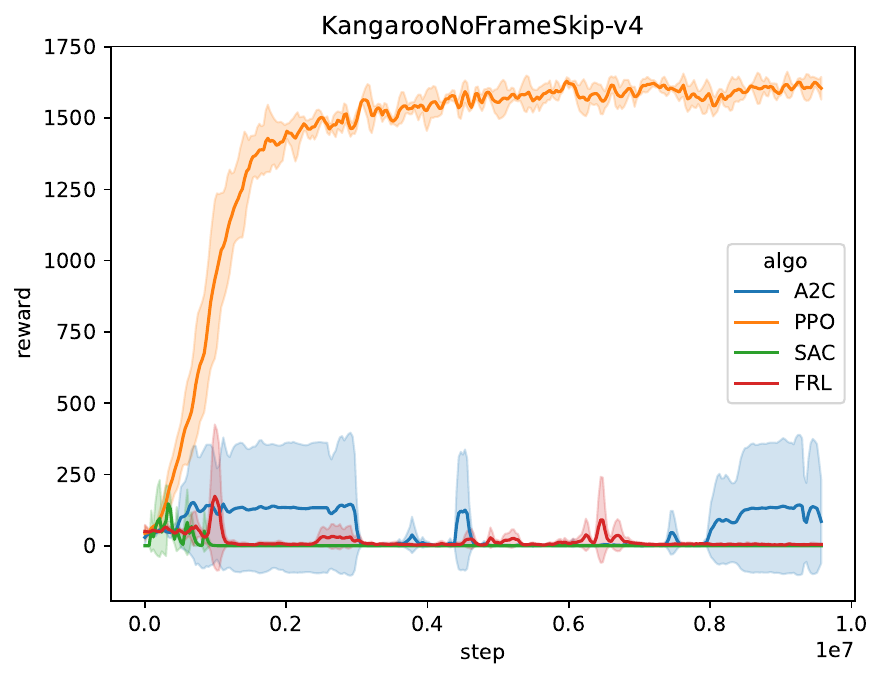}
}
\hspace{-0.4 cm}
\subfigure{
\includegraphics[width=1.2 in]{atari_rewards/Krull.pdf}
}
\subfigure{
\includegraphics[width=1.2 in]{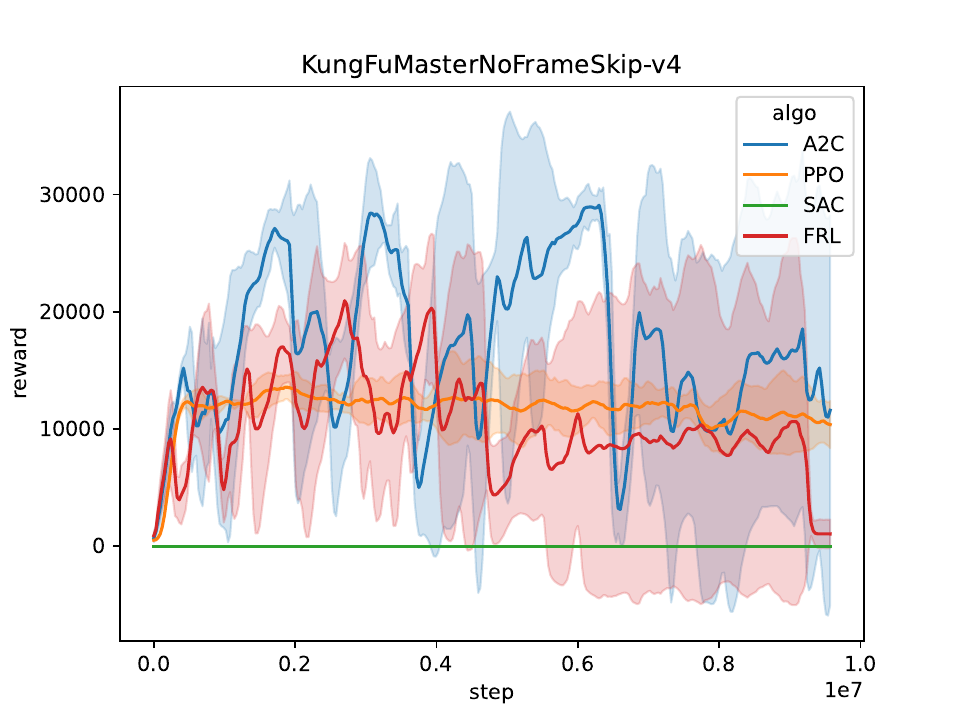}
}
\hspace{-0.6 cm}
\subfigure{
\includegraphics[width=1.2 in]{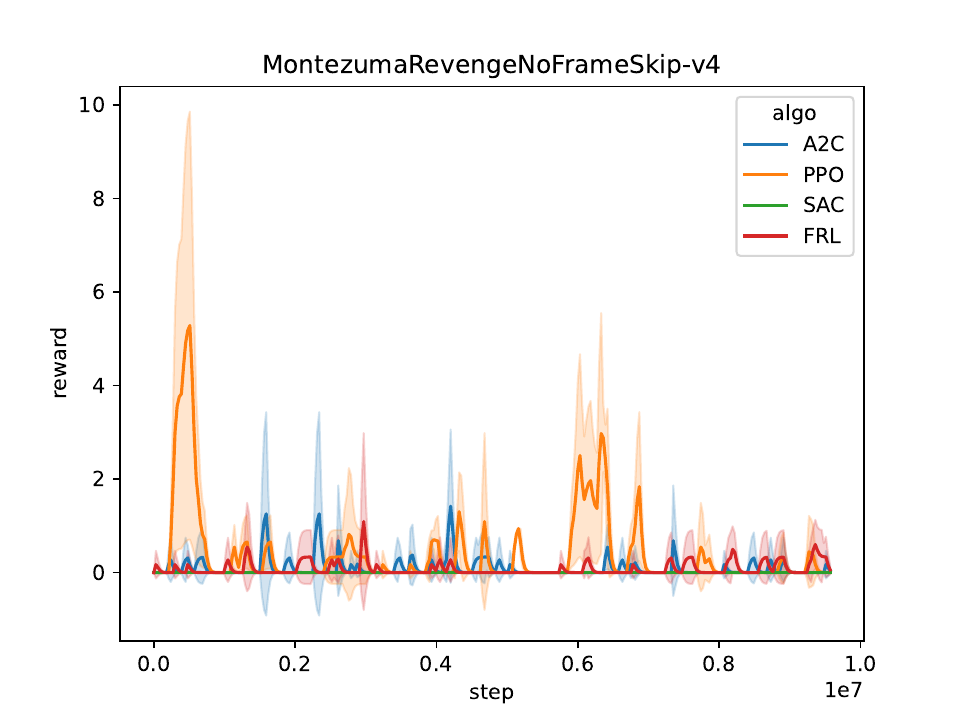}
}
\hspace{-0.6 cm}
\subfigure{
\includegraphics[width=1.2 in]{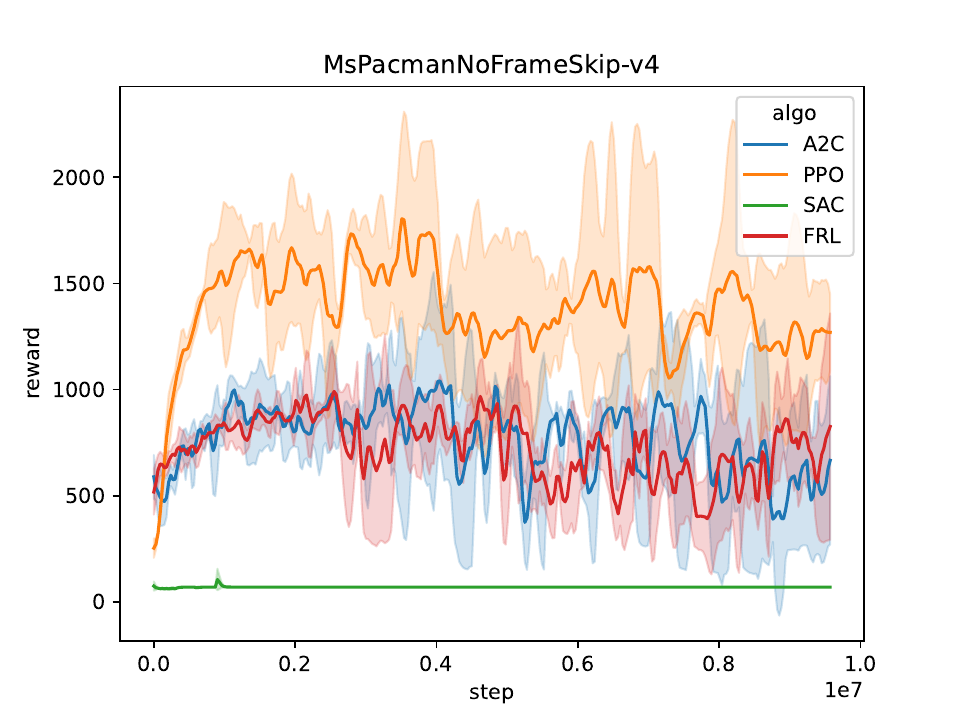}
}
\hspace{-0.6 cm}
\subfigure{
\includegraphics[width=1.2 in]{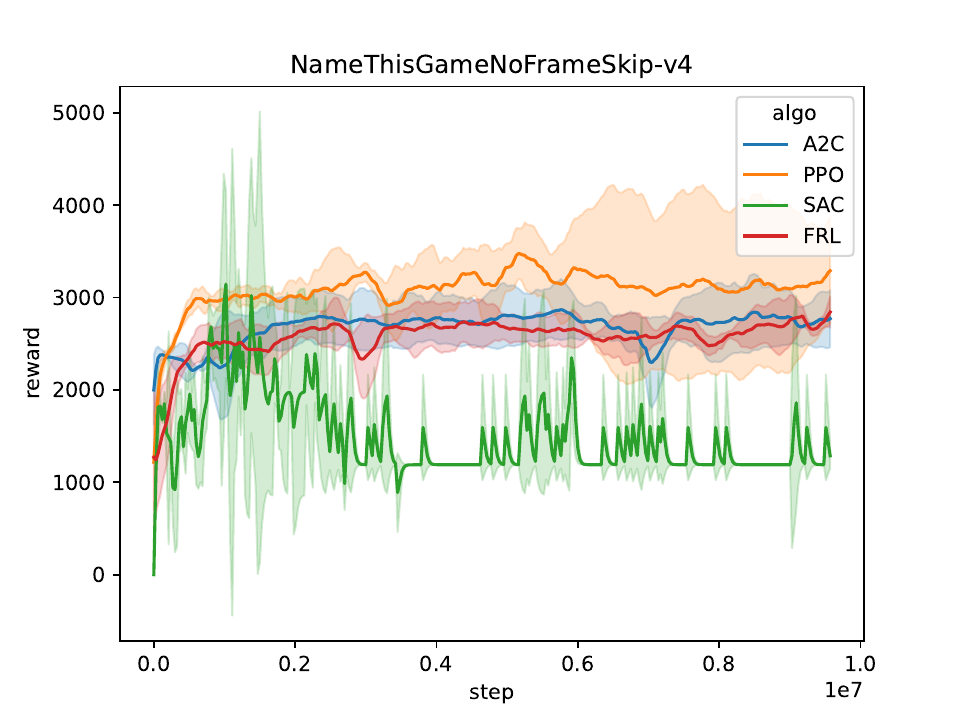}
}
\hspace{-0.6 cm}
\subfigure{
\includegraphics[width=1.2 in]{atari_rewards/Phoenix.pdf}
}
\hspace{-0.6 cm}
\subfigure{
\includegraphics[width=1.05 in]{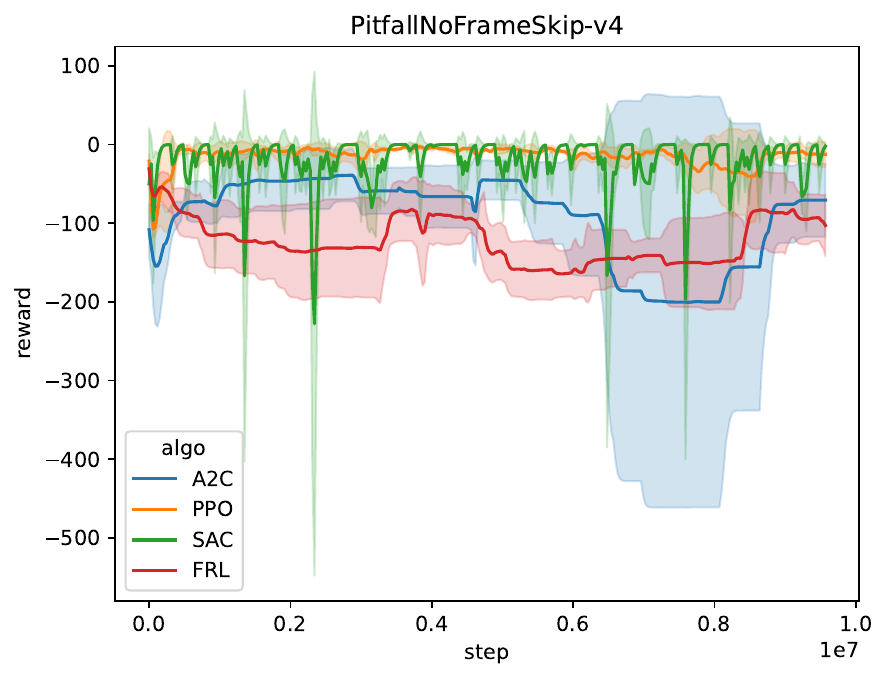}
}
\subfigure{
\includegraphics[width=1.2 in]{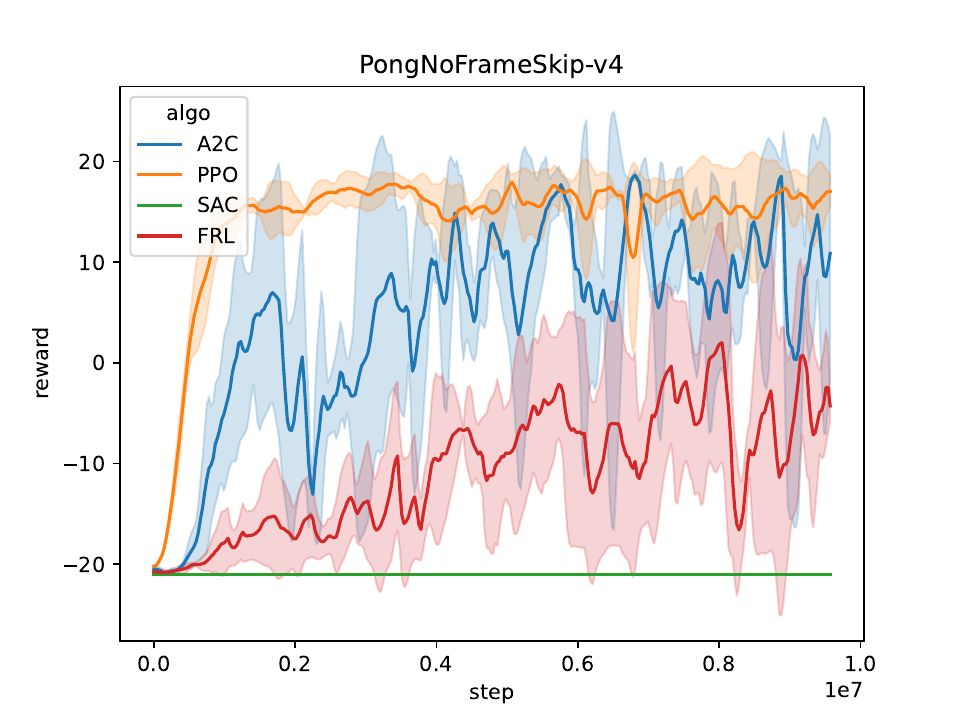}
}
\hspace{-0.6 cm}
\subfigure{
\includegraphics[width=1.2 in]{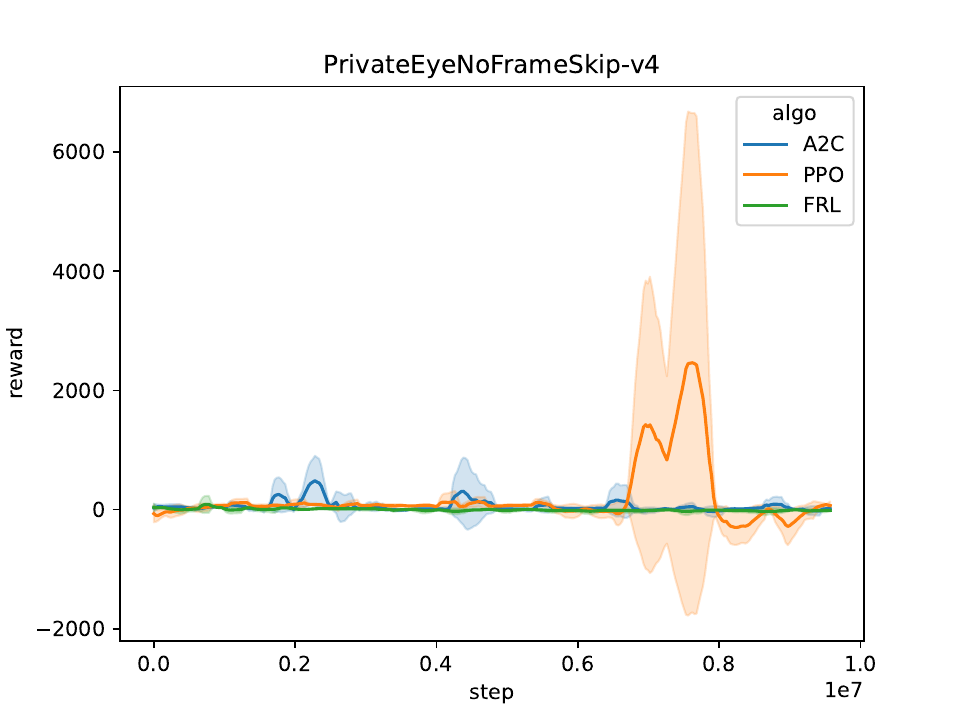}
}
\hspace{-0.6 cm}
\subfigure{
\includegraphics[width=1.2 in]{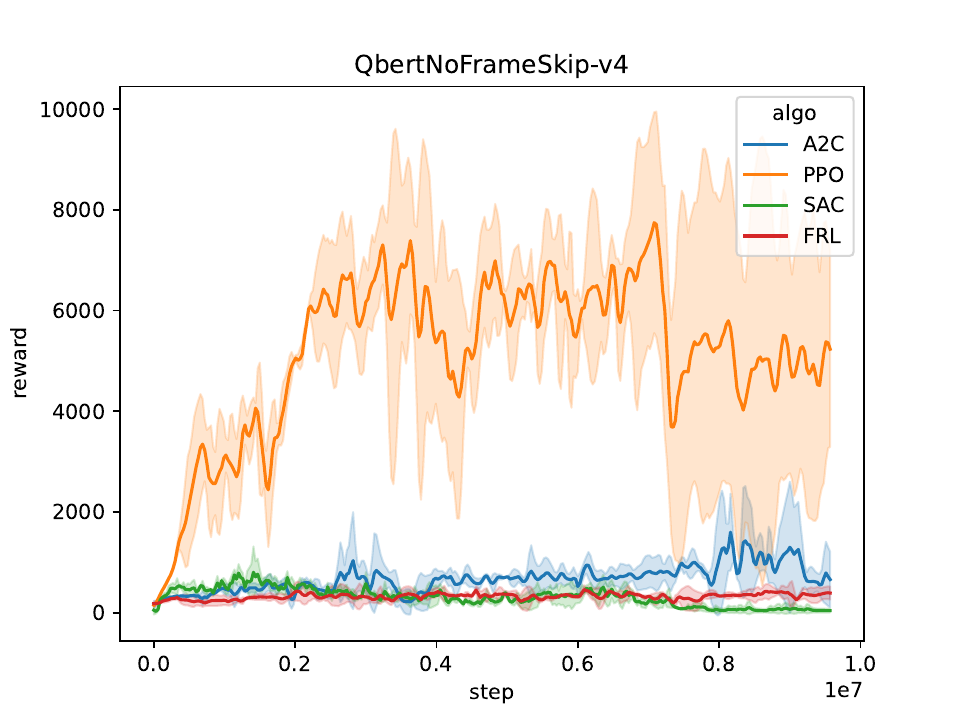}
}
\hspace{-0.6 cm}
\subfigure{
\includegraphics[width=1.2 in]{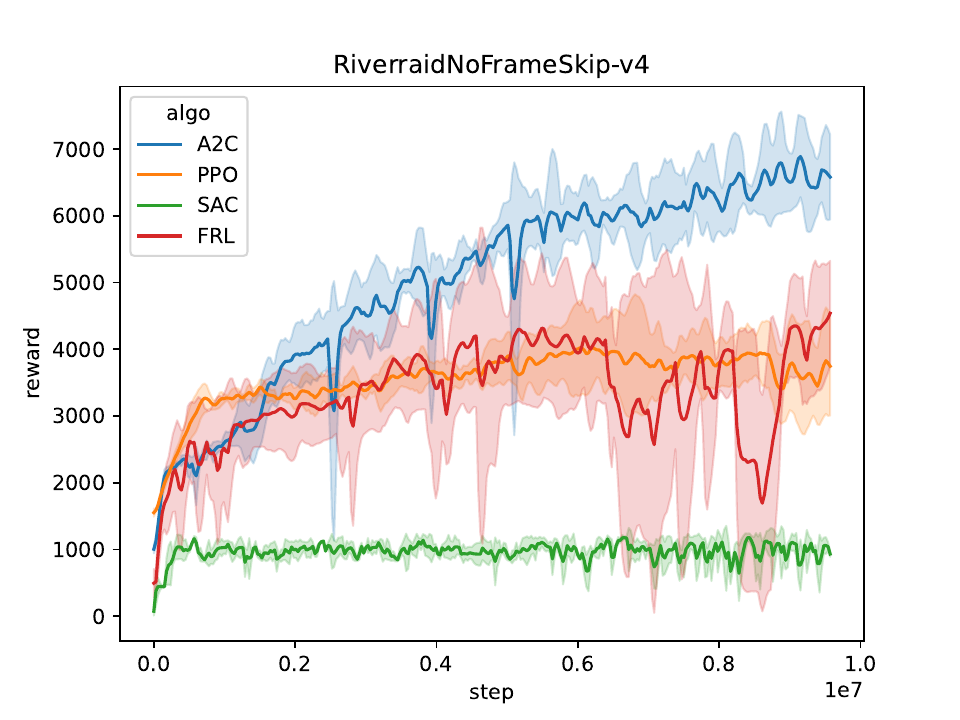}
}
\hspace{-0.6 cm}
\subfigure{
\includegraphics[width=1.2 in]{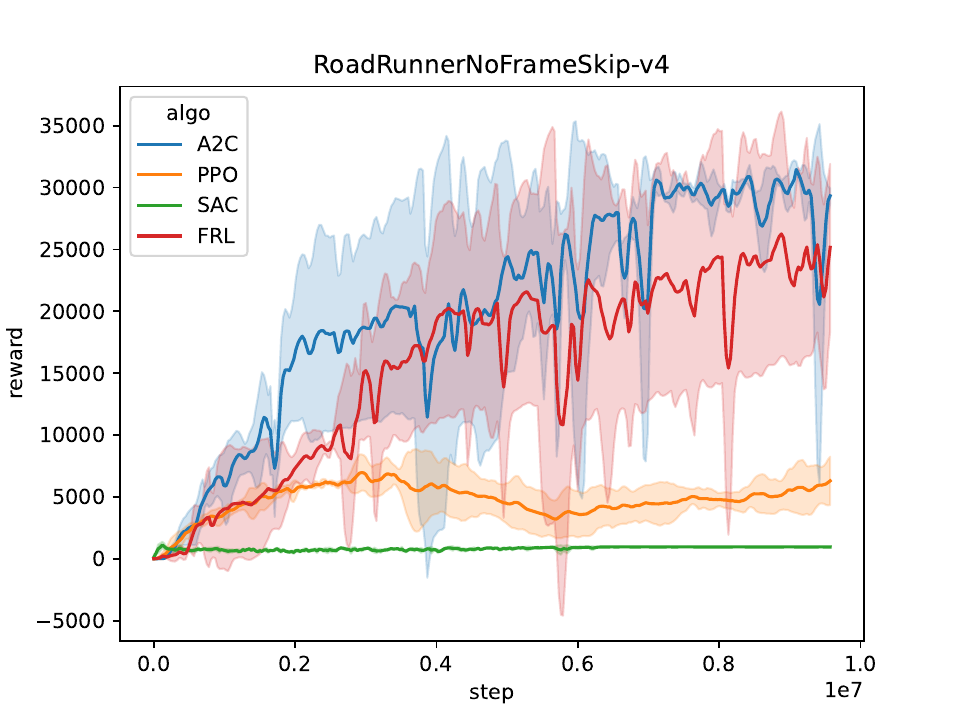}
}
\hspace{-0.6 cm}
\subfigure{
\includegraphics[width=1.2 in]{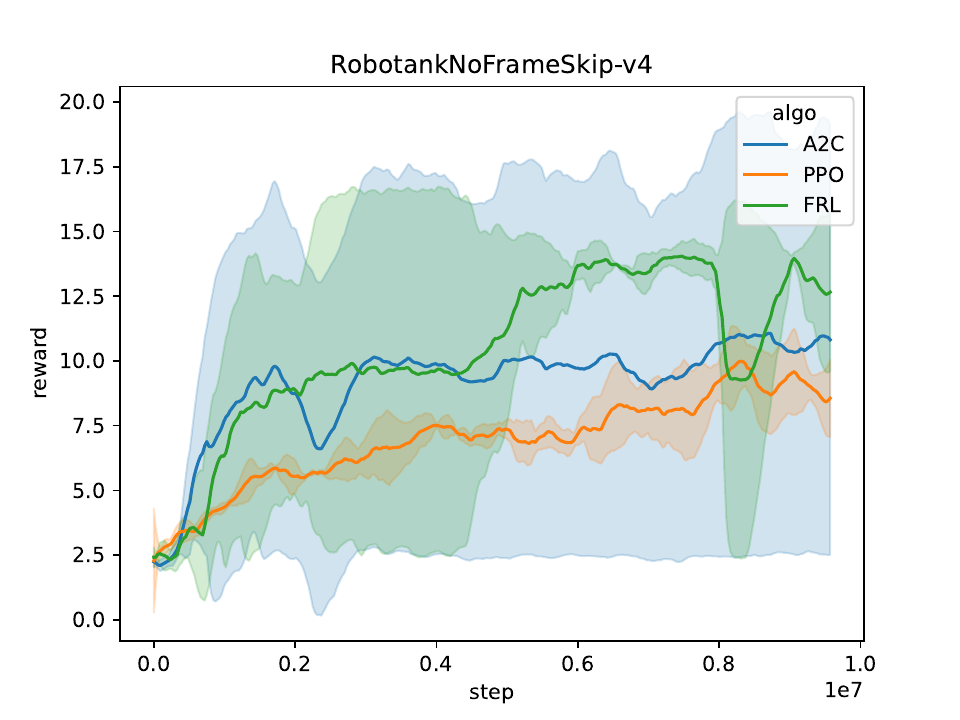}
}
% \{-0.4 cm}
\subfigure{
\includegraphics[width=1.2 in]{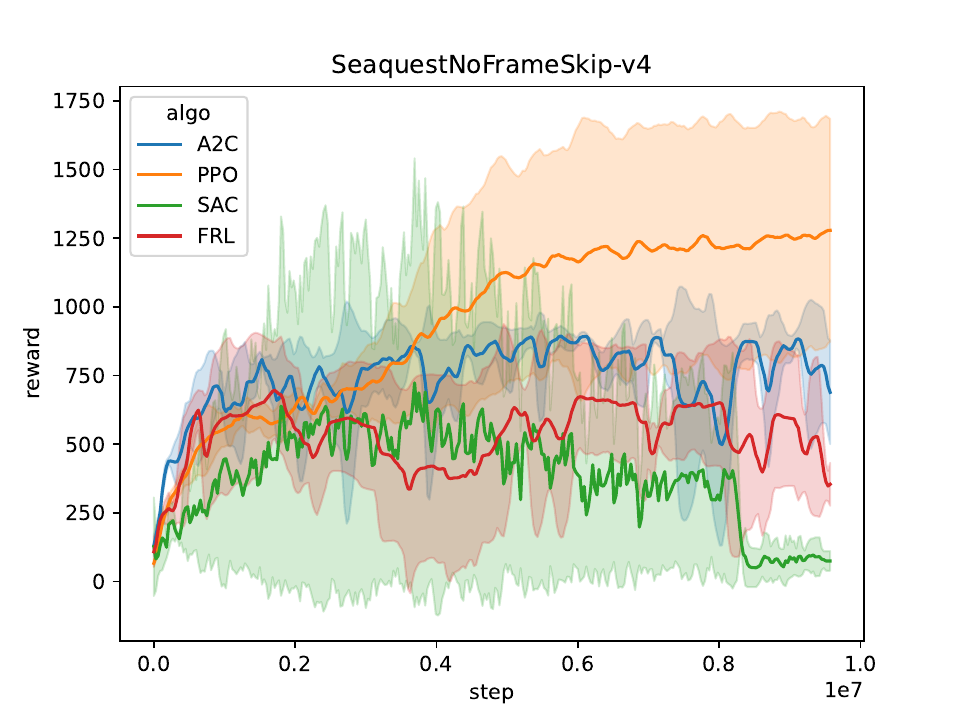}
}
\hspace{-0.6 cm}
\subfigure{
\includegraphics[width=1.2 in]{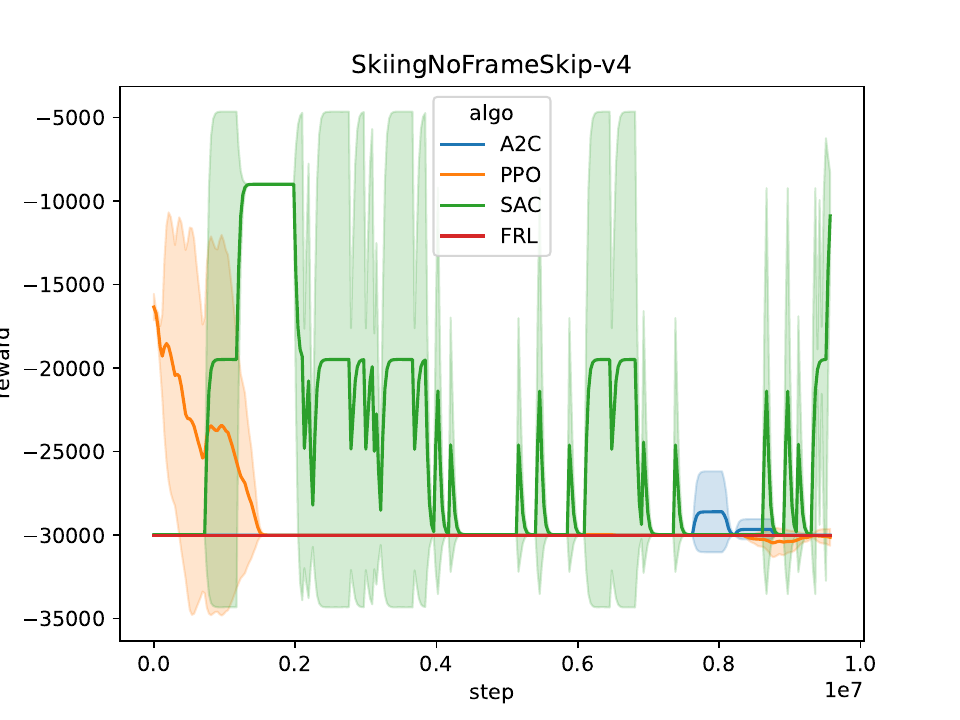}
}
\hspace{-0.6 cm}
\subfigure{
\includegraphics[width=1.2 in]{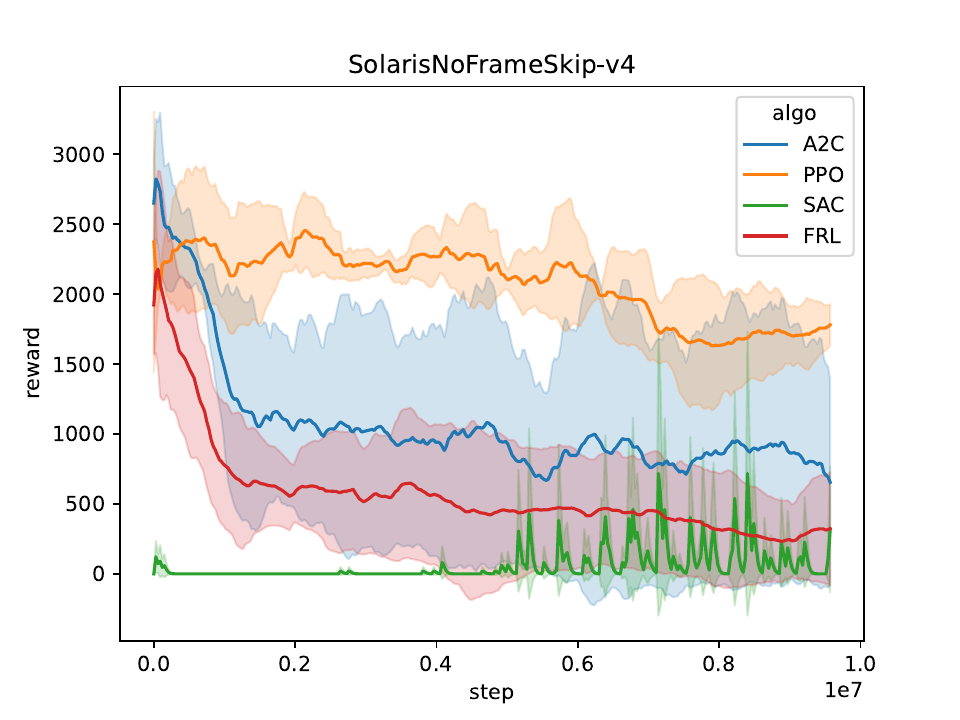}
}
\hspace{-0.6 cm}
\subfigure{
\includegraphics[width=1.2 in]{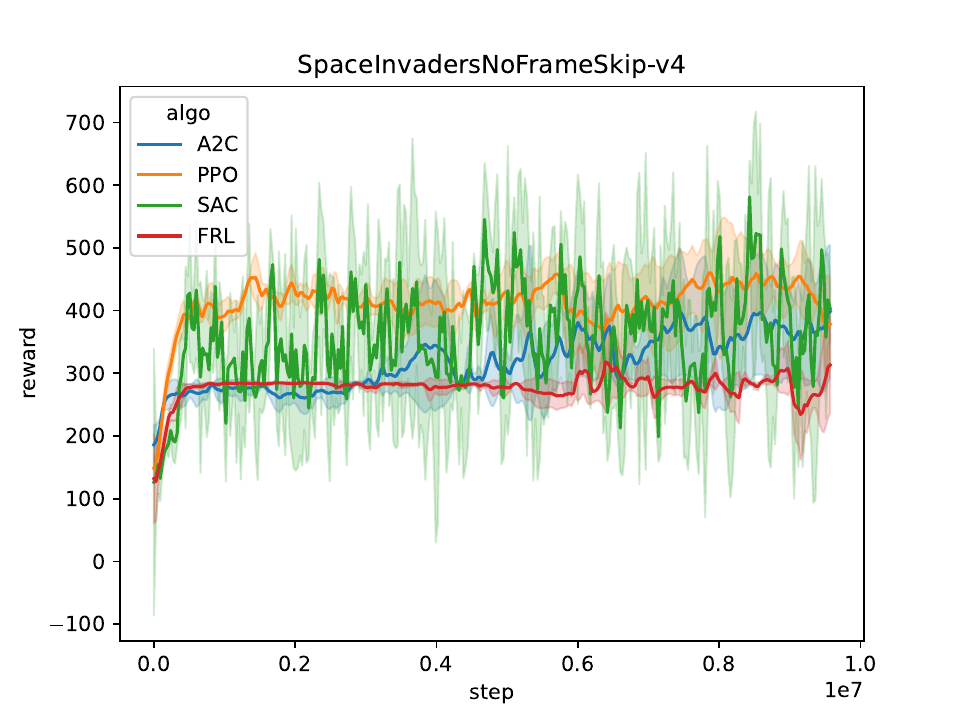}
}
\hspace{-0.6 cm}
\subfigure{
\includegraphics[width=1.2 in]{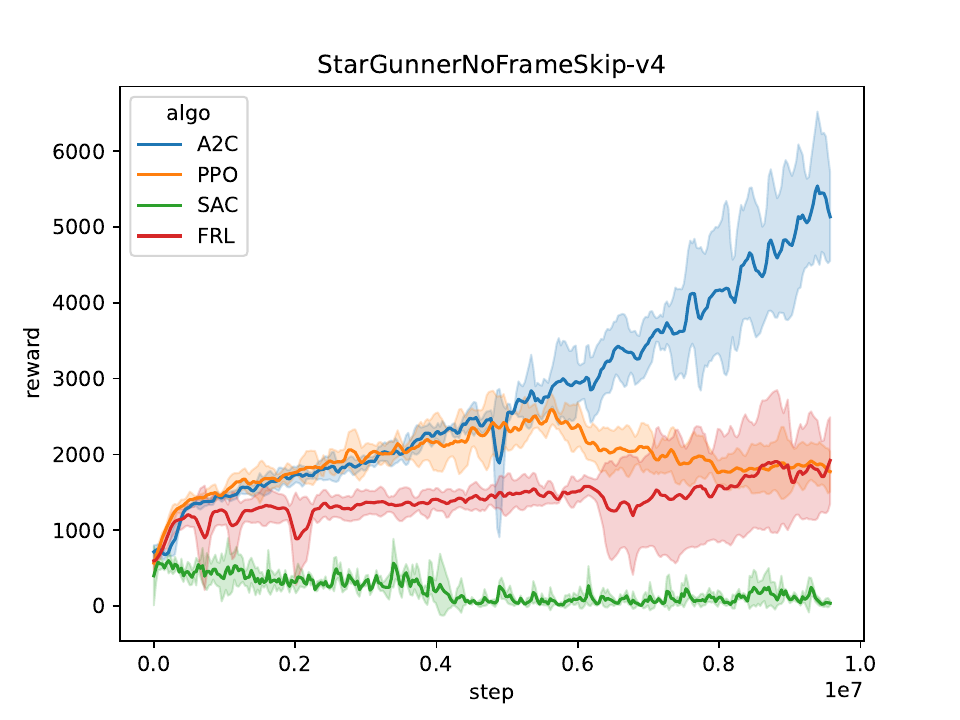}
}
\hspace{-0.6 cm}
\subfigure{
\includegraphics[width=1.2 in]{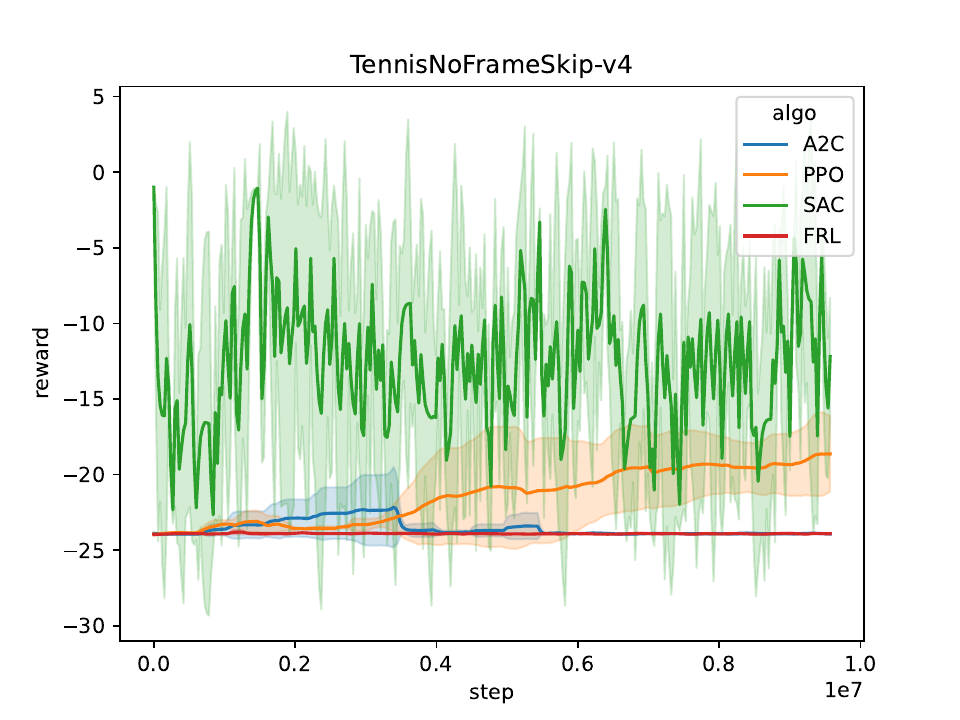}
}
% \vspace{-0.4 cm}
\subfigure{
\includegraphics[width=1.2 in]{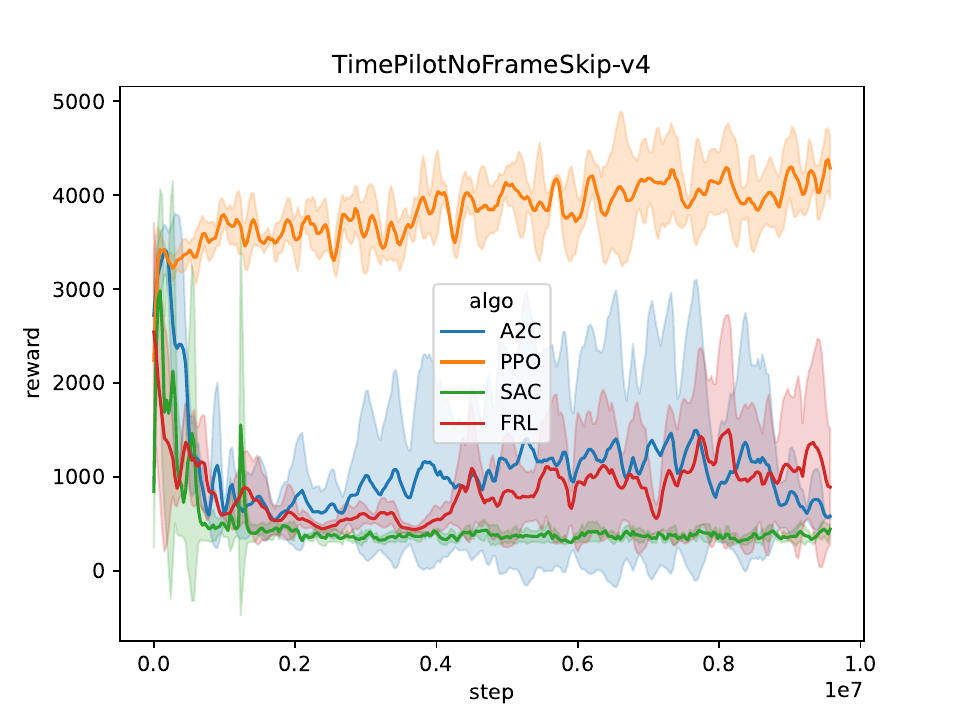}
}
\hspace{-0.6 cm}
\subfigure{
\includegraphics[width=1.2 in]{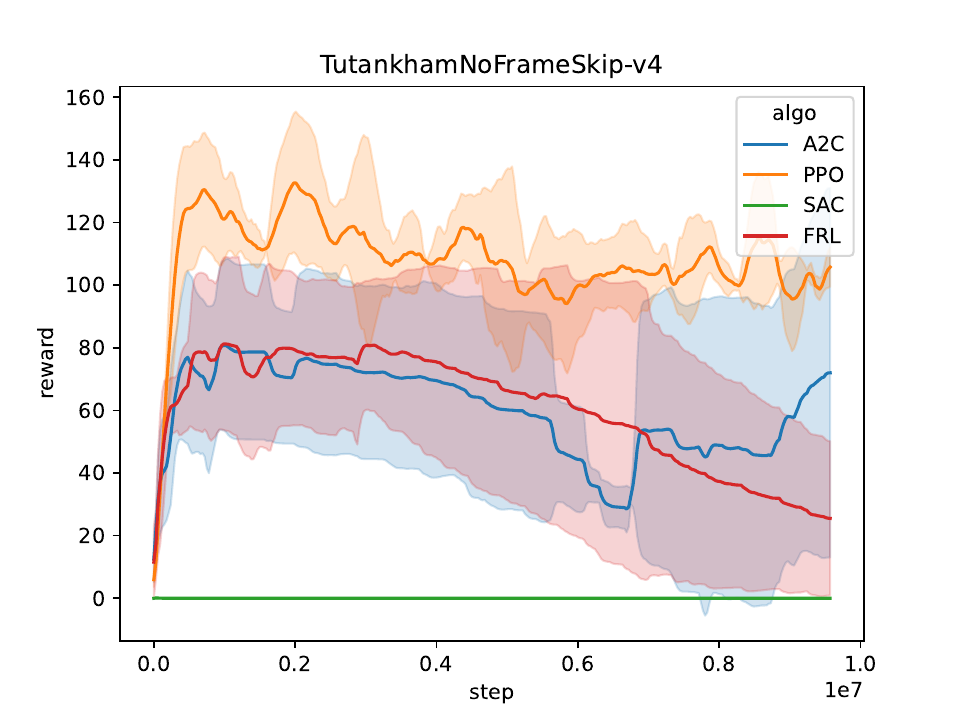}
}
\hspace{-0.6 cm}
\subfigure{
\includegraphics[width=1.2 in]{atari_rewards/UpNDown.pdf}
}
\hspace{-0.6 cm}
\subfigure{
\includegraphics[width=1.2 in]{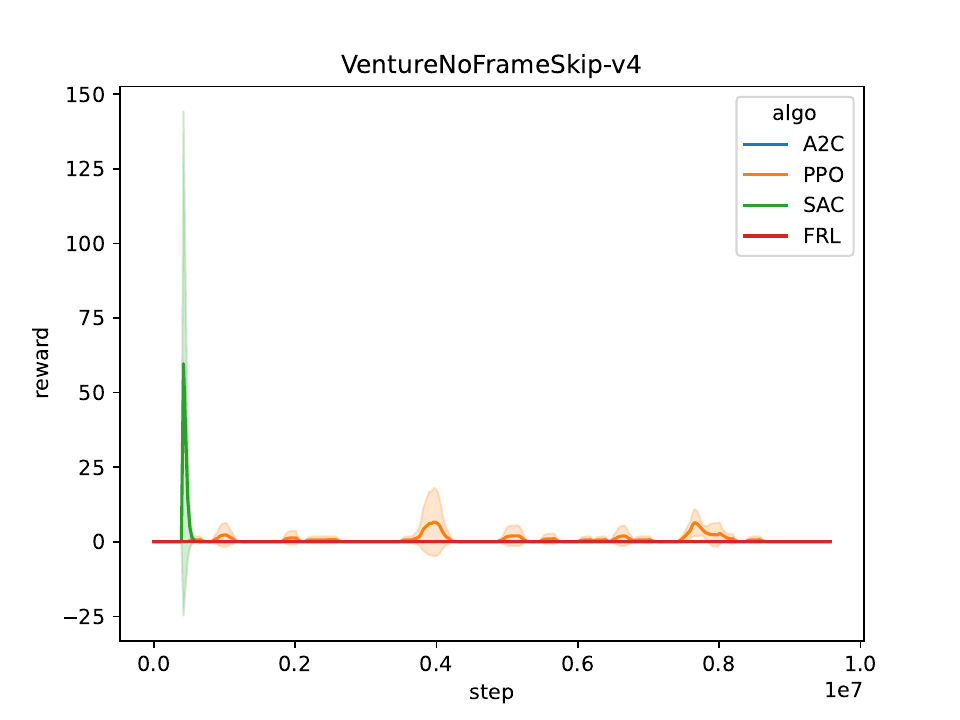}
}
\hspace{-0.6 cm}
\subfigure{
\includegraphics[width=1.2 in]{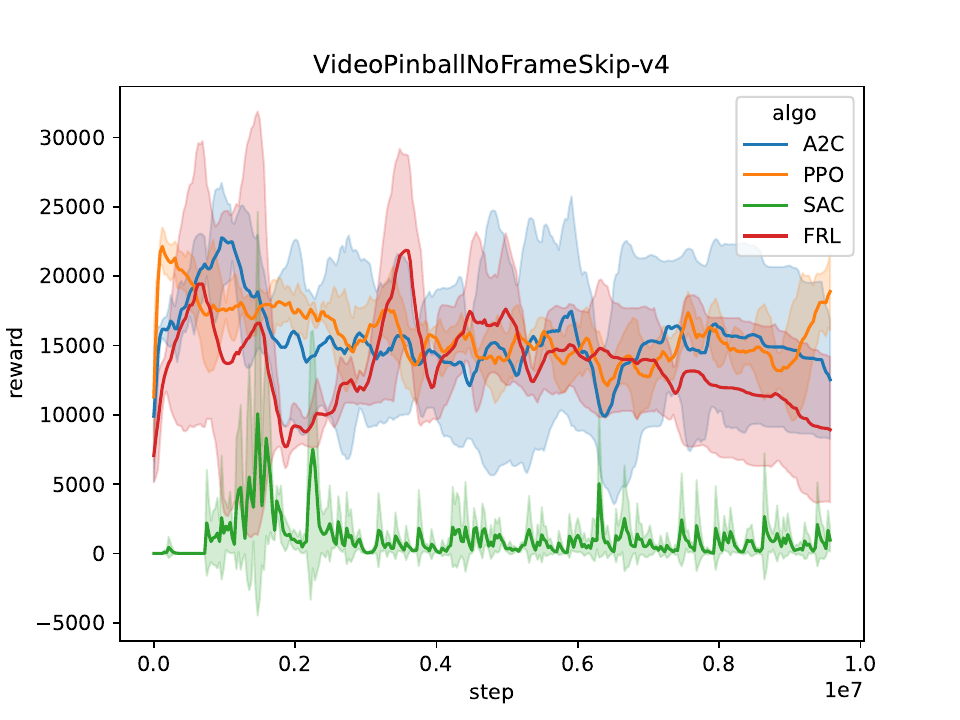}
}
\hspace{-0.6 cm}
\subfigure{
\includegraphics[width=1.2 in]{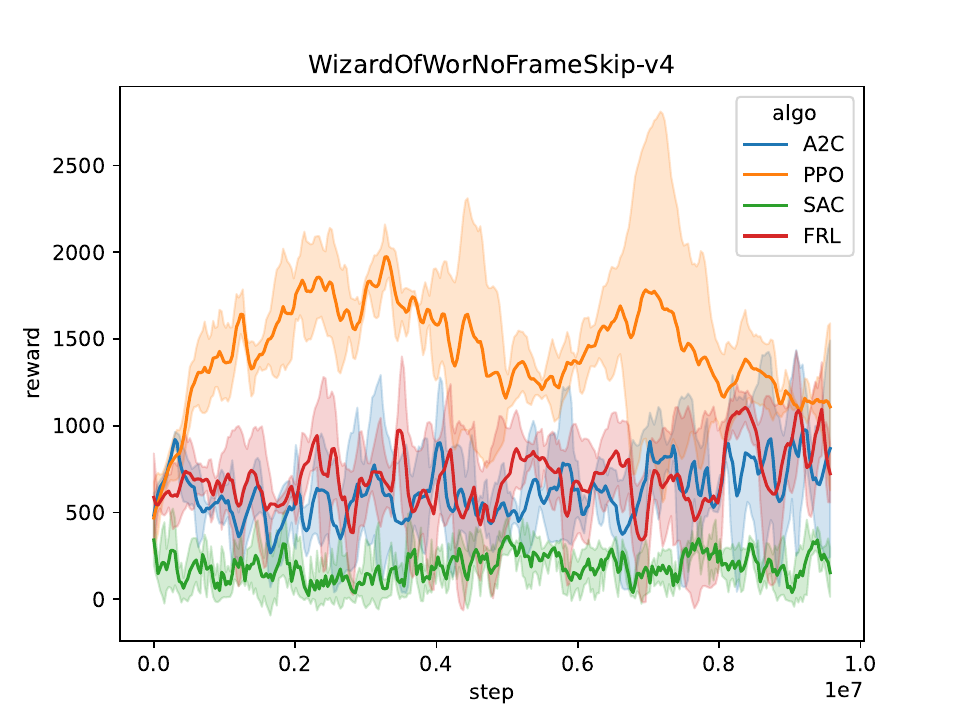}
}

\begin{flushleft}
% \hspace{0.6 cm}
% \vspace{-0.4 cm}
\hspace{0.2 cm}
\subfigure{
\includegraphics[width=1.2 in]{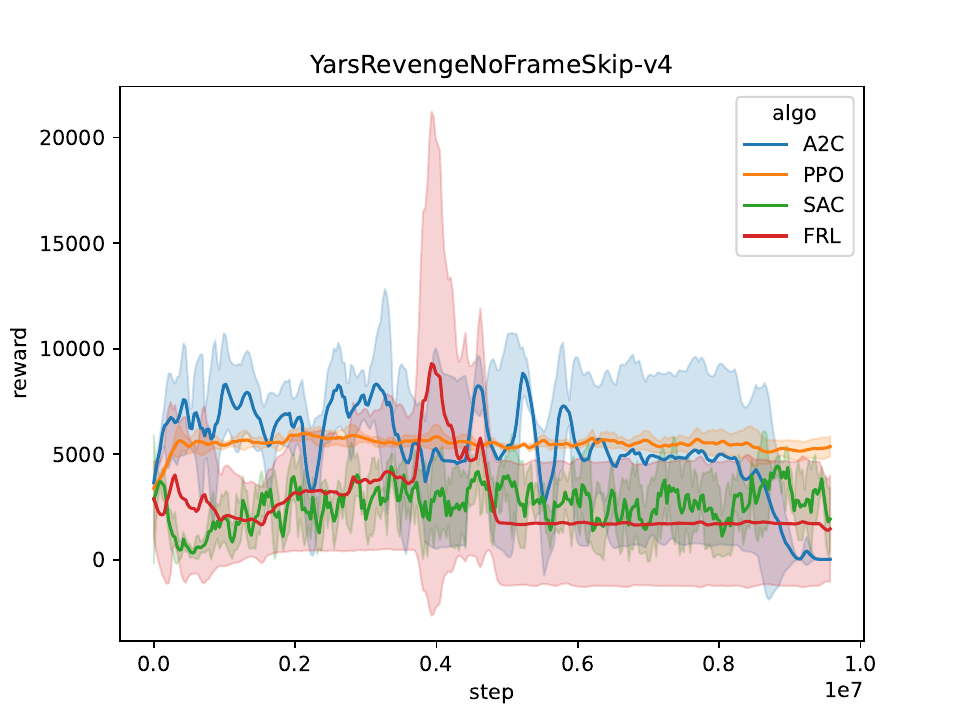}
}
\hspace{-0.6 cm}
\subfigure{
\includegraphics[width=1.2 in]{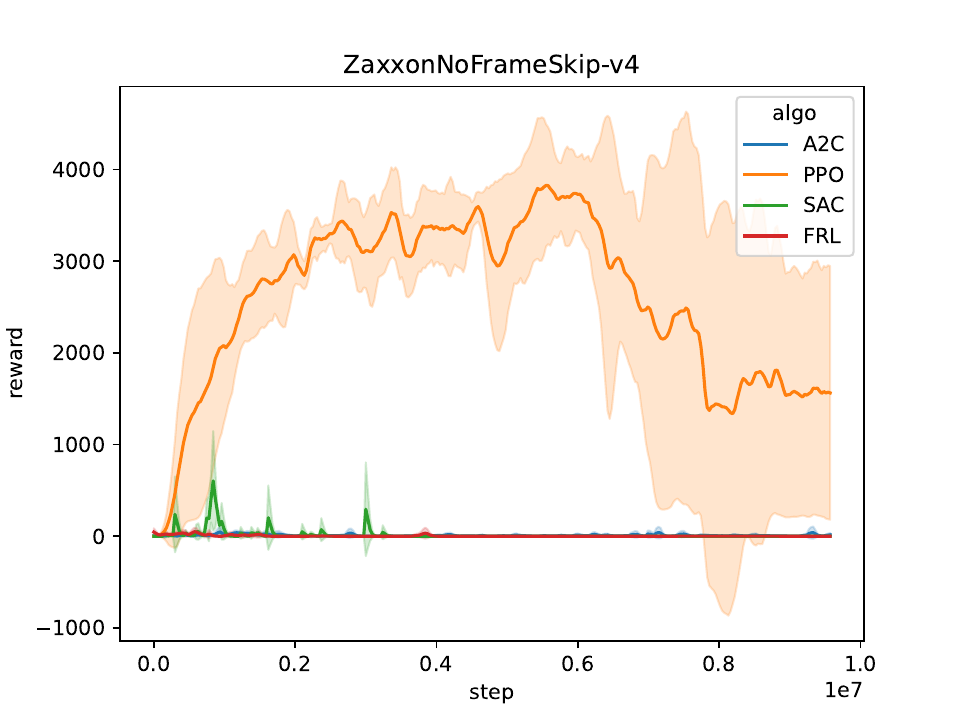}
}
\end{flushleft}
\caption{Performance curves with FRL, A2C, PPO and SAC for the total Atari 2600 video games. Each curve are smoothed by the exponential moving average to with 3 step. The solid line is the mean and the shaded region represents a standard deviation of the average evaluation over five seeds.} 
\label{fig:total_rewards}
% \vspace{-0.4 cm}
\end{figure*}

In this section, we introduce the experimental setup. The Implementation of our method based on the \texttt{Pytorch}\footnote{https://github.com/pytorch/pytorch} and \texttt{Stable Baselines}\footnote{https://github.com/Stable-Baselines-Team/stable-baselines} packages. For the A2C, PPO algorithms, we utilize the implementations of Stable-Baselines3, which are well-known implementations of the DRL methods. What's more, \texttt{Stable-Baselines3} has a wide range of applications in DRL research. For the implementations of A2C and PPO, we dose not change the default hyper-parameters. However, \texttt{Stable-Baselines3} only realize the SAC for the continuous environments, so the implementation of SAC for discrete environments we refer to \cite{sacd}. The experiments are performed on the total Atari \footnote{https://gym.openai.com/envs/\#atari} games (see Figure \ref{fig:atari}, and we use NoFrameskip-v4). Researchers commonly adopt Atari games in academia and industry to evaluate reinforcement learning
algorithms. Besides, the experimental results are present in Figure \ref{fig:total_rewards}. 

For the implementation of FRL, we refer to the A2C algorithms achievement by Stable-baseline3. For fairness, we keep the same hyper-parameters to A2C implementations. For each experiment, we perform 10 million time steps for training. There are 5 agents carried out simultaneously for each seed, and each agent utilizes 8 steps return for the policy gradient estimation. Each experiment runs with 5 different random seeds, and all the curves are averaged by 5 random seeds.

The network of feature extraction used a convolutional layer with 32 filters of size $8 \times 8$ with stride 4, followed by a convolutional layer with 64 filters of size $4 \times 4$ with stride 2, followed by a convolutional layer with 64 filters of size $4 \times 4$ with stride 1,  followed by a fully-connected layer with 512 hidden units. All four hidden layers are followed by ReLU activation function. Because the input of $Q$ network is only state, the $Q$ network and policy network share the feature extraction network. The $Q$ network has a single linear output unit for each action meaning the action value. The policy network is a softmax output representing the
probability of selecting the action. All experiments
use a discount of $ \gamma = 0.99$ and an RMSProp \cite{rmsprop} optimizer decay parameter = $0.99$ with learning rate = $0.0007$. The detail description about
our hyper-parameters are presented in Table \ref{table:hyper-parameters}. We summarize the FRL algorithm with $f(x) = \frac{1}{2}(x-1)^2$ on Algorithm \ref{alg:training process}. One may extend FRL framework for other proper, convex, lower semi-continuous $f$ function.

\bigskip

% \begin{algorithm}
%     \caption{$f-$Divergence Reinforcement Learning Framework (FRL)}
%     \label{algorithm1}
%     \textbf{Requires}: $\theta$: initialize $Q$ value network parameters; $\phi$: initialize policy network parameters; 
%     $s_0$: initial state; $T$: the counter; \\
%     \Repeat{$T > T_{\text{max}}$}{
%         Sample a batch of data $\{s_t,a_t,r_t,s_{t+1}\}_{t=0}^n$ from enviroment \\
%         \For{$k=0,1,\cdots,n-2$}{
%             \# Calculate the quantile distribution of BRD \\
%             \For{$m=0,1,\cdots,N-2$}{
%                 $h_m(s_k) = r_k + \gamma \theta_m(s_{k+1}) - \theta_m(s_{k}) $ \\
%                 $h_m(s_{k+1}) = r_{k+1} + \gamma \theta_m(s_{k+2}) - \theta_m(s_{k+1}) $ \\
%             }
%             \# Calculate the Wasserstein distance  \\ % approximately
%             $W_{1}(U_\theta(s_k), U_\theta(s_{k+1})) = \frac{1}{N} \sum_{i=1}^N \sum_{j=1}^{N}\rho_{\tau_i}^\kappa(\theta_j^M(s_{k+1}) - \theta_i^M(s_{k}))$ \\
%             $\Delta \theta \leftarrow \frac{1}{n-1} \left[ \nabla_\theta \mathcal{L}(\theta) + \lambda \nabla_\theta   W_{1}(U_\theta(s_k), U_\theta(s_{k+1}))\right]$ \\
%             $\theta \leftarrow \theta + \epsilon \Delta \theta$
%         }
        
%         $T \leftarrow T + 1$ \\
%     }
% \end{algorithm}
\begin{table}[htbp]
 \caption{Hyper-parameters settings for our implementation.}
 \label{table:hyper-parameters}
 \renewcommand\tabcolsep{3.0pt} 
 \centering
 \begin{tabular}{l|c}
  \toprule
  \textbf{Hyper-parameter}&Value\\
  \midrule
  \textit{Shared hyper-parameters} & \\
  discount ($\gamma$) & 0.99 \\
  Learning rate for actor & $7 \times 10^{-4}$ \\
  Learning rate for critic & $7 \times 10^{-4}$ \\
  Number of hidden layer for all networks & 4 \\
  Number of hidden units per layer & 512 \\
  Activation function & ReLU \\
  Mini-batch size & 64  \\
  Gradient Clipping & 0.5 \\
  Value net coefficient & 0.5\\
  \midrule
  \textit{FRL} & \\
  Target Entropy & None \\
  Number of step & 5 \\
  Optimizer & RMSprop \\
%   RMSProp decay parameter & 0.99 \\
  \midrule
  \textit{A2C} & \\
  Number of step & 5 \\
  Target Entropy &  - dim of $\mathcal{A}$\\
  Optimizer & RMSprop \\
%   RMSProp decay parameter & 0.99 \\
  \midrule
  \textit{PPO} & \\
  Number of step & 2048 \\
  Target Entropy & None \\
  Learning rate for $\alpha$ & $3\times 10^{-4}$ \\
  Optimizer & Adam \\
  Replay buffer size & $3 \times 10^6$ \\
  Range of clip & 0.2 \\
  \midrule
  \textit{SAC} & \\
  Target Entropy & - dim of $\mathcal{A}$ \\
  Learning rate for $\alpha$ & $7\times 10^{-4}$ \\
  Optimizer & Adam \\
  Replay buffer size & $3 \times 10^6$ \\
  \bottomrule
 \end{tabular}
\end{table}

\subsection*{Appendix F}
In this subsection, we describe in detail the experiments on the overestimation of the $Q$ function. The experiments in this section use the same set of training hyper-parameters and policy network/state-value function
architectures described in Appendix E. We compare the $\text{Loss}_{\text{value}}$ between the true value and the estimated value, where the $\text{Loss}_{\text{value}}$ is to measure the severity of overestimation of the value function. The $\text{Loss}_{\text{value}}$ is defined as
\begin{equation}
    \text{Loss}_{\text{value}} = \mathbb{E}_\pi\left[Q^\pi(s,a)\right] - \text{Value}_{ \text{true}}. 
\end{equation}
Obviously, when the $\text{Loss}_{\text{value}}$ is too large, the overestimation problem of value is serious. On the contrary, if $\text{Loss}_{\text{value}}$ is less than zero, there is a problem with underestimation. For an excellent algorithm, the ideal situation is that the $\text{Loss}_{\text{value}}$ is close to 0. For each time step, $\mathbb{E}_\pi\left[Q^\pi(s,a)\right]$ is calculated by the inner product of the output of our policy network and $Q$ network. Besides, for the calculation of $\text{Value}_{\text{true}}$, we record all the rewards the for sample trajectories, and use $\sum_{i=1}^t \gamma^{i-t} r_t$ to obtain the true value. The curves of $\text{Loss}_{\text{value}}$ for a subset of Atari games are shown in Figure \ref{fig:q_over_complete}.

\begin{figure*}[htb]
    \centering
\subfigure{
\includegraphics[width=1.2 in]{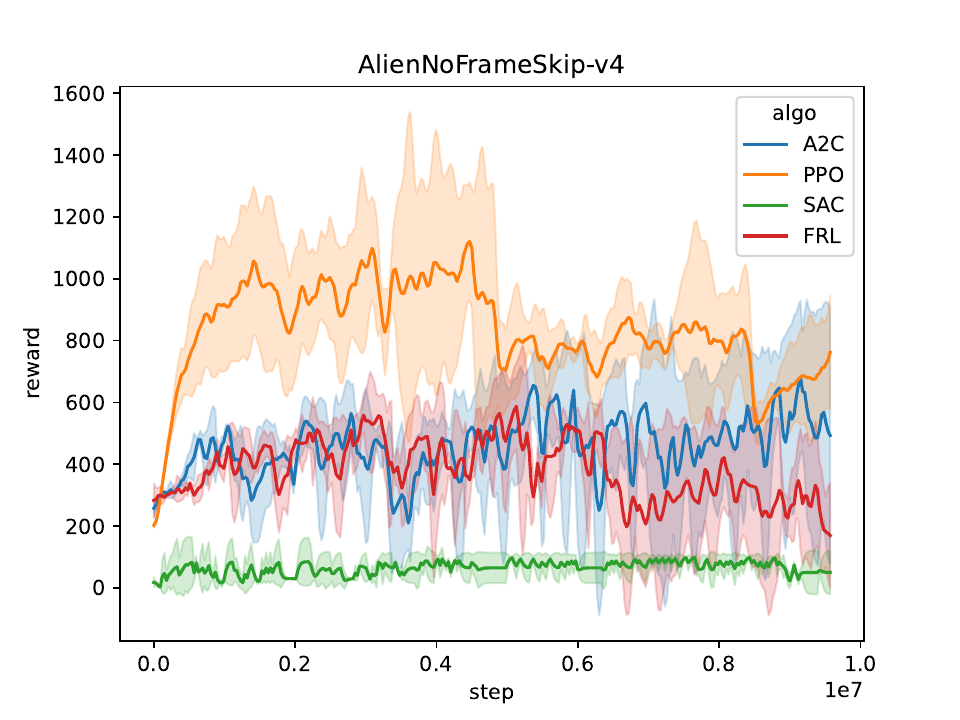}
}
\hspace{-0.6 cm}
\subfigure{
\includegraphics[width=1.2 in]{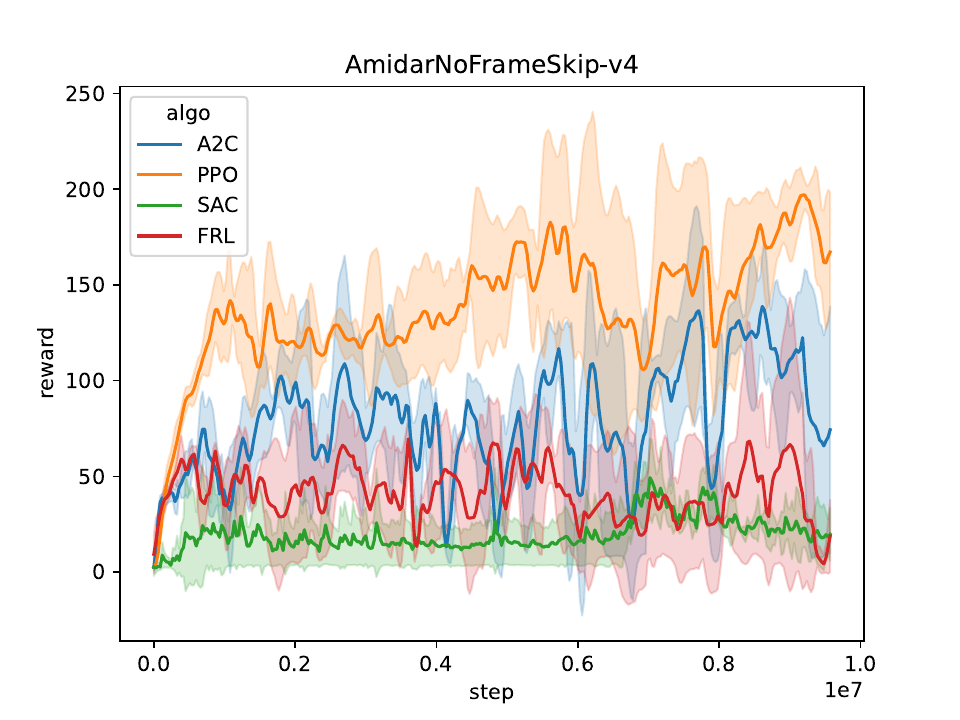}
}
\hspace{-0.6 cm}
\subfigure{
\includegraphics[width=1.2 in]{atari_rewards/Assault.pdf}
}
\hspace{-0.6 cm}
\subfigure{
\includegraphics[width=1.2 in]{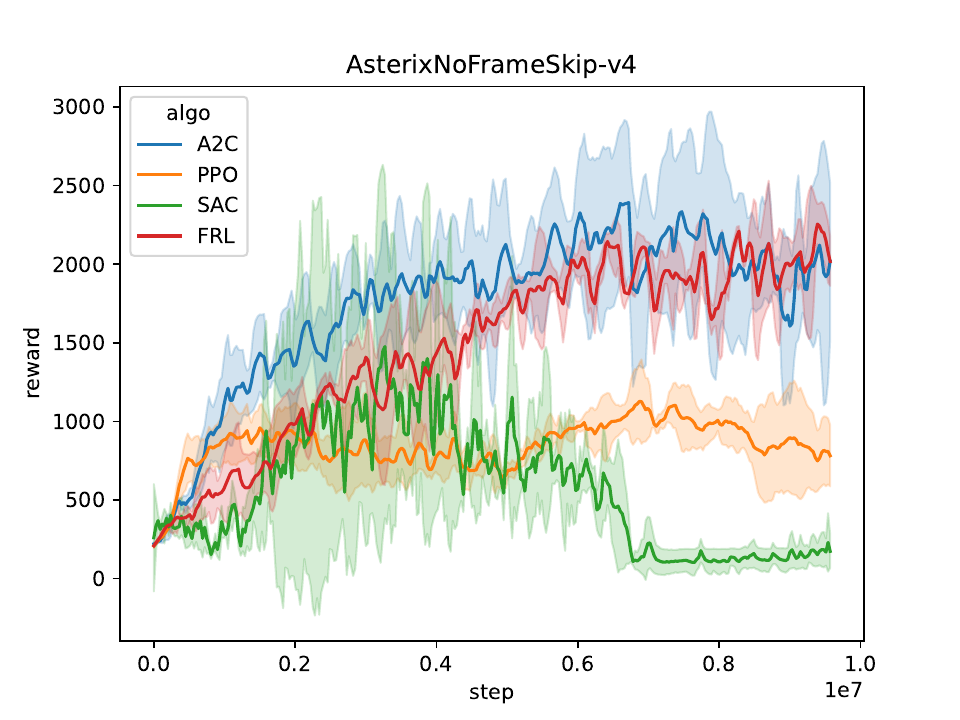}
}
\hspace{-0.6 cm}
\subfigure{
\includegraphics[width=1.2 in]{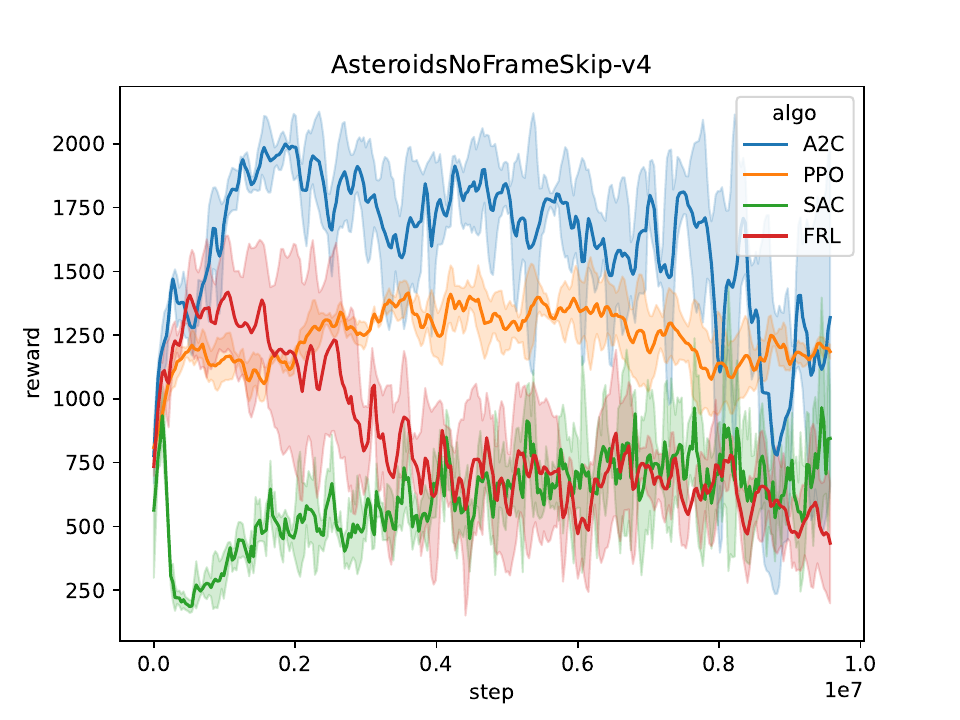}
}
\hspace{-0.6 cm}
\subfigure{
\includegraphics[width=1.2 in]{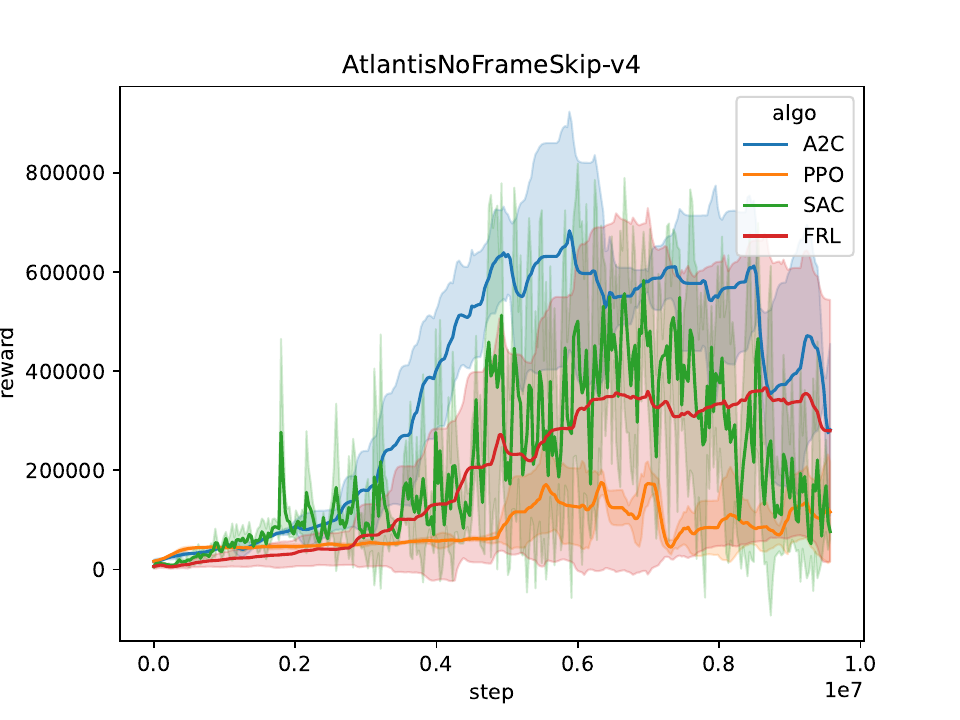}
}
\subfigure{
\includegraphics[width=1.2 in]{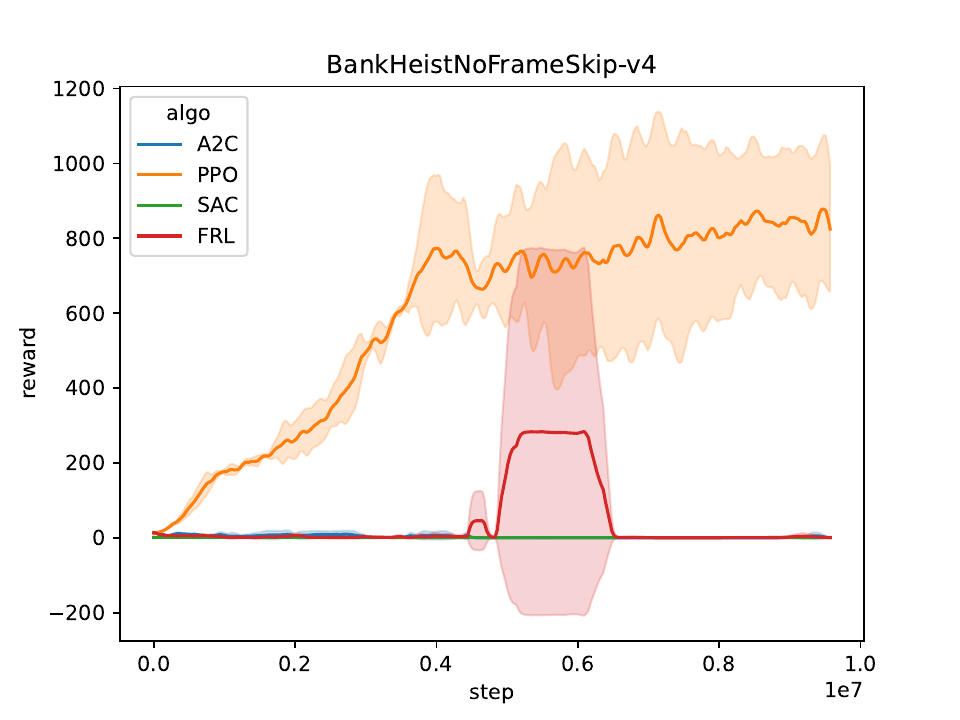}
}
\hspace{-0.6 cm}
\subfigure{
\includegraphics[width=1.2 in]{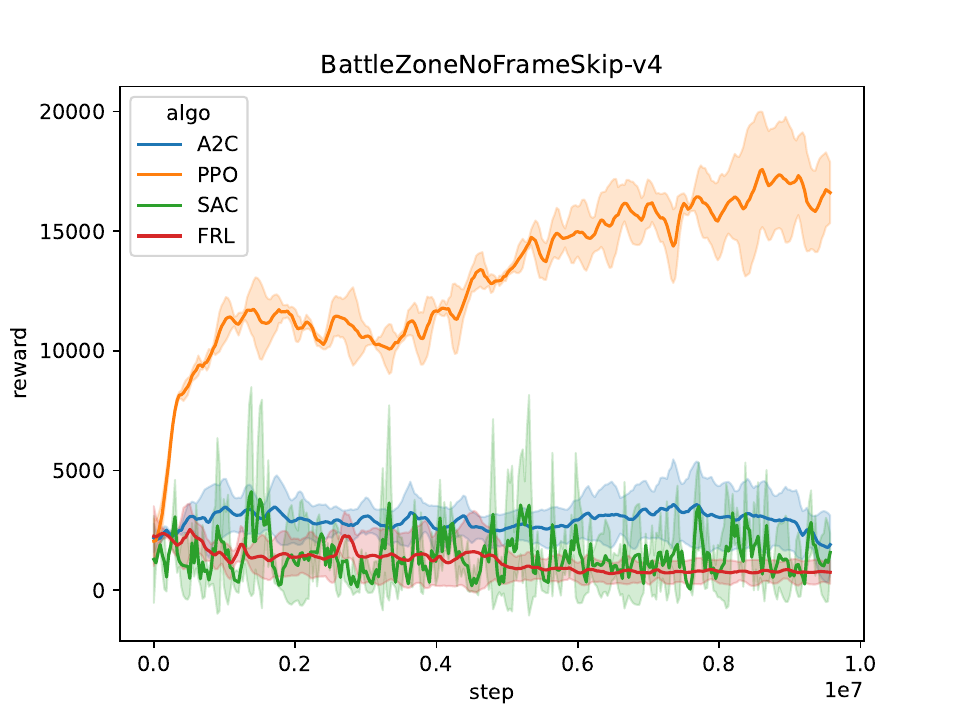}
}
\hspace{-0.6 cm}
\subfigure{
\includegraphics[width=1.2 in]{atari_rewards/BeamRider.pdf}
}
\hspace{-0.6 cm}
\subfigure{
\includegraphics[width=1.2 in]{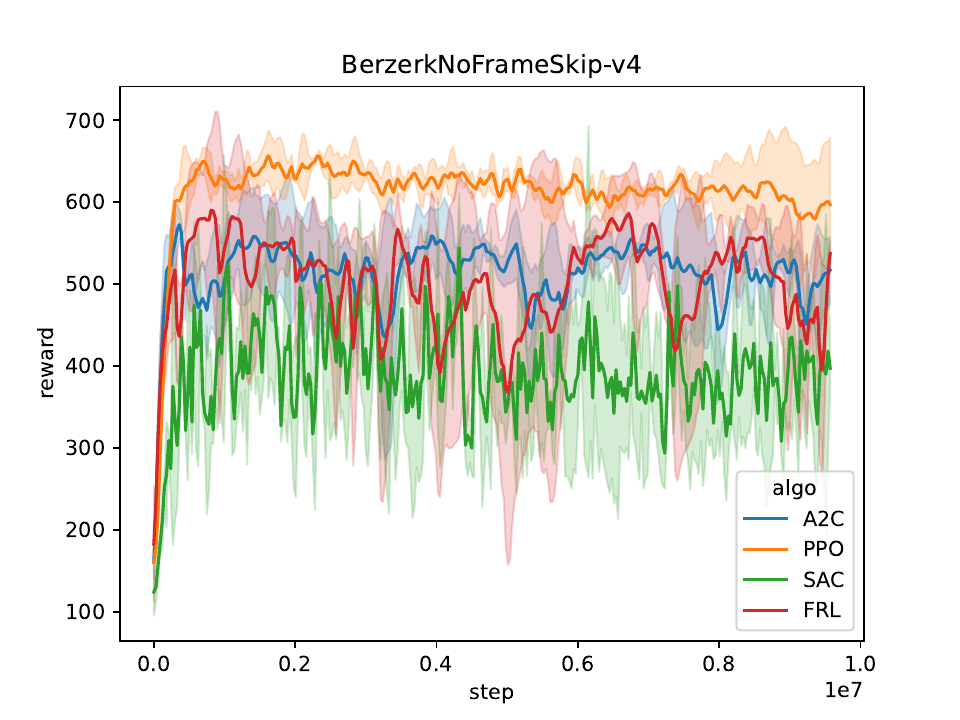}
}
\hspace{-0.6 cm}
\subfigure{
\includegraphics[width=1.2 in]{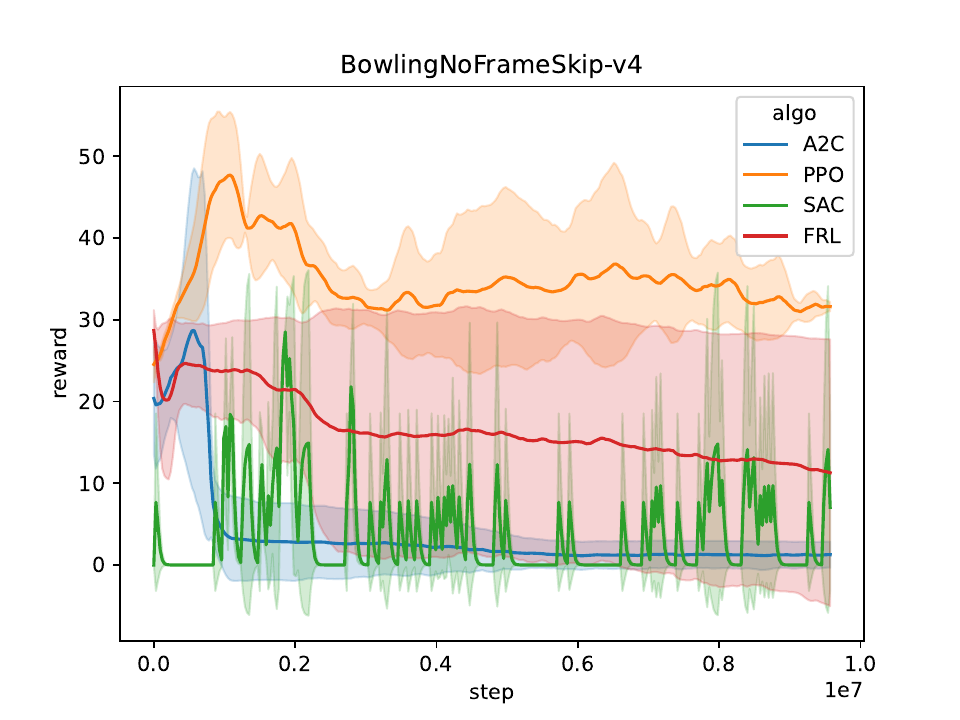}
}
\hspace{-0.6 cm}
\subfigure{
\includegraphics[width=1.2 in]{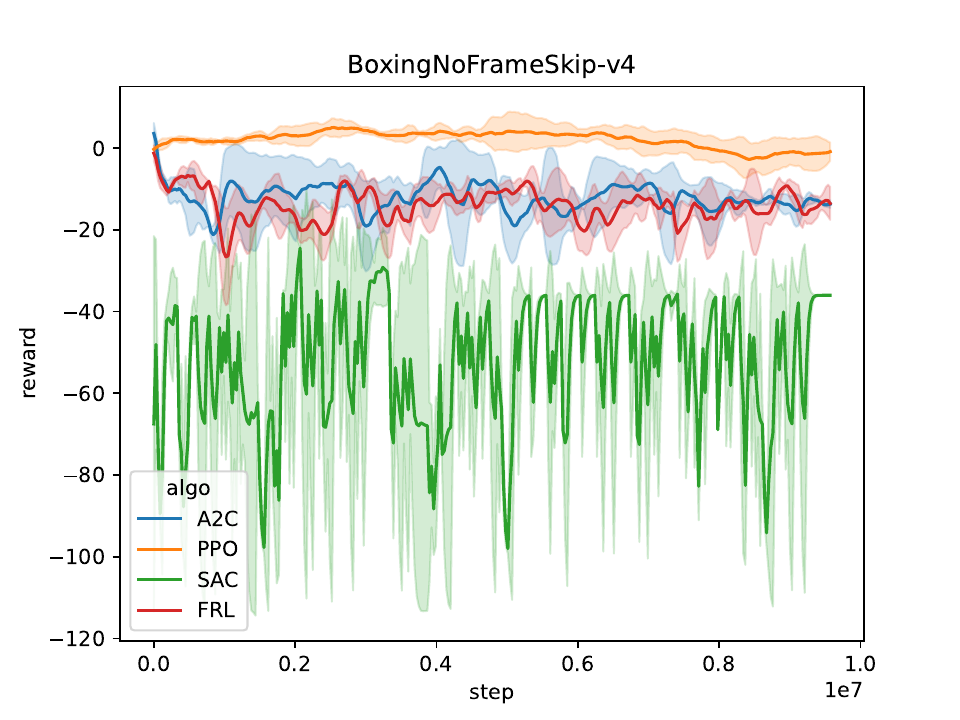}
}
\subfigure{
\includegraphics[width=1.2 in]{atari_rewards/Breakout.pdf}
}
\hspace{-0.6 cm}
\subfigure{
\includegraphics[width=1.2 in]{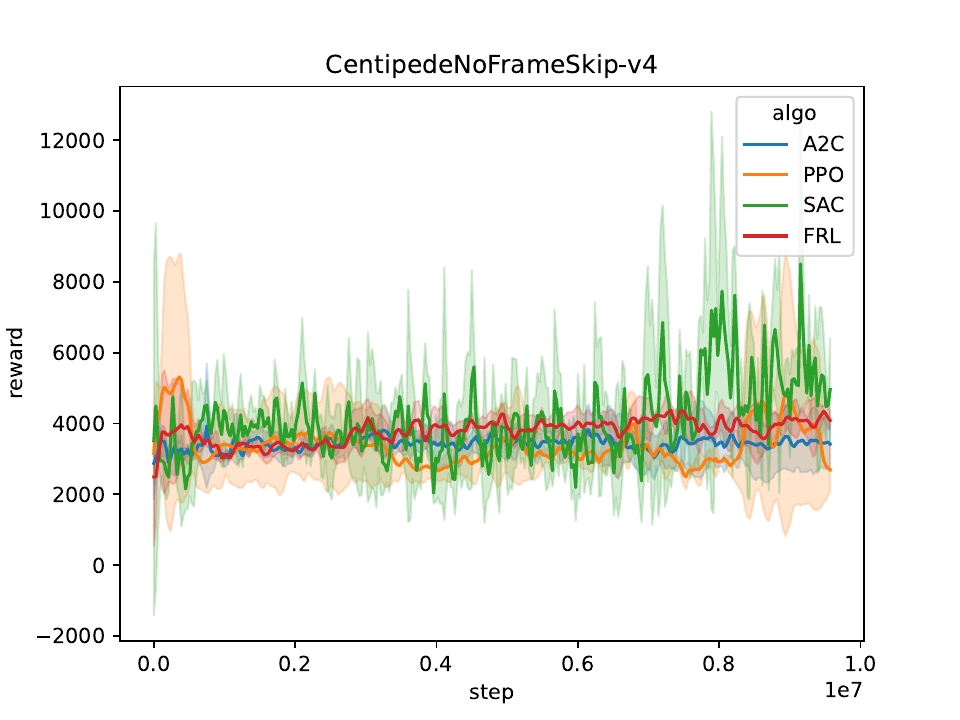}
}
\hspace{-0.6 cm}
\subfigure{
\includegraphics[width=1.2 in]{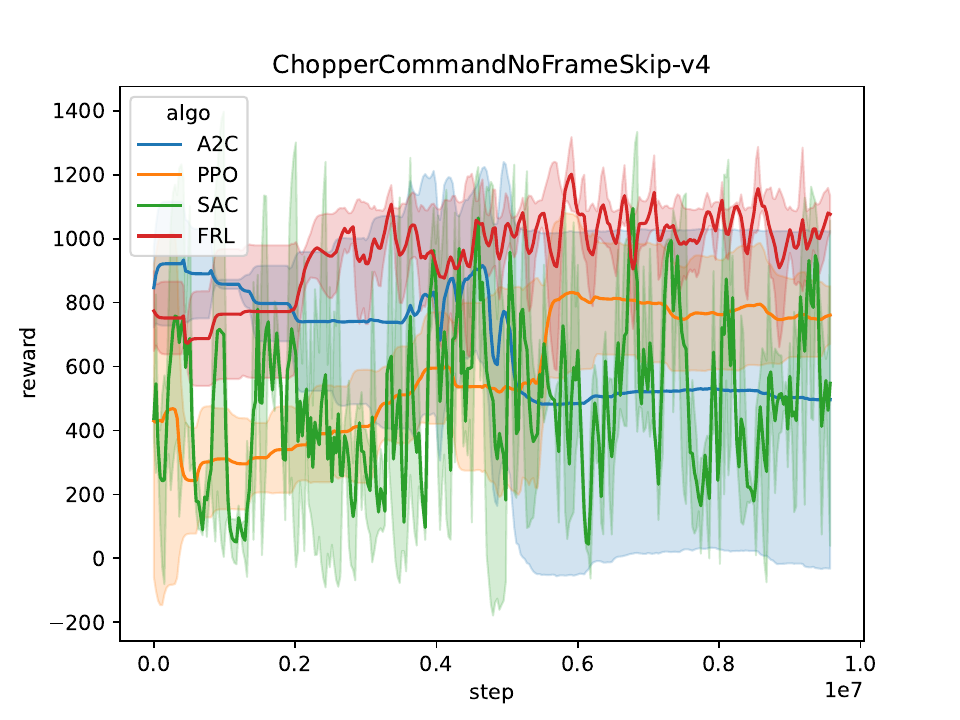}
}
\hspace{-0.6 cm}
\subfigure{
\includegraphics[width=1.2 in]{atari_rewards/CrazyClimber.pdf}
}
\hspace{-0.6 cm}
\subfigure{
\includegraphics[width=1.2 in]{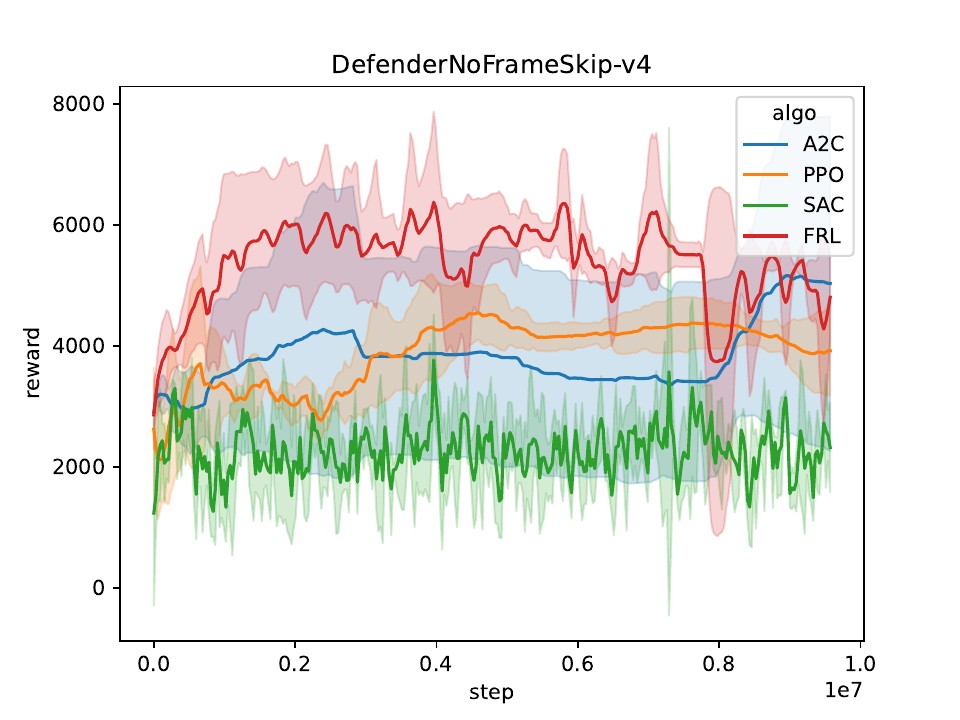}
}
\hspace{-0.6 cm}
\subfigure{
\includegraphics[width=1.2 in]{atari_rewards/DemonAttack.pdf}
}
\end{figure*}
\end{document}